\pdfoutput=1

\documentclass[12pt, twoside,openright]{book}
\usepackage[T1]{fontenc} 
\usepackage[english]{babel}
\usepackage[table,xcdraw,dvipsnames]{xcolor}
\usepackage{rotating}
\usepackage{svg}
\usepackage[ruled,vlined]{algorithm2e}
\usepackage{algorithmic}
\usepackage{amsmath}
\usepackage{pdfpages}
\usepackage{subcaption}
\usepackage{stmaryrd}   
\usepackage{tcolorbox}
\usepackage[hyphens]{url}
\usepackage{multirow}
\usepackage{multicol}
\usepackage{colortbl} 
\usepackage{paralist} 
\usepackage{booktabs} 
\usepackage{setspace} 
\usepackage{graphicx}
\usepackage{latexsym}
\usepackage{float}
\usepackage{titlesec}
\usepackage{caption}
\usepackage{placeins}
\usepackage{enumitem}
\usepackage{times}
\usepackage{amsfonts}
\usepackage{svg}
\usepackage{tabularx}
\usepackage{latexsym}
\usepackage[]{todonotes}
\usepackage{adjustbox}
\usepackage{wrapfig}
\usepackage{arydshln}
\usepackage{amsmath}
\usepackage{tcolorbox}
\usepackage{scalerel} 

\newcolumntype{g}{>{\columncolor{Gray!10}}c}

\DeclareUnicodeCharacter{1EBC}{\~E}
\DeclareUnicodeCharacter{1EBD}{\~e}
\DeclareUnicodeCharacter{1ABC}{\~A}
\DeclareUnicodeCharacter{1ABD}{\~a}

\usepackage{url}
\usepackage{dsfont}

\usepackage{wasysym}

\usepackage{etoolbox} 
\makeatletter
\patchcmd{\ttlh@hang}{\parindent\z@}{\parindent\z@\leavevmode}{}{}
\patchcmd{\ttlh@hang}{\noindent}{}{}{}
\makeatother

\usepackage{soul} 
\usepackage{amssymb}
\usepackage{pifont}
\usepackage[Sonny]{fncychap}

\usepackage{fancyhdr}

\usepackage[width=0.9\textwidth,font=small,labelsep=endash,labelfont=bf]{caption}

\usepackage[comma]{natbib} 

\usepackage{tocbibind} 
\usepackage[pdfstartview=FitH, pdftitle={Cross-Lingual Transfer for Low-Resource Natural Language Processing}, pdfauthor={Iker García-Ferrero}, pdfproducer={pdfLaTeX}]{hyperref} 

\usepackage{textcomp} 
\usepackage{listings}
\usepackage{estilo/Tesis}

\tcbuselibrary{many}
\tcbset{
    enhanced,
    breakable,
    attach boxed title to top left={
        xshift=0.5cm,
        yshift= -3.5mm, 
    },
    top=4mm,
    coltitle=black,
    beforeafter skip=\baselineskip,
}

\definecolor{boxbackground_tesis_contributions}{HTML}{FFE6CC}
\definecolor{boxborder_tesis_contributions}{HTML}{D79B00}
\definecolor{titlebackground_tesis_contributions}{HTML}{D79B00} 
\definecolor{titletext_tesis_contributions}{HTML}{000000}

\newenvironment{part_of_the_thesis}[1]{%
    \begin{tcolorbox}[colback=boxbackground_tesis_contributions, 
                      colframe=boxborder_tesis_contributions, 
                      title={\parbox{\linewidth}{\centering#1}}, 
                      coltitle=titletext_tesis_contributions, 
                      colbacktitle=titlebackground_tesis_contributions, 
                      fonttitle=\bfseries,
                      rounded corners, 
                      boxrule=0.5mm,
                      width=\textwidth,
                      title={\parbox{\dimexpr\linewidth-2mm\relax}{#1}}]}%
 {\end{tcolorbox}}

 \newenvironment{smallbox}[0]{%
    \begin{tcolorbox}[colback=boxbackground_tesis_contributions, 
                      colframe=boxborder_tesis_contributions, 
                      boxrule=0.5mm,
                      width=0.36\textwidth,
                      before=\hskip-3mm,
                      after=\newline\hskip-2mm,
                      left=2mm, right=1mm, top=0mm, bottom=0mm]}%
 {\end{tcolorbox}}

\definecolor{boxbackground_not_tesis_contributions}{HTML}{DAE8FC}
\definecolor{boxborder_not_tesis_contributions}{HTML}{6C8EBF}
\definecolor{titlebackground_not_tesis_contributions}{HTML}{6C8EBF} 
\definecolor{titletext_not_tesis_contributions}{HTML}{000000}

 \newenvironment{not_part_of_the_thesis}[1]{%
    \begin{tcolorbox}[colback=boxbackground_not_tesis_contributions, 
                      colframe=boxborder_not_tesis_contributions, 
                      title={\parbox{\linewidth}{\centering#1}}, 
                      coltitle=titletext_not_tesis_contributions, 
                      colbacktitle=titlebackground_not_tesis_contributions, 
                      fonttitle=\bfseries,
                      rounded corners, 
                      boxrule=0.5mm,
                      width=\textwidth,
                      title={\parbox{\dimexpr\linewidth-2mm\relax}{#1}}]}%
 {\end{tcolorbox}}

 \definecolor{boxbackground_resources}{HTML}{D5E8D4}
\definecolor{boxborder_resources}{HTML}{82B366}
\definecolor{titlebackground_resources}{HTML}{82B366} 
\definecolor{titletext_resources}{HTML}{000000}

 \newenvironment{resources}[1]{%
    \begin{tcolorbox}[colback=boxbackground_resources, 
                      colframe=boxborder_resources, 
                      title={\parbox{\linewidth}{\centering#1}}, 
                      coltitle=titletext_resources, 
                      colbacktitle=titlebackground_resources, 
                      fonttitle=\bfseries,
                      rounded corners, 
                      boxrule=0.5mm,
                      width=\textwidth,
                      title={\parbox{\dimexpr\linewidth-2mm\relax}{#1}}]}%
 {\end{tcolorbox}}
 
 \newenvironment{resourcessmall}[0]{%
 \begin{tcolorbox}[colback=boxbackground_resources, 
                   colframe=boxborder_resources, 
                   boxrule=0.5mm,
                   width=0.38\textwidth,
                   before=\hskip-3mm,
                   after=\newline\hskip-2mm,
                   left=2mm, right=1mm, top=0mm, bottom=0mm]}%
{\end{tcolorbox}}

\usepackage{estilo/gb4e}
\noautomath 

\begin{document}

\frontmatter  

\titleformat{\chapter}[display]
  {\Large\bfseries}
  {\filleft\normalfont{\Huge\thechapter. \large\MakeUppercase{\chaptertitlename}}}
  {4ex}
  {\titlerule
    \vspace{2ex}%
    \filleft}
  [\vspace{2ex}%
   \titlerule]

\includepdf[]{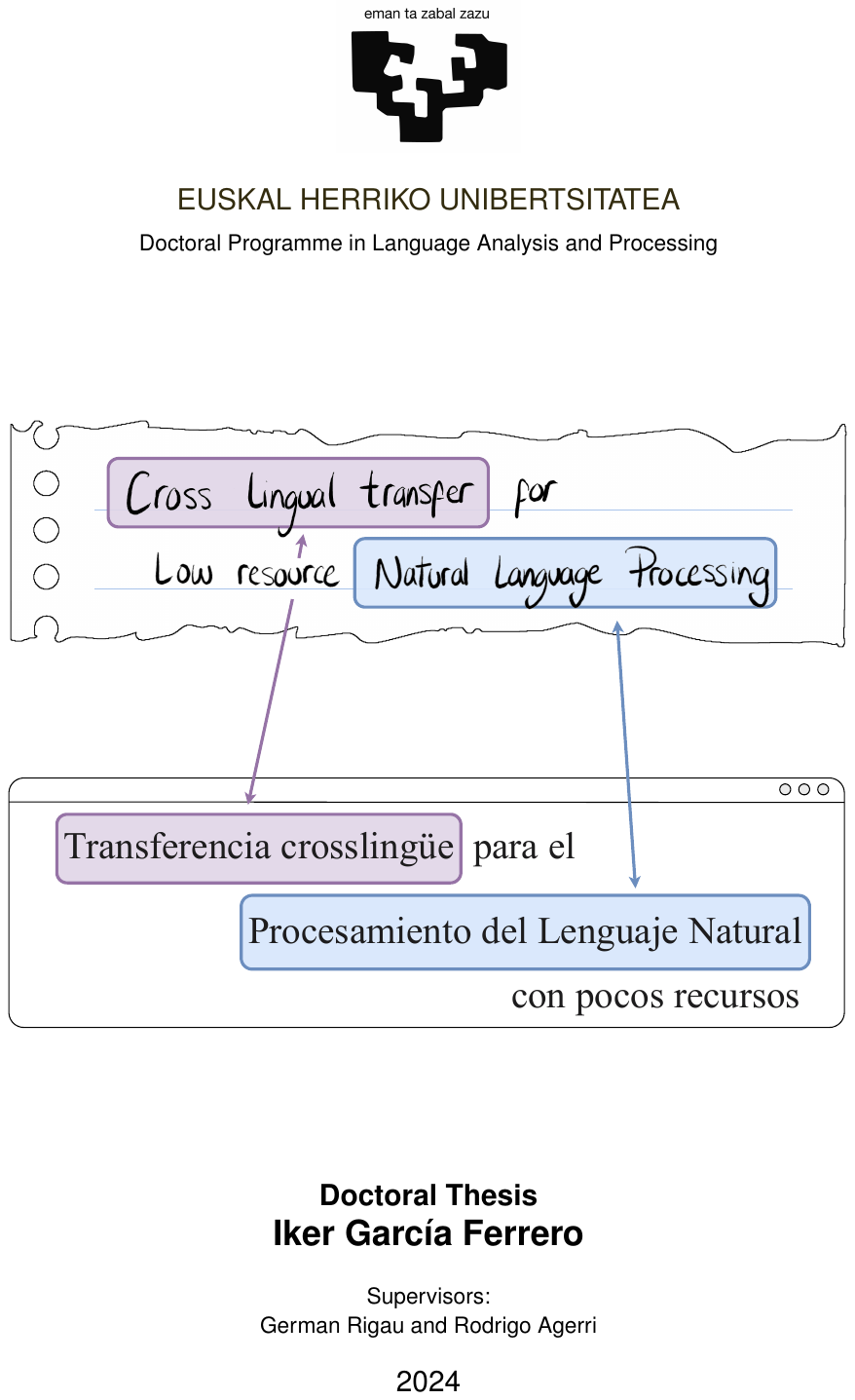}
\cleardoublepage
\setcounter{page}{1} 
\thispagestyle{empty}

\begin{center}
  \includegraphics[width=0.25\textwidth]{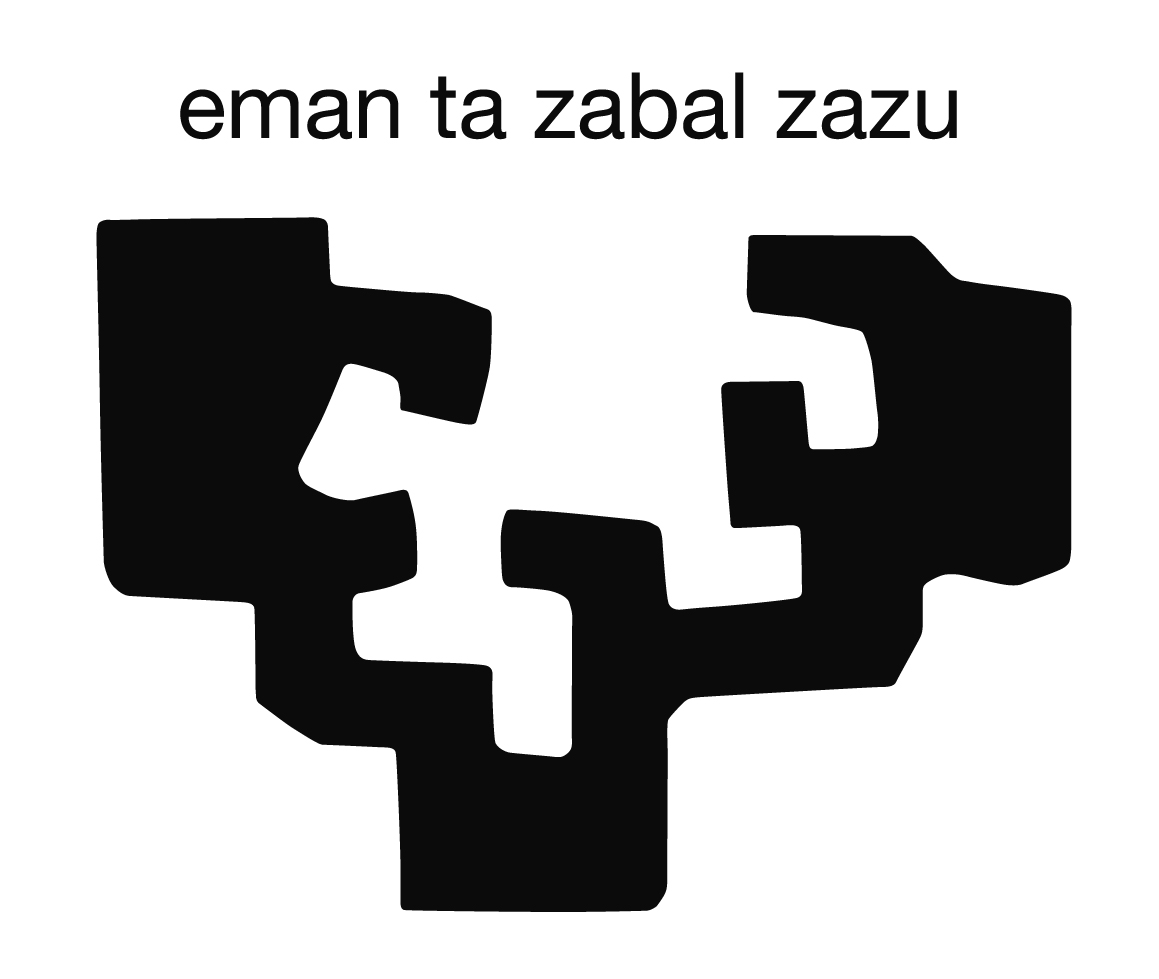} \\[0.3cm]
  \textsf{EUSKAL HERRIKO UNIBERTSITATEA}\\[0.15cm]
   \textsf{Doctoral Programme in Language Analysis and Processing}\\[2cm]

\vspace{1cm}
{ \LARGE 
\begin{spacing}{1}
\textbf{\fontsize{19}{22} \selectfont Cross-lingual Transfer for Low-Resource Natural Language Processing}
\end{spacing}
}

\vspace{2.5cm}
\end{center}

\hspace{5cm}
\begin{minipage}{8.1cm}
This thesis report was made by Iker García Ferrero under the supervision of German Rigau and Rodrigo Agerri, and submitted to obtain a PhD degree at the University of the Basque Country UPV/EHU\\
\vspace{0.5cm}

\noindent Donostia, December 2024.

\end{minipage} 
\cleardoublepage
\thispagestyle{empty}
\label{aipua}

\vspace*{10mm}

\begin{center}

$\cdots$

``You must never think of the whole street at once, understand? You must only concentrate on the next step, the next breath, the next stroke of the broom, and the next, and the next. Nothing else.''

\smallskip
\smallskip
Again he paused for thought before adding, ``That way you enjoy your work, 
which is important, because then you make a good job of it. And that's how 
it ought to be.''

$\cdots$

\end{center}

\begin{flushright}
Michael Ende (MOMO, 1973)
\end{flushright}

\cleardoublepage
\chapter*{Acknowledgments}

Thank you / Gracias / Eskerrik Asko ...

\vspace{5mm}

... A German Rigau, por haberme guiado desde que era un alumno de grado que no sabía a qué quería dedicarse. Gracias por enseñarme la pasión por la investigación.

\vspace{5mm}

... A Rodrigo Agerri, por ayudarme a poner orden en los brainstormings y tablas de resultados infinitas, y por enseñarme cómo ser un buen investigador.

\vspace{5mm}

... A mi familia, por haberme dado la oportunidad de poder estudiar y trabajar en algo que me apasiona y me hace disfrutar. Y por haberme apoyado en todo el camino. 

\vspace{5mm}

...  A Raquel Barbero, por descubrirme el mundo fuera de las teclas, por ayudarme a desconectar cuando no sabía desconectar, y por apoyarme siempre. 

\vspace{5mm}

... IXA taldeari. Lanera joatea inoiz ez delako lanera joatea bezala sentitzen. Egun bakoitza dibertigarria egiteagatik. Eta eman didazuen laguntza guztiagatik.

\vspace{5mm}

... 318 bulegori, brainstorming ordu guztiengatik eta ideia on guztiengatik. Eskerrik asko Ander Salaberriari bihurrikeria guztietan konplize izateagatik eta Oscar Sainzi elkarrekin egin dugun lan guztiengatik.

\vspace{5mm}

... to Dan Roth for hosting me at the University of Pennsylvania, Jennifer Sheffield for all the work that made it possible, and all the members of the Cognitive Computation Group for giving me the chance to collaborate with you.

\clearpage

This thesis has been supported by a PhD Grant from the Basque Government (PRE\_2020\_2\_0208).

\cleardoublepage

\pagestyle{fancy}
\cleardoublepage
\phantomsection 
\addcontentsline{toc}{chapter}{Abstract}
\selectlanguage{english}
\chapter*{Abstract}

Natural Language Processing (NLP) has seen remarkable advances in recent years, particularly with the emergence of Large Language Models that have achieved unprecedented performance across many tasks. However, these developments have mainly benefited a small number of high-resource languages such as English. The majority of languages still face significant challenges due to the scarcity of training data and computational resources. To address this issue, this thesis focuses on cross-lingual transfer learning, a research area aimed at leveraging data and models from high-resource languages to improve NLP performance for low-resource languages. Specifically, we focus on Sequence Labeling tasks such as Named Entity Recognition, Opinion Target Extraction, and Argument Mining.

The research is structured around three main objectives: (1) advancing data-based cross-lingual transfer learning methods through improved translation and annotation projection techniques, (2) developing enhanced model-based transfer learning approaches utilizing state-of-the-art multilingual models, and (3) applying these methods to real-world problems while creating open-source resources that facilitate future research in low-resource NLP.

More specifically, this thesis presents a new method to improve data-based transfer with T-Projection, a state-of-the-art annotation projection method that leverages text-to-text multilingual models and machine translation systems. T-Projection significantly outperforms previous annotation projection methods by a wide margin. For model-based transfer, we introduce a constrained decoding algorithm that enhances cross-lingual Sequence Labeling in zero-shot settings using text-to-text models. Finally, we develop Medical mT5, the first multilingual text-to-text medical model, demonstrating the practical impact of our research on real-world applications.
\cleardoublepage 
\phantomsection 
\addcontentsline{toc}{chapter}{Resumen}
\selectlanguage{spanish}

\chapter*{Resumen}

El Procesamiento del Lenguaje Natural (PLN) ha experimentado avances notables en los últimos años, particularmente con la aparición de Modelos de Lenguaje de Gran Tamaño que han logrado un rendimiento sin precedentes en numerosas tareas. Sin embargo, estos desarrollos han beneficiado principalmente a un pequeño número de idiomas con abundantes recursos, como el inglés. Así, la mayoría de los idiomas aún se enfrentan a desafíos significativos debido a la escasez de datos de entrenamiento y recursos computacionales. Para abordar este problema, esta tesis se centra en el aprendizaje por transferencia crosslingüe, un área de investigación destinada a aprovechar los datos y modelos de idiomas con abundantes recursos para mejorar el rendimiento del PLN en idiomas con recursos más limitados. Específicamente, nos esta tesis se enfoca en tareas de Etiquetado Secuencial como el Reconocimiento de Entidades Nombradas, la Extracción de Foco  de Opinión y la Minería de Argumentos.

La investigación se estructura en torno a tres objetivos principales: (1) avanzar en los métodos de aprendizaje por transferencia crosslingüe basados en datos mediante técnicas mejoradas de traducción y proyección de anotaciones, (2) desarrollar enfoques mejorados de aprendizaje por transferencia basados modelos multilingües de última generación, y (3) aplicar estos métodos a problemas del mundo real mediante la creación de recursos de código abierto que faciliten la investigación futura en PLN con recursos limitados.

Más concretamente, en esta tesis se presenta un nuevo método para mejorar la transferencia basada en datos con T-Projection, una técnica de proyección de anotaciones de última generación que aprovecha los modelos multilingües texto-a-texto y los sistemas de traducción automática. T-Projection supera significativamente todos los métodos anteriores de proyección de anotaciones. Para la transferencia basada en modelos, introducimos un algoritmo de decodificación restringida que mejora el Etiquetado Secuencial crosslingüe en entornos sin recursos utilizando modelos texto-a-texto. Finalmente, desarrollamos Medical mT5, el primer modelo médico multilingüe texto-a-texto, demostrando el impacto práctico de nuestra investigación en aplicaciones del mundo real.
\cleardoublepage 
\phantomsection 
\addcontentsline{toc}{chapter}{Laburpena}
\selectlanguage{basque}

\chapter*{Laburpena}

Hizkuntzaren Prozesamenduan aurrerapen nabarmenak ikusi dira azken urteetan, bereziki ataza askotan aurrekaririk gabeko errendimendua lortu duten Hizkuntza Eredu Handien agerpenarekin. Hala ere, garapen hauek batez ere baliabide handiko hizkuntza gutxi batzuen onurarako izan dira, ingelesa kasu. Hizkuntza gehienek oraindik ere erronka handiei aurre egin behar diete entrenamendu-datuen eta baliabide konputazionalen urritasuna dela eta. Arazo honi aurre egiteko, tesi honek hizkuntzen arteko transferentzia-ikasketan jartzen du arreta, hots, baliabide handiko hizkuntzetako datuak eta ereduak aprobetxatuz baliabide urriko Hizkuntzetarako Prozesamenduanaren errendimendua hobetzea helburu duen ikerketa-arloan. Zehazki, Sekuentzia Etiketatze atazetan zentratzen gara, hala nola Izendun Entitateen Erauzketan, Iritzien Xedeen Erauzketan eta Argudio Meatzaritzan.

Ikerketa hiru helburu nagusiren inguruan egituratzen da: (1) datuetan oinarritutako hizkuntzen arteko transferentzia-ikasketa metodoak hobetzea itzulpen eta anotazio-proiekzio tekniken bidez, (2) ereduetan oinarritutako transferentzia-ikasketa hurbilpenak garatzea puntako eredu eleaniztunak erabiliz, eta (3) metodo hauek benetako arazoei aplikatzea, baliabide urriko Hizkuntzetarako Prozesamenduan etorkizuneko ikerketa erraztuko duten kode irekiko baliabideak sortuz.

Zehazki, datuen transferentzia hobetzen dugu T-Projection bidez, testutik testurako eredu eleaniztunak eta itzulpen automatikoko sistemak erabiltzen dituen puntako anotazio-proiekzio metodoa. T-Projection metodoak nabarmen gainditzen ditu aurreko anotazio-proiekzio metodoak. Ereduetan oinarritutako transferentziarako, deskodifikazio murriztuko algoritmo bat aurkezten dugu, zero-shot testuinguruetan hizkuntzen arteko Sekuentzia Etiketatzea hobetzen duena testutik testurako ereduak erabiliz. Azkenik, Medical mT5 garatu dugu, testutik testurako lehen eredu mediko eleaniztuna, gure ikerketaren eragin praktikoa erakutsiz benetako aplikazioetan.
\cleardoublepage

\renewcommand{\contentsname}{Table of Contents}
\renewcommand{\listfigurename}{Figure List}
\renewcommand{\listtablename}{Table List}
\tableofcontents
\listoftables
\listoffigures

\mainmatter  
\setcounter{page}{1}

\selectlanguage{english}
\chapter[Introduction]{Introduction}
\label{ch:instroduction}

This thesis is framed within the area of Natural Language Processing (NLP). Natural Language Processing is a multidisciplinary research field within Artificial Intelligence (AI), Computer Science, and Linguistics. NLP involves a wide range of tasks, including, Natural Language Understanding, Machine Translation, Information Extraction, and Text Generation, among others. The main goal of NLP is to enable computers to understand, interpret, and generate human language in a way that is valuable for humans. The Ixa group, within the HiTZ center at the University of the Basque Country, is one of the leading research teams working in NLP. Since its foundation more than 30 years ago, the Ixa group has been a pioneer in developing NLP tools for many different applications, with a special focus on creating language tools for the Basque language. Moreover, Ixa has been involved in many European and international research projects, significantly contributing to languages beyond Basque.

The primary objective of this thesis is to develop cross-lingual transfer learning solutions to address the resource constraints faced by many languages, tasks, and domains. Cross-lingual transfer learning is a research area focused on creating models for low-resource languages by leveraging knowledge from high-resource languages. Specifically, this thesis explores cross-lingual transfer learning for Sequence Labeling tasks, such as Named Entity Recognition, Opinion Target Extraction, and Argument Mining. We propose novel methods for knowledge transfer from high-resource to low-resource languages through translation and annotation projection, as well as multilingual NLP models. Thus, our goal is to develop publicly available models that achieve state-of-the-art performance in low-resource languages and to make these models accessible to the research community. This thesis work was aligned with the objectives of the projects DeepReading\footnote{\url{https://ixa2.si.ehu.eus/deepreading/}}, DeepKnowledge \footnote{\url{http://ixa.si.ehu.es/node/13582}} and Andidote\footnote{\url{https://univ-cotedazur.eu/antidote}}.

\section{Motivation}
\label{motivation}

\begin{figure}[htb]
    \centering
    \includegraphics[width=\textwidth]{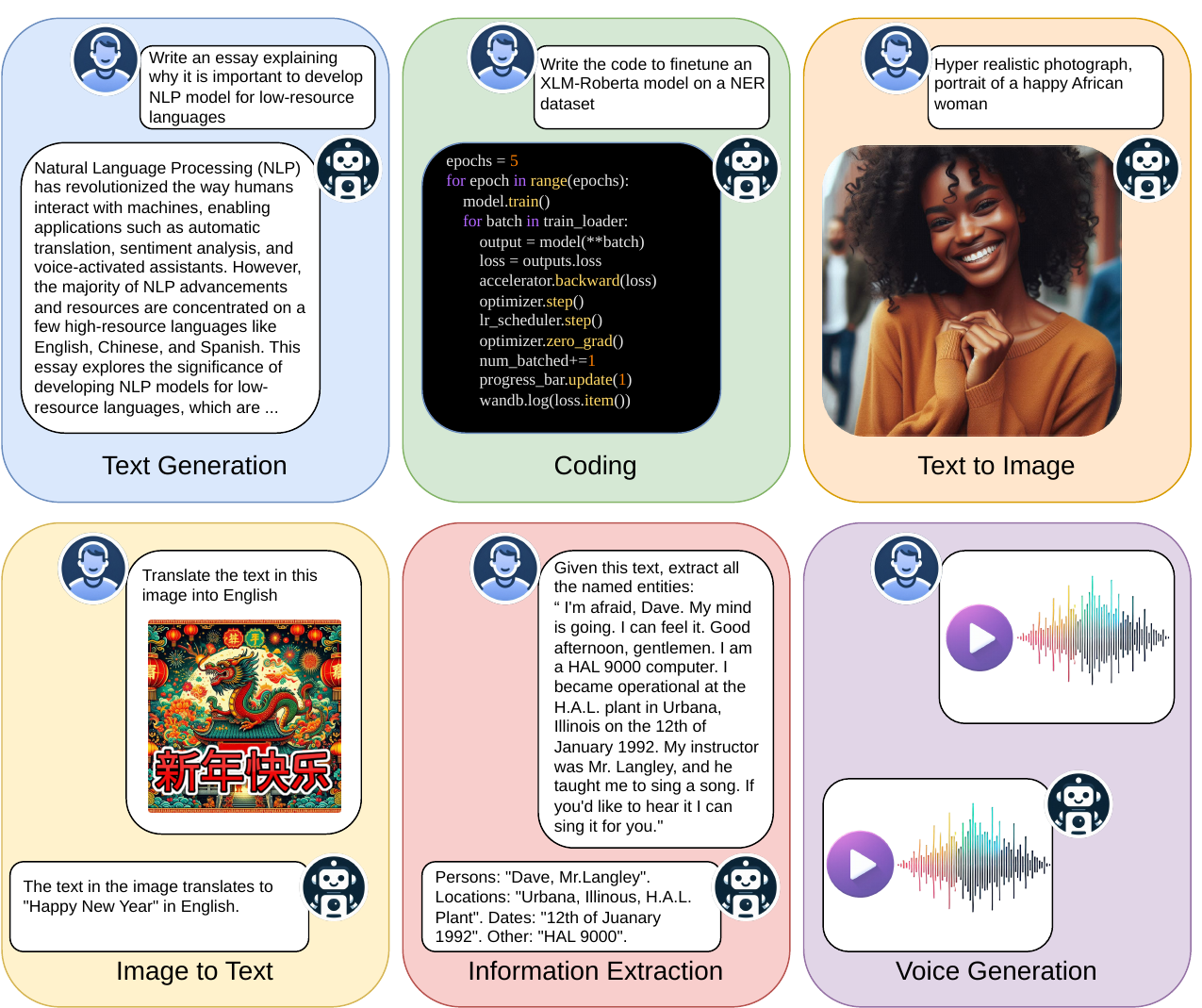}
    \caption{Modern LLMs, which support text, image, and other multimodal representations, have achieved outstanding performance in a wide range of NLP tasks. They have been applied in many real-world applications.}
    \label{fig:llms}
\end{figure}

Neural networks have become an indispensable resource in Natural Language Processing (NLP). Driven by the success of the Transformer architecture (\cite{DBLP:conf/nips/VaswaniSPUJGKP17}), they have demonstrated outstanding performance in various challenging NLP tasks (\cite{DBLP:journals/csur/MinRSVNSAHR24}), such as General Language Understanding (\cite{DBLP:conf/iclr/WangSMHLB19}), Question Answering (\cite{DBLP:conf/acl/RajpurkarJL18}), Text Generation (\cite{brown2020language}), Dialogue (\cite{DBLP:journals/corr/abs-2201-08239}), and Conditional Image Generation (\cite{DBLP:conf/cvpr/RombachBLEO22}), among others. Scaling up these models in terms of parameter count and training data (\cite{chung-flan-instruction-models}) has led to the development of current state-of-the-art NLP systems. Large Language Models (LLMs) such as GPT-4 (\cite{openai2024gpt4technicalreport}) and LLaMA-3 (\cite{llama3modelcard}), trained on hundreds of terabytes of text data and billions of parameters, have proven capable of generating human-like text and have been applied in a wide range of applications, such as the ones depicted in Figure \ref{fig:llms}. These cutting-edge NLP systems hold the potential to bring significant societal changes (\cite{DBLP:journals/corr/abs-2108-07258}).

Despite the remarkable progress in NLP, many challenges remain. LLMs require vast amounts of data and computational resources to achieve optimal performance (\cite{DBLP:journals/corr/abs-2203-15556}). In addition to English, only a handful of Western European languages (principally German, French, and Spanish) and even fewer non-Indo-European languages (primarily Chinese, Japanese, and Arabic) dominate the field (\cite{joshi-etal-2020-state}). While speakers of these languages benefit from the latest innovations in Language Technology—such as quick and accurate access to information using smart assistants, online translation services, interaction with machines using natural language, or speeding-up their work with automatic summarization tools, coding assistants, or image generation tools—speakers of low-resource languages are being left behind (\cite{blasi-etal-2022-systematic}).

Models consistently perform better on high-resource languages, especially English (\cite{etxaniz-etal-2024-multilingual}), while their performance on low-resource languages is significantly lower (\cite{DBLP:journals/corr/abs-2311-07978, DBLP:conf/africanlp/OjoO23}). This disparity is due to the fact that the quality and quantity of the data directly impact the performance of the models (\cite{DBLP:conf/aaai/Liu0YDJCMF21}). For the large majority of the approximately more than 7,000 languages spoken worldwide, this data is scarce or non-existent (\cite{joshi-etal-2020-state}). Therefore, obtaining optimal results would require manually generating annotated data for each application domain and language. Given the rapidly increasing number of tasks and domains to which NLP is applied, this is an unfeasible task in terms of monetary cost and human effort.

\begin{figure}[htb]
    \centering
    \includegraphics[width=0.7\textwidth]{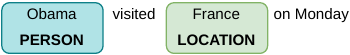}
    \caption{Illustration of the Named Entity Recognition (NER) sequence labelling task. The goal is to identify and classify named entities in running text.}
    \label{fig1:Ner}
\end{figure}

The primary objective of this thesis is to develop cross-lingual transfer learning solutions to address the resource constraints faced by many languages, tasks, and domains. \emph{Cross-lingual transfer learning} is a research area focused on creating models for low-resource languages by leveraging knowledge from high-resource languages (\cite{DBLP:conf/nips/ConneauL19}). Cross-lingual transfer learning uses the data and models available in high-resource languages (typically English) to solve tasks in low-resource languages where these resources are scarce or non-existent.

This thesis explores cross-lingual transfer learning for sequence labeling tasks. \emph{Sequence labeling} is the task of assigning a label to each token in a given input sequence (\cite{DBLP:conf/icml/LaffertyMP01}). Figure \ref{fig1:Ner} illustrates the Named Entity Recognition (NER) sequence labeling task, where the goal is to identify and classify named entities in a text. Sequence labeling tasks are essential for many NLP applications, such as Information Extraction, Question Answering, and Sentiment Analysis, among others. By applying cross-lingual transfer learning techniques, such as translation and annotation projection, alongside multilingual NLP models, we aim to leverage resources from high-resource languages to perform sequence labeling in low-resource languages. Our final goal is to develop publicly available models that achieve state-of-the-art performance in low-resource languages.

\section{Goals and research lines}
\label{goals}

The main goal of this thesis is to develop state-of-the-art cross-lingual transfer learning methods for sequence labeling tasks. We aim to apply these methods to real-world problems where the lack of resources is a significant issue. Additionally, we intend to provide the research community with a set of tools, as well as generate freely available data and models that can be used in the future. The research lines of this thesis are as follows:

\begin{itemize}
    \item \textbf{RL1: Develop better data-based cross-lingual transfer learning methods for sequence labeling tasks.} Data-transfer methods focus on transferring knowledge from high-resource to low-resource languages through translation and annotation projection. At the start of this thesis, most data-based approaches relied on statistical word alignment methods and sub-optimal Machine Translation models. Our goal was to develop improved data-based methods that leverage the latest advances in Machine Translation and NLP models. We also aim to explore the use of multilingual NLP models for data transfer, which have shown promising results in other NLP tasks.
    
    \item \textbf{RL2: Develop better model-based cross-lingual transfer learning methods for sequence labeling tasks.} Model-transfer methods are based on transferring knowledge from high-resource to low-resource languages through pre-trained models. A multilingual NLP model is fine-tuned on data from high-resource languages and then directly applied to low-resource languages. At the start of this thesis, this approach was offering good results in many NLP tasks using encoder-only models. Our objective is to develop improved model-based methods by leveraging the multilingual capabilities of state-of-the-art text-to-text pre-trained models.
    
    \item \textbf{RL3: Real-world application of cross-lingual transfer learning methods.} We aim to apply the developed methods to real-world problems where the lack of resources is a significant issue. By doing so, we aim to better understand which scenarios are best suited for different techniques in cross-lingual transfer learning. Additionally, we develop open-source tools, datasets, and models to support the research community in replicating our experiments and extending our work. These resources are intended to facilitate advancements in NLP for low-resource languages and enable their application across diverse tasks, languages, and domains.

\end{itemize}
\clearpage
\section{Structure of the thesis}

This thesis is structured as a series of interconnected papers, each building on the previous one. The chapters are organized as follows:

In Chapter \ref{ch:related-work}, we present the background of the thesis, review the state-of-the-art in cross-lingual transfer learning for sequence labeling tasks, and introduce the main concepts and techniques used in this research.

Chapter \ref{ch:model-vs-data} focuses on the effectiveness of model-based and data-based cross-lingual transfer learning methods for sequence labeling tasks. We identify the advantages and shortcomings of each method, as well as the challenges faced by current techniques for cross-lingual zero-resource sequence labeling. These insights provide a foundation for the subsequent chapters.

In Chapter \ref{ch:data-transfer}, we introduce a novel data-based method for cross-lingual transfer learning in zero-resource settings. We propose T-Projection, a method that achieves state-of-the-art performance on annotation projection tasks.

Chapter \ref{ch:model-transfer} presents a constrained decoding algorithm that improves the performance of the model-based cross-lingual transfer learning approach. We demonstrate that the constrained decoding algorithm successfully leverages text-to-text models for sequence labeling tasks in low-resource languages achieving state-of-the-art results. 

Chapter \ref{ch:medicalmt5} offers a case study on the application of cross-lingual transfer learning to the medical domain. We show that the methods developed in this thesis can be successfully applied to real-world problems where resource scarcity is a significant issue. By applying both data-based and model-based methods, we develop a comprehensive multilingual pre-training, fine-tuning, and evaluation framework for the medical domain, culminating in the first open-source text-to-text multilingual model for the medical domain.

Finally, Chapter \ref{ch:final-chapter} summarizes the conclusions of the thesis, discusses the main contributions and limitations of the work, and proposes future research directions.

\clearpage

\section{List of scientific contributions}

In this section, we present the scientific contributions developed throughout this thesis. This section is divided into three parts. First, we present the publications that are included in this manuscript. Next, we provide a list of publications closely related to the thesis topic but not included in this manuscript. Finally, we list publications from other lines of research that are outside the scope of this thesis. All papers are listed in chronological order.
\clearpage
\subsection{Contributions included in the thesis}
\label{publi}

These three publications are included in this thesis manuscript, as they present the main contributions of the thesis. Their content will be explained in the following chapters.

\begin{part_of_the_thesis}{Model and Data Transfer for Cross-Lingual Sequence Labelling in Zero-Resource Settings.}
    \begin{smallbox}
     Presented in Chapter \ref{ch:model-vs-data}.
    \end{smallbox}
    \underline{Iker García-Ferrero}, Rodrigo Agerri, and German Rigau. \\
    Findings of the Association for Computational Linguistics: EMNLP 2022. \\
    \textit{\href{https://doi.org/10.18653/v1/2022.findings-emnlp.478}{https://doi.org/10.18653/v1/2022.findings-emnlp.478}}
    \end{part_of_the_thesis}

\begin{part_of_the_thesis}{T-projection: High quality annotation projection for sequence labeling tasks}
        \begin{smallbox}
         Presented in Chapter \ref{ch:data-transfer}.
        \end{smallbox}
        \underline{Iker García-Ferrero}, Rodrigo Agerri, and German Rigau. \\
        Findings of the Association for Computational Linguistics: EMNLP 2023 \\
        \textit{\href{https://doi.org/10.18653/v1/2023.findings-emnlp.1015}{https://doi.org/10.18653/v1/2023.findings-emnlp.1015}}
        \end{part_of_the_thesis}

\begin{part_of_the_thesis}{Medical mT5: An Open-Source Multilingual Text-to-Text LLM for The Medical Domain}
    \begin{smallbox}
        Presented in Chapter \ref{ch:medicalmt5}.
    \end{smallbox}
    \underline{Iker García-Ferrero}, Rodrigo Agerri, Aitziber Atutxa Salazar, Elena Cabrio, Iker de la Iglesia, Alberto Lavelli, Bernardo Magnini, Benjamin Molinet, Johana Ramirez-Romero, German Rigau, Jose Maria Villa-Gonzalez, Serena Villata, Andrea Zaninello. \\
    LREC-COLING 2024 \\
    \textit{\href{https://aclanthology.org/2024.lrec-main.974}{https://aclanthology.org/2024.lrec-main.974}}
    \end{part_of_the_thesis}

\clearpage

\subsection{Closely Related Contributions}
These contributions are not included in this manuscript, as they are not directly related to the cross-lingual transfer paradigm. However, they explore complementary topics aligned with the thesis’s main research direction, sharing the objective of advancing Information Extraction systems.

\begin{not_part_of_the_thesis}{Benchmarking meta-embeddings: What works and what does not}
    \underline{Iker García-Ferrero}, Rodrigo Agerri, and German Rigau. \\
    Findings of the Association for Computational Linguistics: EMNLP 2021 \\
    \textit{\href{https://doi.org/10.18653/v1/2021.findings-emnlp.333}{https://doi.org/10.18653/v1/2021.findings-emnlp.333}}
\end{not_part_of_the_thesis}

\begin{not_part_of_the_thesis}{	
    IXA/Cogcomp at SemEval-2023 Task 2: Context-enriched Multilingual Named Entity Recognition using Knowledge Bases}
    \underline{Iker García-Ferrero}, Jon Ander Campos, Oscar Sainz, Ander Salaberria, and Dan Roth. \\
    Proceedings of the 17th International Workshop on Semantic Evaluation (SemEval-2023) \\
    \textit{\href{https://doi.org/10.18653/v1/2023.semeval-1.186}{https://doi.org/10.18653/v1/2023.semeval-1.186}}
\end{not_part_of_the_thesis}

\begin{not_part_of_the_thesis}{	
    \scalerel*{\includegraphics{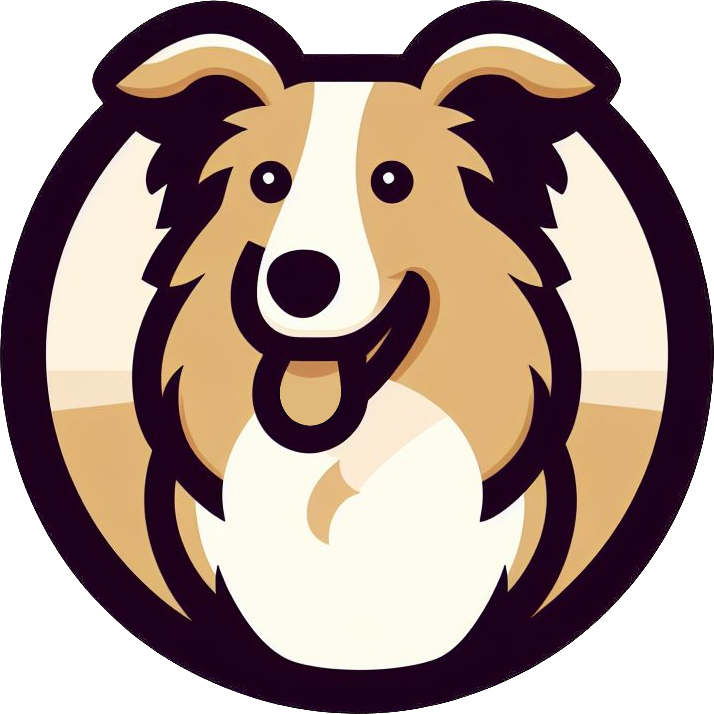}}{\textrm{\textbigcircle}} GoLLIE: Annotation Guidelines improve Zero-Shot Information-Extraction}
    Oscar Sainz, \underline{Iker García-Ferrero}, Rodrigo Agerri, Oier Lopez de Lacalle, German Rigau, Eneko Agirre \\
    The Twelfth International Conference on Learning Representations, 2024 \\
    \textit{\href{https://openreview.net/forum?id=Y3wpuxd7u9}{https://openreview.net/forum?id=Y3wpuxd7u9}}
\end{not_part_of_the_thesis}

\clearpage
 
\subsection{Contributions that are not part of the Thesis}
These contributions result from collaborations with other members of the IXA group and the research community. They focus on evaluating language models, particularly regarding Bias and Data Contamination. Although not directly aligned with the main goals of this thesis, their insights have direct implications for the work conducted in the thesis. 

\begin{not_part_of_the_thesis}{Itzulpen Automatikoko Sistemen Analisia: Genero Alborapenaren Kasua.}
    Ander Salaberria, Jon Ander Campos, \underline{Iker García-Ferrero}, Joseba Fernandez de Landa \\
    In Proceedings of the IV. Ikergazte (2021). Nazioarteko ikerketa euskaraz. Kongresuko artikulu bilduma. Ingeniaritza eta Arkitektura. \\
    \textit{\href{http://ixa.si.ehu.es/node/13328}{http://ixa.si.ehu.es/node/13328}}
\end{not_part_of_the_thesis}

\begin{not_part_of_the_thesis}{Twitterreko Euskal Komunitatearen Eduki Azterketa Pandemia Garaian.}
    Joseba Fernandez de Landa, \underline{Iker García-Ferrero}, Ander Salaberria, Jon Ander Campos \\
    In Proceedings of the IV. Ikergazte (2021). Nazioarteko ikerketa euskaraz. Kongresuko artikulu bilduma. Ingeniaritza eta Arkitektura. \\
    \textit{\href{http://ixa.si.ehu.es/node/13327}{http://ixa.si.ehu.es/node/13327}}
\end{not_part_of_the_thesis}

\begin{not_part_of_the_thesis}{	
    This is not a Dataset: A Large Negation Benchmark to Challenge Large Language Models}
    \underline{Iker García-Ferrero}, Begoña Altuna, Javier Álvez, Itziar Gonzalez-Dios, German Rigau. \\
    Proceedings of the 2023 Conference on Empirical Methods in Natural Language Processing \\
    \textit{\href{https://doi.org/10.18653/v1/2023.emnlp-main.531}{https://doi.org/10.18653/v1/2023.emnlp-main.531}}
\end{not_part_of_the_thesis}

\clearpage

\begin{not_part_of_the_thesis}{	
    NLP Evaluation in trouble: On the Need to Measure LLM Data Contamination for each Benchmark}
    Oscar Sainz, Jon Ander Campos, \underline{Iker García-Ferrero}, Julen Etxaniz, Oier Lopez de Lacalle, Eneko Agirre \\
    Findings of the Association for Computational Linguistics: EMNLP 2023 \\
    \textit{\href{https://doi.org/10.18653/v1/2023.findings-emnlp.722}{https://doi.org/10.18653/v1/2023.findings-emnlp.722}}
\end{not_part_of_the_thesis}

\begin{not_part_of_the_thesis}{	
    Uncovering Social Changes of the Basque Speaking Twitter Community During COVID-19 Pandemic}
    Joseba Fernandez de Landa, \underline{Iker García-Ferrero}, Ander Salaberria, Jon Ander Campos\\
    Proceedings of the 3rd Annual Meeting of the Special Interest Group on Under-resourced Languages @ LREC-COLING 2024 \\
    \textit{\href{https://aclanthology.org/2024.sigul-1.44}{https://aclanthology.org/2024.sigul-1.44}}
\end{not_part_of_the_thesis}

\begin{not_part_of_the_thesis}{	
    \scalerel*{\includegraphics{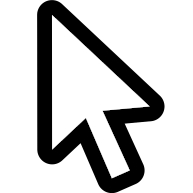}}{\textrm{\textbigcircle}}NoticIA: A Clickbait Article Summarization Dataset in Spanish}
    \underline{Iker García-Ferrero}, Begoña Altuna \\
    Journal Procesamiento del Lenguaje Natural, 2024 \\
    \textit{\href{http://journal.sepln.org/sepln/ojs/ojs/index.php/pln/article/view/6610}{http://journal.sepln.org/sepln/ojs/ojs/index.php/pln/article/view/6610}}
\end{not_part_of_the_thesis}

\begin{not_part_of_the_thesis}{	
    Data Contamination Report from the 2024 CONDA Shared Task}
    Oscar Sainz, \underline{Iker García-Ferrero}, Alon Jacovi, Jon Ander Campos, Yanai Elazar, Eneko Agirre, Yoav Goldberg, Wei-Lin Chen, Jenny Chim, Leshem Choshen, Luca D'Amico-Wong, Melissa Dell, Run-Ze Fan, Shahriar Golchin, Yucheng Li, Pengfei Liu, Bhavish Pahwa, Ameya Prabhu, Suryansh Sharma, Emily Silcock, Kateryna Solonko, David Stap, Mihai Surdeanu, Yu-Min Tseng, Vishaal Udandarao, Zengzhi Wang, Ruijie Xu, Jinglin Yang \\
    Proceedings of The 1st Workshop on Data Contamination (CONDA) @ ACL 2024  \\
    \textit{\href{https://aclanthology.org/2024.conda-1.4/}{https://aclanthology.org/2024.conda-1.4/}}
\end{not_part_of_the_thesis}

\clearpage

\section{List of open-source resources}

As mentioned above, this thesis emphasizes reproducibility and the development of tools and resources freely available for the research community. We have developed open-source software, datasets, and models that other researchers can use to replicate our experiments and build upon our work. The following is a list of open-source resources developed during the years of the thesis:

\subsection{Open source software}

\begin{resources}
    {
    \begin{adjustbox}{valign=c}
    \includegraphics[width=0.05\textwidth]{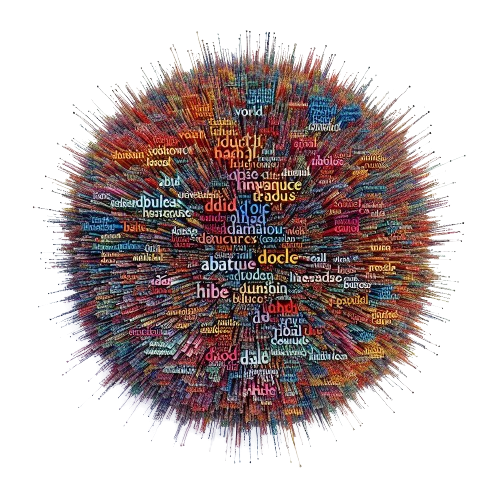}
    \end{adjustbox}
    \begin{adjustbox}{valign=c}
    MetaVec
    \end{adjustbox}
    }
    A monolingual and cross-lingual meta-embedding generation and evaluation framework. \\
    \textit{\href{https://github.com/ikergarcia1996/MetaVec}{https://github.com/ikergarcia1996/MetaVec}}
\end{resources}

\begin{resources}
    {
    \begin{adjustbox}{valign=c}
    \includegraphics[width=0.04\textwidth]{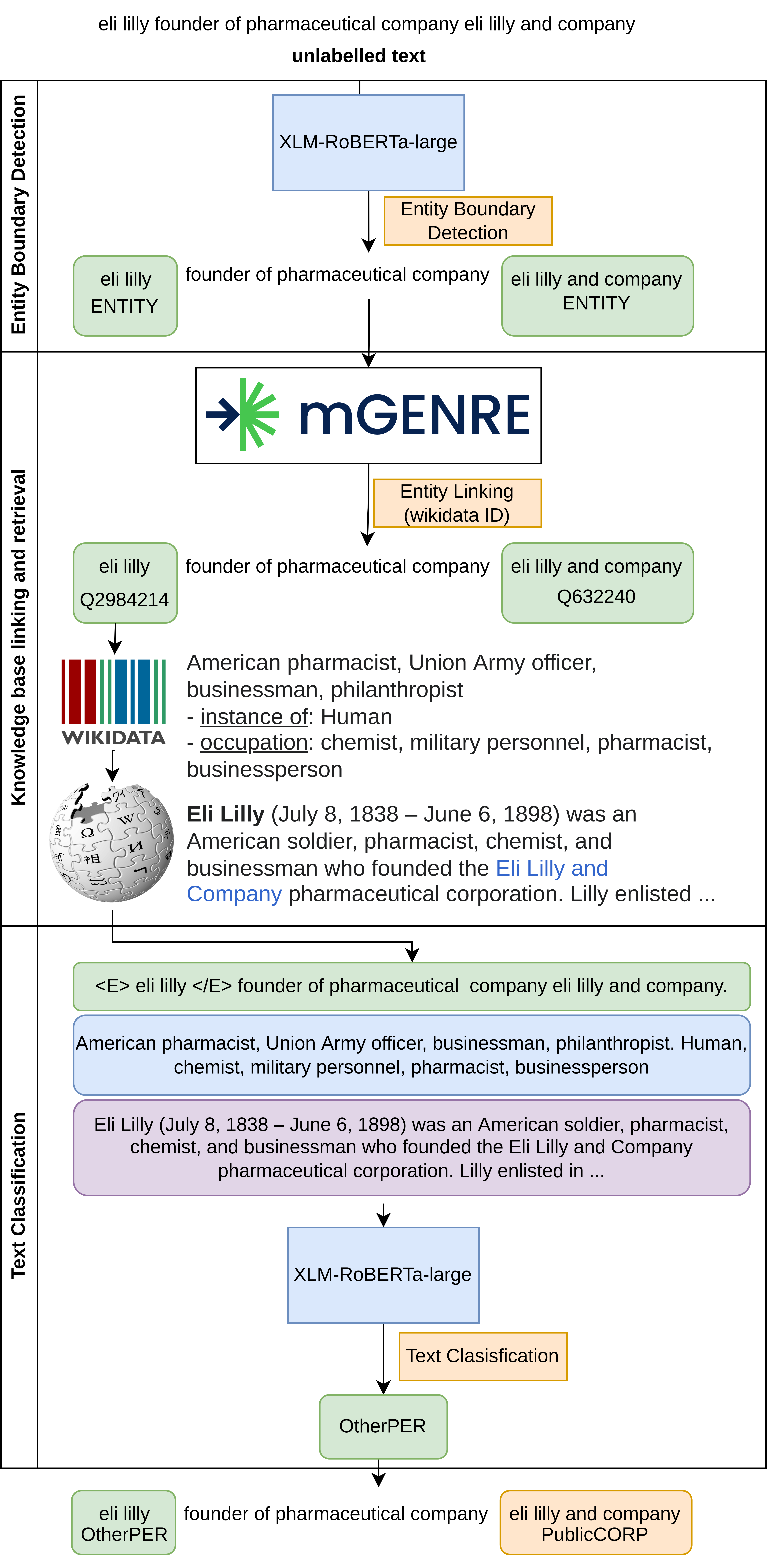}
    \end{adjustbox}
    \begin{adjustbox}{valign=c}
    \begin{minipage}{\textwidth}
    Context-enriched multilingual named entity recognition \\ using knowledge bases.
    \end{minipage}
    \end{adjustbox}
    }
    A NER frameworks that (1) identifies possible entity candidates by analyzing the input sentence structure, (2) links the candidate to an existing updated knowledge base if possible, and (3) performs the fine-grained classification using the input sentence plus the retrieved information from the KB about the entity. \\
    \textit{\href{https://github.com/ikergarcia1996/Context-enriched-NER}{https://github.com/ikergarcia1996/Context-enriched-NER}}
\end{resources}

\begin{resources}
    {
    \includegraphics[height=0.8cm]{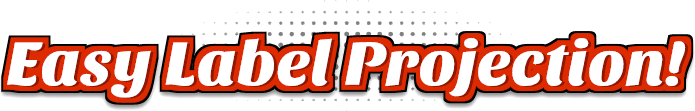}
    }
    \begin{resourcessmall}
    Developed in Chapter \ref{ch:model-vs-data}
    \end{resourcessmall}
    Easy Label Projection is a library that allows to project labels from one dataset into another easily. You can automatically generate datasets for languages for which you do not have any labelled data using mGiza, FastAlign, SimALign or AWESOME. \\
    \textit{\href{https://github.com/ikergarcia1996/Easy-Label-Projection}{https://github.com/ikergarcia1996/Easy-Label-Projection}}
\end{resources}

\begin{resources}
    {
    \includegraphics[height=0.6cm]{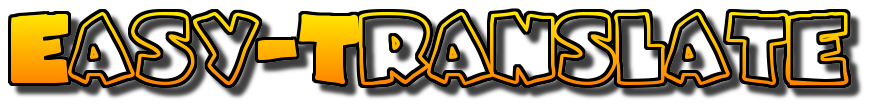}
    }
    \begin{resourcessmall}
    Developed in Chapter \ref{ch:model-vs-data}
    \end{resourcessmall}
    Easy-Translate is a script for translating large text files with a SINGLE COMMAND. Easy-Translate is designed to be as easy as possible for beginners and as seamless customizable and as possible for advanced users. \\
    \textit{\href{https://github.com/ikergarcia1996/Easy-Translate}{https://github.com/ikergarcia1996/Easy-Translate}}
\end{resources}

\begin{resources}
    {
    \begin{adjustbox}{valign=c}
    \includegraphics[width=0.06\textwidth]{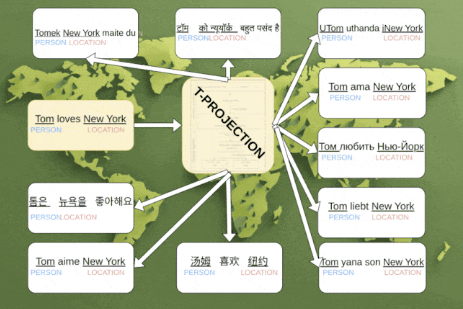}
    \end{adjustbox}
    \begin{adjustbox}{valign=c}
    \begin{minipage}{\textwidth}
    T-Projection
    \end{minipage}
    \end{adjustbox}
    }
    \begin{resourcessmall}
    Developed in Chapter \ref{ch:data-transfer}
    \end{resourcessmall}
    T-Projection is a method to perform high-quality Annotation Projection of Sequence Labeling datasets. The code is built on top of HuggingFace's Transformers and HuggingFace's Accelerate library. \\
    \textit{\href{https://github.com/ikergarcia1996/T-Projection}{https://github.com/ikergarcia1996/T-Projection}}
\end{resources}

\begin{resources}
    {
    \begin{adjustbox}{valign=c}
    \includegraphics[width=0.08\textwidth]{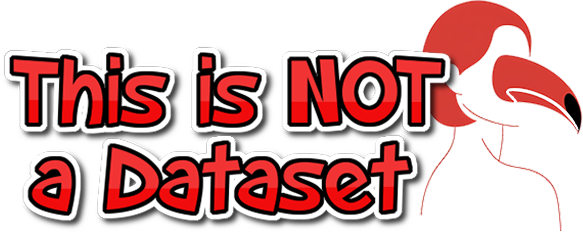}
    \end{adjustbox}
    \begin{adjustbox}{valign=c}
    \begin{minipage}{\textwidth}
    TINAD Framework
    \end{minipage}
    \end{adjustbox}
    }
    A LLM finetuning and LLM evaluation library for the "This Is NOT a dataset" (TINAD) dataset.  \\
    \textit{\href{https://github.com/hitz-zentroa/This-is-not-a-Dataset}{https://github.com/hitz-zentroa/This-is-not-a-Dataset}}
\end{resources}

\begin{resources}
    {
    \begin{adjustbox}{valign=c}
    \includegraphics[width=0.04\textwidth]{Capitulos/1_Introduccion/logos/GoLLIE.png}
    \end{adjustbox}
    \begin{adjustbox}{valign=c}
    \begin{minipage}{\textwidth}
    GoLLIE Framework
    \end{minipage}
    \end{adjustbox}
    }
    The framework to finetune and evaluate GoLLIE-style models. Allows to implement any IE task by defining a set of categories and guidelines. Co-developed with Oscar Sainz. \\
    \textit{\href{https://github.com/hitz-zentroa/GoLLIE}{https://github.com/hitz-zentroa/GoLLIE}}
\end{resources}

\begin{resources}
    {
    \begin{adjustbox}{valign=c}
    \includegraphics[width=0.04\textwidth]{Capitulos/1_Introduccion/logos/noticia.png}
    \end{adjustbox}
    \begin{adjustbox}{valign=c}
    \begin{minipage}{\textwidth}
    NoticIA Framework
    \end{minipage}
    \end{adjustbox}
    }
    A LLM finetuning and LLM evaluation library for the NoticIA dataset.  \\
    \textit{\href{https://github.com/ikergarcia1996/NoticIA}{https://github.com/ikergarcia1996/NoticIA}}
\end{resources}

\subsection{Open source datasets}

\begin{resources}
    {
    \begin{adjustbox}{valign=c}
    \includegraphics[width=0.08\textwidth]{Capitulos/1_Introduccion/logos/tinad.png}
    \end{adjustbox}
    \begin{adjustbox}{valign=c}
    \begin{minipage}{\textwidth}
    This is NOT a Dataset
    \end{minipage}
    \end{adjustbox}
    }
    A large semi-automatically generated dataset of ~400,000 descriptive sentences about commonsense knowledge that can be true or false in which negation is present in about 2/3 of the corpus in different forms that we use to evaluate LLMs.  \\
    \textit{\href{https://huggingface.co/datasets/HiTZ/This-is-not-a-dataset}{https://huggingface.co/datasets/HiTZ/This-is-not-a-dataset}}
\end{resources}

\begin{resources}
    {
    \begin{adjustbox}{valign=c}
    \includegraphics[width=0.08\textwidth]{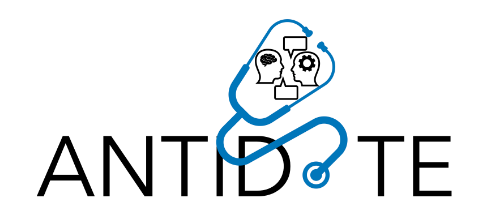}
    \end{adjustbox}
    \begin{adjustbox}{valign=c}
    \begin{minipage}{\textwidth}
    Multilingual Medical Corpus
    \end{minipage}
    \end{adjustbox}
    }
    \begin{resourcessmall}
    Developed in Chapter \ref{ch:medicalmt5}
    \end{resourcessmall}
    A Multilingual-Medical-Corpus a 3 billion word multilingual corpus for training LLMs adapted to the medical domain. Multilingual-Medical-Corpus includes four languages, namely, English, Spanish, French, and Italian.  \\
    \textit{\href{https://huggingface.co/datasets/HiTZ/Multilingual-Medical-Corpus}{https://huggingface.co/datasets/HiTZ/Multilingual-Medical-Corpus}}
\end{resources}

\begin{resources}
    {
    \begin{adjustbox}{valign=c}
    \includegraphics[width=0.08\textwidth]{Capitulos/1_Introduccion/logos/antidote.png}
    \end{adjustbox}
    \begin{adjustbox}{valign=c}
    \begin{minipage}{\textwidth}
    Multilingual AbstRCT
    \end{minipage}
    \end{adjustbox}
    }
    \begin{resourcessmall}
    Developed in Chapter \ref{ch:medicalmt5}
    \end{resourcessmall}
    We translate the AbstRCT English Argument Mining Dataset dataset to generate parallel French, Italian and Spanish versions using the NLLB200 3B parameter model and projected using word alignment tools. The projections have been manually corrected.  \\
    \textit{\href{https://huggingface.co/datasets/HiTZ/multilingual-abstrct}{https://huggingface.co/datasets/HiTZ/multilingual-abstrct}}
\end{resources}
\clearpage
\begin{resources}
    {
    \begin{adjustbox}{valign=c}
    \includegraphics[width=0.08\textwidth]{Capitulos/1_Introduccion/logos/antidote.png}
    \end{adjustbox}
    \begin{adjustbox}{valign=c}
    \begin{minipage}{\textwidth}
    Multilingual BioASQ-6B
    \end{minipage}
    \end{adjustbox}
    }
    \begin{resourcessmall}
    Developed in Chapter \ref{ch:medicalmt5}
    \end{resourcessmall}
    We translate the BioASQ-6B English Question Answering dataset to generate parallel French, Italian and Spanish versions using the NLLB200 3B parameter model.  \\
    \textit{\href{https://huggingface.co/datasets/HiTZ/Multilingual-BioASQ-6B}{https://huggingface.co/datasets/HiTZ/Multilingual-BioASQ-6B}}
\end{resources}

\begin{resources}
    {
    \begin{adjustbox}{valign=c}
    \includegraphics[width=0.04\textwidth]{Capitulos/1_Introduccion/logos/noticia.png}
    \end{adjustbox}
    \begin{adjustbox}{valign=c}
    \begin{minipage}{\textwidth}
    NoticIA
    \end{minipage}
    \end{adjustbox}
    }
    A dataset consisting of 850 Spanish news articles featuring prominent clickbait headlines, each paired with high-quality, single-sentence generative summarizations written by humans.  \\
    \textit{\href{https://huggingface.co/datasets/Iker/NoticIA}{https://huggingface.co/datasets/Iker/NoticIA}}
\end{resources}

\subsection{Open source models}

\begin{resources}
    {
    \begin{adjustbox}{valign=c}
    \includegraphics[width=0.08\textwidth]{Capitulos/1_Introduccion/logos/antidote.png}
    \end{adjustbox}
    \begin{adjustbox}{valign=c}
    \begin{minipage}{\textwidth}
    medical mT5
    \end{minipage}
    \end{adjustbox}
    }
    \begin{resourcessmall}
    Developed in Chapter \ref{ch:medicalmt5}
    \end{resourcessmall}
    The first open-source text-to-text multilingual model for the medical domain. Medical mT5 is an encoder-decoder model developed by continuing the training of publicly available mT5 checkpoints on medical domain data for English, Spanish, French, and Italian.  \\
    \textit{\href{https://huggingface.co/HiTZ/Medical-mT5-xl}{https://huggingface.co/HiTZ/Medical-mT5-xl}}
\end{resources}
\clearpage
\begin{resources}
    {
    \begin{adjustbox}{valign=c}
    \includegraphics[width=0.04\textwidth]{Capitulos/1_Introduccion/logos/GoLLIE.png}
    \end{adjustbox}
    \begin{adjustbox}{valign=c}
    \begin{minipage}{\textwidth}
    GoLLIE
    \end{minipage}
    \end{adjustbox}
    }
    A Large Language Model trained to follow annotation guidelines. GoLLIE outperforms previous approaches on zero-shot Information Extraction and allows the user to perform inferences with annotation schemas defined on the fly. Unlike previous approaches, GoLLIE can follow detailed definitions and not only rely on the knowledge already encoded in the LLM. \\
    \textit{\href{https://huggingface.co/HiTZ/GoLLIE-34B}{https://huggingface.co/HiTZ/GoLLIE-34B}}
\end{resources}

\begin{resources}
    {
    \begin{adjustbox}{valign=c}
    \includegraphics[width=0.05\textwidth]{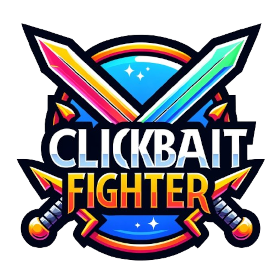}
    \end{adjustbox}
    \begin{adjustbox}{valign=c}
    \begin{minipage}{\textwidth}
    ClickbaitFighter
    \end{minipage}
    \end{adjustbox}
    }
    A model finetuned with the NoticIA Dataset. This model can generate summaries of clickbait headlines. \\
    \textit{\href{https://huggingface.co/Iker/ClickbaitFighter-10B}{https://huggingface.co/Iker/ClickbaitFighter-10B}}
\end{resources}
\selectlanguage{english}

\chapter[Related Work]{Related Work}
\label{ch:related-work}

In this chapter, we will present the state-of-the-art in Natural Language Processing (NLP) and cross-lingual transfer. We will begin with a brief overview of recent advancements in NLP and multilingual language models. Next, we will discuss the different approaches for cross-lingual transfer learning. More specifically, we will present the two main paradigms for cross-lingual transfer: data-based transfer and model-based transfer. For each paradigm, we will describe the various methods proposed in previous research. These methods will serve as the baselines for the experiments presented in the following chapters.

\section{NLP and Deep Learning: Scaling compute and data}
\label{sc:deep-learning-sota}

In the last few years, the Natural Language Processing paradigm has switched from building pipelines of different processing modules (\cite{DBLP:conf/lrec/AgerriBR14}) into end-to-end neural networks trained with vast amounts of text data (\cite{DBLP:journals/corr/abs-2111-01243}).

\begin{figure}[ht]
    \centering
    \includegraphics[height=0.65\linewidth]{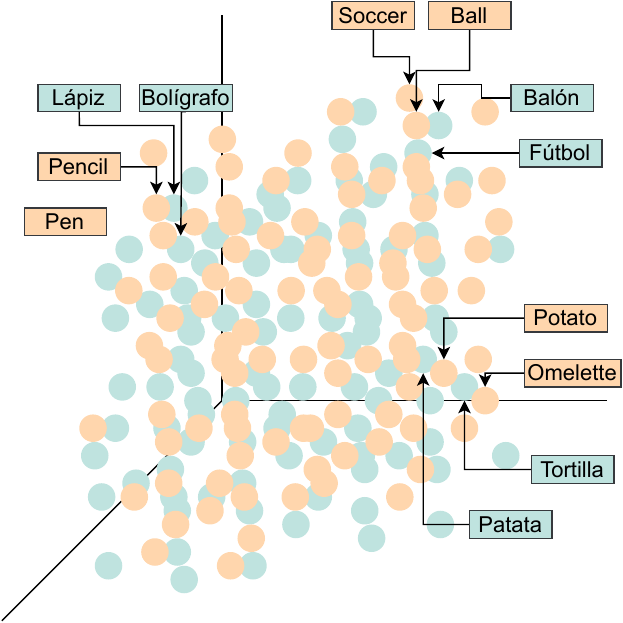}
    \caption{Illustration of multilingual embeddings, where two languages are mapped into a shared vector space. Words with similar meanings are placed close together.}
    \label{fig:chap3_multilingual_embeddings}
\end{figure}

The first step in this direction was the introduction of word embeddings, such as Word2Vec (\cite{DBLP:journals/corr/abs-1301-3781}), GloVe (\cite{pennington-etal-2014-glove}), and FastText (\cite{DBLP:journals/tacl/BojanowskiGJM17}). These embeddings are trained on large corpora of text and capture the semantic and syntactic properties of words. Word embeddings are used as input features for neural networks that perform a wide range of NLP tasks. Multilinguality was added to this paradigm with the introduction of multilingual word embeddings (\cite{DBLP:journals/jair/RuderVS19}). As depicted in Figure \ref{fig:chap3_multilingual_embeddings}, multilingual word embeddings are trained on text from multiple languages and map words from different languages into a shared vector space (\cite{DBLP:conf/icml/GouwsBC15,DBLP:conf/naacl/LuongPM15}). Alternatively, multilingual word embeddings can be trained on monolingual data and then projected into a shared space using a bilingual dictionary (\cite{zhang-etal-2016-ten,artetxe-etal-2016-learning,DBLP:conf/iclr/SmithTHH17}). This shared space allows for the transfer of knowledge across languages, enabling the training of models in one language and applying them to another. However, the cross-lingual transfer capabilities of word embeddings-based systems are limited and often perform poorly on low-resource languages (\cite{conneau-etal-2018-xnli}).

The introduction of the Transformer architecture (\cite{DBLP:conf/nips/VaswaniSPUJGKP17}) marked a significant shift in the NLP field. Transformers have achieved state-of-the-art performance on a wide range of NLP tasks, such as Machine Translation (\cite{DBLP:conf/nips/VaswaniSPUJGKP17}), Text Classification (\cite{devlin-etal-2019-bert}), General Language Understanding (\cite{DBLP:conf/iclr/WangSMHLB19}), Question Answering (\cite{DBLP:conf/acl/RajpurkarJL18}), Text Generation (\cite{brown2020language}) or Dialogue (\cite{DBLP:journals/corr/abs-2201-08239}) among many others (\cite{DBLP:journals/corr/abs-2111-01243}).  The success of the Transformer architecture has led to the development of a broad range of Transformer-based language models.

\begin{figure}[ht]
    \centering
    \includegraphics[width=0.85\linewidth]{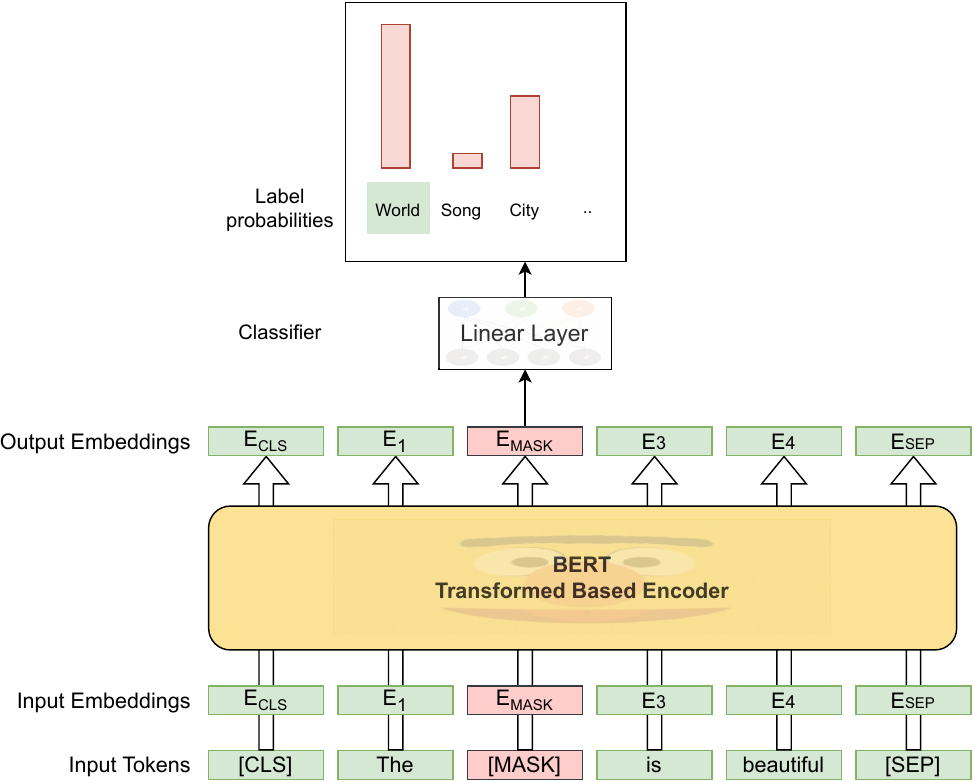}
    \caption{Representation of the BERT architecture. During training, BERT learns to predict missing words in a sentence based on the contextual representations produced by the model.}
    \label{fig:chap3_bert}
\end{figure}

The first prominent Transformer-based language model was BERT (\cite{devlin-etal-2019-bert}). BERT is trained on large corpora of text data and learns to predict missing words in a sentence, producing contextual embeddings that capture the meaning of words in context. This process is illustrated in Figure \ref{fig:chap3_bert}. Similar to word embeddings, BERT embeddings can be used as input features for a wide range of NLP tasks. BERT has been extended to support multiple languages with mBERT (\cite{devlin-etal-2019-bert}), which is trained on text from over 100 languages. mBERT achieved state-of-the-art performance in multiple languages and demonstrated strong performance when trained in English and applied to other languages (\cite{pires-etal-2019-multilingual,artetxe-schwenk-2019-massively}). The success of mBERT has led to the development of other multilingual models, such as XLM-RoBERTa (\cite{conneau-etal-2020-unsupervised}) and DeBERTa (\cite{DBLP:conf/iclr/HeLGC21/deberta}). These models have increasingly larger sizes and are trained on progressively larger corpora of text data.

\begin{figure}[ht]
    \centering
    \includegraphics[width=0.95\linewidth]{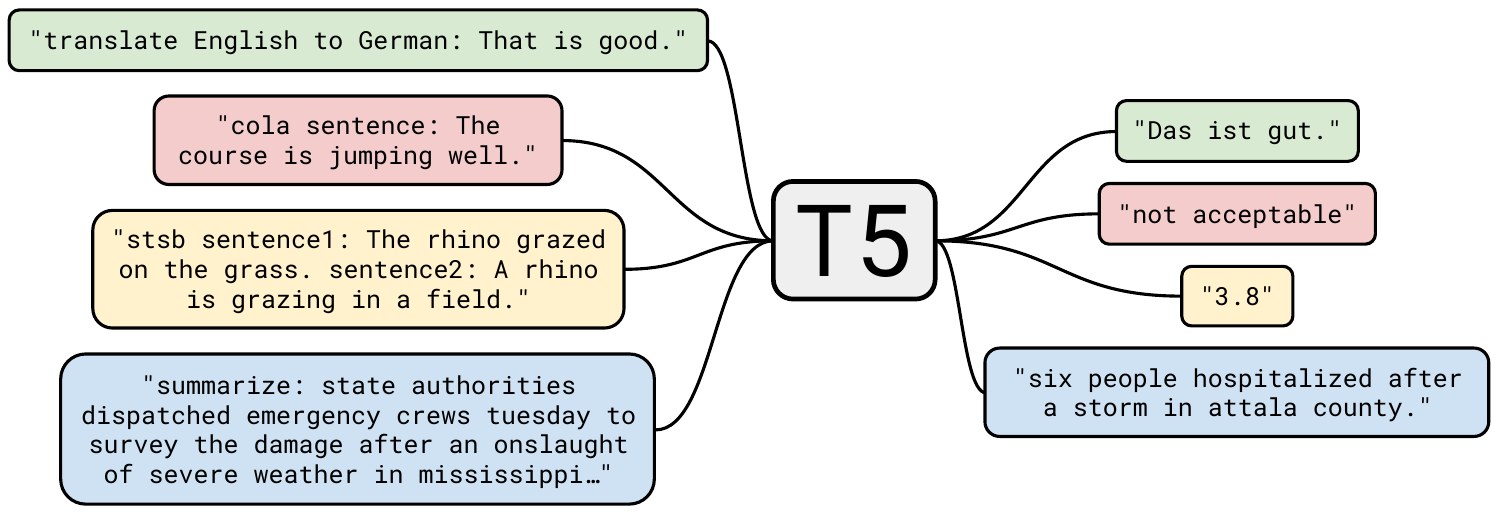}
    \caption{Representation of the text-to-text framework in T5. Every task is framed as a text input and the model is trained to generate the desired output as text. Figure reproduced from \cite{DBLP:journals/jmlr/RaffelSRLNMZLL20-T5}.}
    \label{fig:chap3_textotext}
\end{figure}

The introduction of T5 (\cite{DBLP:journals/jmlr/RaffelSRLNMZLL20-T5}) and GPT (\cite{radford2019language,brown2020language}) shifted the focus in NLP from learning word representations to a text-to-text approach. T5 is designed to map input text to output text, enabling it to handle a wide array of NLP tasks. Consequently, all NLP tasks are framed as text-to-text tasks, where the input is a description of the task or a prompt, and the output is the desired result, as illustrated in Figure \ref{fig:chap3_textotext}. Unlike previous NLP models, which were fine-tuned for specific tasks, T5 can be trained on a broad spectrum of tasks with a single training objective (\cite{chung-flan-instruction-models}). 

On a similar research line, \cite{radford2019language} introduced the GPT line of models, demonstrating that Large Language Models (LLMs) trained on extensive internet data can perform, given a natural language task description, tasks such as Question Answering, Machine Translation, and Summarization without explicit supervision. This finding led to the emergence of instruction tuning, also known as multitask fine-tuning, as the leading method for achieving generalization in large models to solve diverse NLP tasks. In this approach, models are first trained on vast amounts of unlabeled data and subsequently fine-tuned on a diverse collection of tasks (\citep{DBLP:conf/emnlp/WangMAKMNADASPK22,chung-flan-instruction-models}) framed as text-to-text problems.

Subsequent research has demonstrated that increasing the parameter count of language models (\citep{brown2020language}), coupled with improvements in the size and quality of the instruction tuning dataset, results in enhanced generalization capabilities. Consequently, models have been increasingly scaled up (\cite{chung-flan-instruction-models}) in both the number of parameters and the amount of training data. This scaling has led to the development of state-of-the-art Large Language Models (LLMs) such as GPT-4 (\cite{openai2024gpt4technicalreport}), LLaMA (\cite{DBLP:journals/corr/abs-2407-21783}), and Mistral (\cite{jiang2023mistral7b}). These models, which have billions of parameters, are trained on hundreds of terabytes of text data. They are also trained on a large number of diverse tasks and instructions, enabling them to perform a wide range of NLP tasks.

In addition to achieving state-of-the-art performance on various NLP tasks, these models can also solve tasks for which they were not explicitly trained (\cite{radford2019language,lieber2021jurassic,DBLP:journals/corr/abs-2201-11990,rae2022scalinglanguagemodelsmethods,DBLP:journals/jmlr/ChowdheryNDBMRBCSGSSTMRBTSPRDHPBAI23}). Since these models are trained on a substantial portion of internet data, they are inevitably multilingual and can be applied to a wide range of languages and tasks.

While the latest generation of NLP models has made a huge step forward in terms of performance, they require a huge amount of data and computational resources to train. This has led to a growing gap (\cite{blasi-etal-2022-systematic}) between high-resource languages, such as English, and low-resource languages, for which there is very little data available (\cite{joshi-etal-2020-state}). A notable example is African languages for which both Open Source (\cite{DBLP:journals/corr/abs-2311-07978}) and Comercial (\cite{DBLP:conf/africanlp/OjoO23}) NLP models produce lower performance for African languages. This has led to the development of \emph{cross-lingual transfer} methods, which aim to leverage the knowledge learned from high-resource languages to improve or enable NLP tasks in low-resource languages. In the following sections, we will present the different approaches for cross-lingual transfer.

\section{Cross-Lingual Transfer Methods}
\label{sc:transfer-methods}

In this section, we will present the different approaches for cross-lingual transfer. More specifically, we will present the data-transfer and model-transfer approaches that will constitute the baselines in the following chapters. 

Cross-lingual transfer in Natural Language Processing (NLP) is a method in which knowledge learned from one language (typically a high-resource language with abundant data and resources) is applied to improve or enable NLP tasks in another language (often a low-resource language with limited or no data). This process can be achieved through various techniques, including translating datasets (\cite{Ehrmann}) (data-transfer) or using multilingual models (\cite{devlin-etal-2019-bert,conneau-etal-2020-unsupervised}) that understand multiple languages (model-transfer). The goal is to overcome the scarcity of annotated data in many languages, thus facilitating multilinguality in NLP applications. 

\subsection{Data-based transfer}

Data transfer leverages parallel data and/or Machine Translation to bridge the gap between languages in cross-lingual NLP tasks. Data transfer methods make the assumption of annotation preservation across translations. In the data transfer paradigm, the NLP model is trained and performs inference in the same language. There are two main approaches for data transfer: Translate-Train and Translate-Test. 

\subsubsection{Translate-Train}

\begin{wrapfigure}{r}{0.55\linewidth}
\vspace{-0.481cm}
    \centering
    \includegraphics[width=\linewidth]{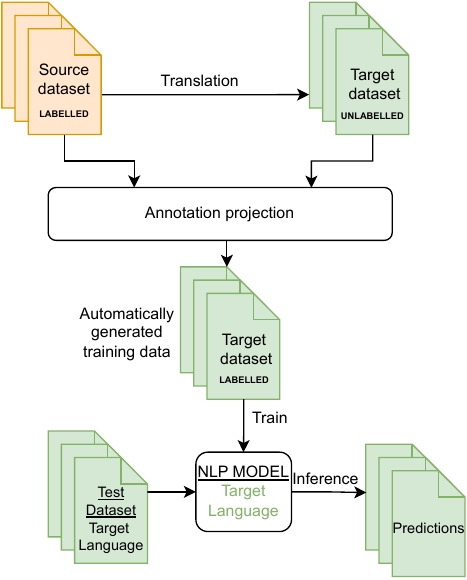}
    \caption{Illustration of the Translate-Train cross-lingual transfer approach: Given gold data in the source language, this method utilizes translation and annotation projection to create silver-standard training data in the target language.}
    \label{fig:chap3_translatetrain}
    \vspace{-0.5cm}
\end{wrapfigure} 

The Translate-Train approach, illustrated in Figure \ref{fig:chap3_translatetrain},  aims to automatically generate annotated data in languages where such data is scarce by leveraging annotated datasets from a high-resource source language. This method begins with a dataset that is fully annotated in a well-resourced source language, which is then translated into the target language (\citet{DBLP:conf/emnlp/JainPL19,fei-etal-2020-cross}). Following translation, the annotations from the source language are projected onto the translated text, resulting in a new, silver-standard annotated dataset in the target language. This newly created dataset can then be used to train NLP models directly on tasks in the target language, making this approach particularly valuable when original annotated data in the target language is not available. High-quality Machine Translation systems are essential for this approach to succeed. An alternative strategy involves automatically annotating the English version of a multi-parallel corpus and then projecting these annotations to all other languages in the corpus (\citet{Ehrmann}). In situations where neither Machine Translation systems nor parallel data are available, \citet{DBLP:conf/acl/GuoR21} translate labeled data on a word-by-word basis using a dictionary and then constructing target-language text from the source-language annotations using a constrained pre-trained language model trained with unlabeled data in the target language. In any case, this approach results in a silver-standard annotated dataset in the target language. The quality of this automatically generated dataset depends on the quality of the translation or parallel data and the effectiveness of the annotation projection algorithm.

 \subsubsection{Translate-Test}
 
\begin{figure}[ht]
    \centering
    \includegraphics[width=10cm]{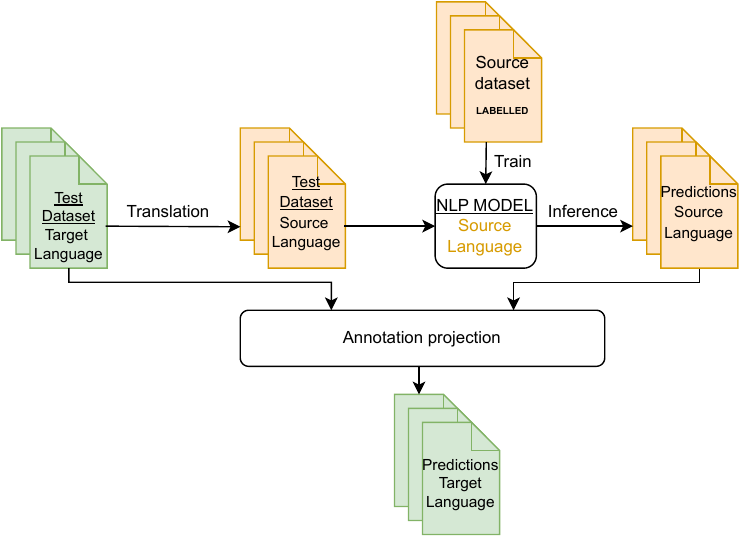}
    \caption{Illustration of the Translate-Test cross-lingual transfer approach: A model is trained using gold data in the source language. During inference, inputs in the target language are first translated into the source language, after which predictions are made and then projected back into the target language.}
    \label{fig:chap3_translatetest}
\end{figure}

Instead of building models for the target language, the translate-test approach aims to take advantage of the ability of the models to produce better results for high-resource languages such as English (\cite{etxaniz-etal-2024-multilingual}). The Translate-Test approach is illustrated in Figure \ref{fig:chap3_translatetest}. In this method (\cite{shah2010synergy,10.1007/978-3-540-45175-4_13,tebbifakhr-etal-2020-machine}), the model is trained in a source language with abundant resources (gold standard data). At inference, the inputs in the target language are first translated into the source language. The model, which is trained on the source language data, performs its inference on these translated inputs. Subsequently, the predictions made by the model are translated or projected back into the target language. This approach enables the deployment of NLP models across languages without retraining the model with annotated data in the target language. However, as with the Translate-Train approach, the performance of the Translate-Test approach relies heavily on the quality of the translation of the input data and the projection of the output predictions into the target language. 

\subsubsection{Annotation Projection}

\begin{figure}
    \centering
    \includegraphics[width=0.9\textwidth]{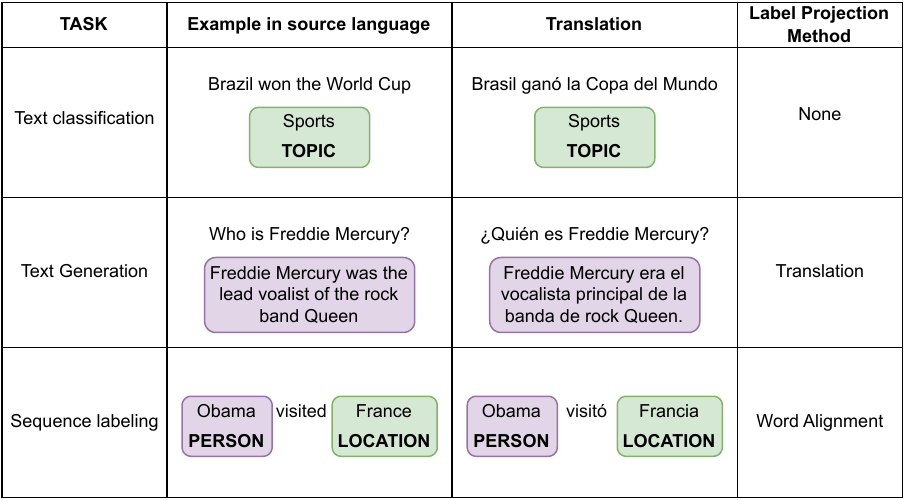}
    \caption{Illustration of data transfer for different NLP tasks. Each task requires a different method to transfer the labels from the source into the target language.}
    \label{fig:chap3_taskexamples}
\end{figure}

In both the Translate-Train and Translate-Test approaches, it is necessary to project the labels from the source language into the target language or vice versa. As depicted in Table \ref{fig:chap3_taskexamples}, the annotation projection method required varies significantly across different NLP tasks. For instance, in Text Classification, annotation projection is straightforward, as the translated sentence will be labeled with the same category as the source sentence. For text generation tasks, such as Abstractive Question Answering, the label can be translated along with the input sentence. The most challenging tasks are those involving Sequence Labeling. For sequence labeling tasks, which involve span-level annotations, it is necessary to identify the sequence of words in the translated sentence that corresponds to the labeled spans in the source text. The majority of previous work published in this research area explores the application of word-alignments (\cite{Ehrmann}).

\subsubsection{Annotation projection using word alignments}
\label{ch:2_word_alignments}

\begin{figure}
    \centering
    \includegraphics[width=8cm]{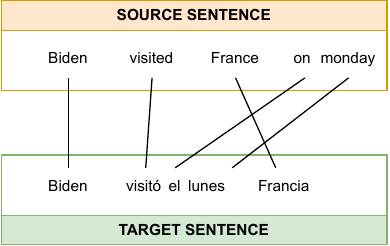}
    \caption{Illustration of word alignments represented as a bidirectional graph.}
    \label{fig:chap3_wordalignmentsraw}
\end{figure}

Word alignments refer to the process of matching words in a sentence in the source language to their corresponding translations in a target language. As illustrated in Figure \ref{fig:chap3_wordalignmentsraw}, word alignments are represented as a bidirectional graph between words in a parallel sentence.

Most word-alignment algorithms are based on statistical Machine Translation systems. Giza++ (\cite{och-ney-2003-systematic-giza}) is based on the IBM Models. The IBM alignment models (\cite{brown-etal-1993-mathematics}) use statistical methods to learn the probability of translation between words in a source language and their counterparts in a target language, based on a given corpus of aligned texts. Building upon this research, FastAlign (\cite{dyer-etal-2013-simple}) introduces a log-linear reparameterization of IBM Model 2, which achieves an accuracy comparable to GIZA++ but with improved computational efficiency, thereby enabling faster inference throughput. On a similar line of research, Efmaral and Eflomal (\cite{Ostling2016efmaral}) extend the IBM models with a Bayesian model with Markov Chain Monte Carlo (MCMC) inference for improved accuracy and computational efficiency.

As neural networks started to outperform previous statistical approaches for most NLP tasks (\cite{DBLP:journals/csur/MinRSVNSAHR24}), a new line of research emerged aiming to generate word alignments using neural networks. This line of research employs multilingual language models that have been pretrained using data from both the source and the target languages. In this line of research, \cite{DBLP:journals/pbml/PeterNN17} built an attention-based neural network in which the attention probabilities are trained to closely align with those obtained from statistical MT toolkits. Building on this work, \cite{DBLP:journals/corr/abs-1901-11359} proposed a method that adds an extra layer of attention on top of the Transformer architecture and directly optimizes its activations towards a given target word.
\cite{garg-etal-2019-jointly} trained a Transformer model to produce both accurate translations and alignments, jointly performing the Machine Translation and Word Alignment tasks. The model outputs the translation during inference while alignments are extracted from the attention probabilities. This approach achieves competitive results compared with GIZA++ without sacrificing translation accuracy. 
However, the model only achieves better performance than GIZA++ when existing word alignments are provided for fine-tuning. 
\cite{DBLP:conf/acl/ZenkelWD20} combines both previous approaches and extends them with a loss function that encourages contiguity in the alignment matrix and a symmetrization algorithm that jointly optimizes the alignment matrix within two models trained in opposite directions. This approach outperforms the alignments generated by GIZA++ and other statistical Word lignment systems.

\begin{figure}
    \centering
    \includegraphics[width=\textwidth]{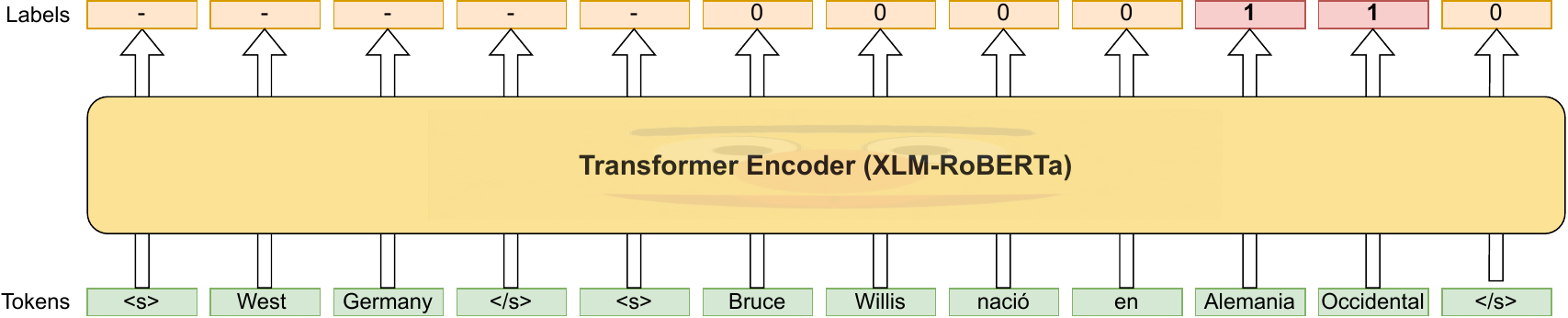}
    \caption{Illustration of word alignments by fine-tuning language models. The query is on the left ``West Germany'' and the translated sentence on the right. The model predicts that ``Alemania Occidental'' is the Spanish translation of ``West Germany''.}
    \label{fig:chap3_wordalignmentsXLMR}
\end{figure}

Instead of extracting alignments from the attention layer of a multilingual language model, \cite{Li2021CrossLingualNE} optimizes a Transformer encoder directly to generate word alignments in its output. As illustrated in Figure \ref{fig:chap3_wordalignmentsXLMR}, the task is formulated as a token classification problem. The alignment model is constructed by concatenating an English text span, representing a labeled sequence, with a sentence in the target language. The model is fine-tuned to predict which tokens in the target sentence correspond to the source text span. They automatically generate a silver fine-tuning dataset using Wikipedia texts in the target language. Anchor text in hyperlinks indicates the location of named entities. The anchored text is machine-translated into the source language. 

\begin{figure}
    \centering
    \includegraphics[width=8cm]{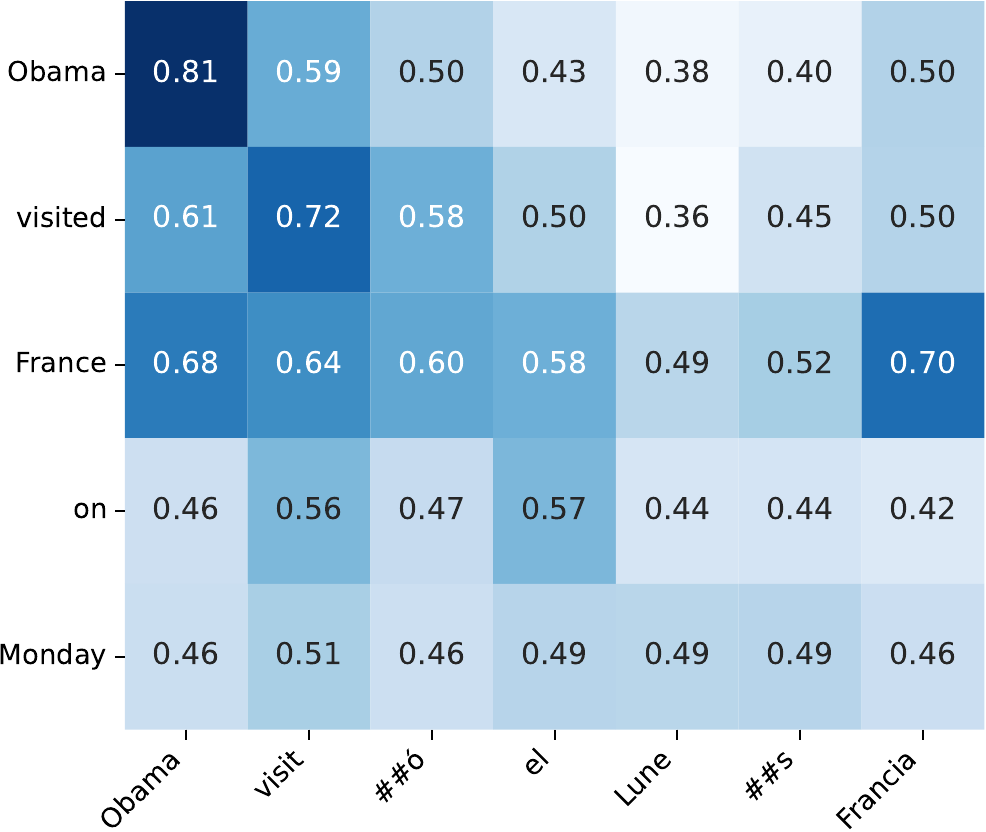}
    \caption{Illustration of the cosine similarity between token embedding representations using Multilingual BERT.}
    \label{fig:chap3_simalign}
\end{figure}

Previous methods that employ neural networks to compute alignments require fine-tuning data, which is very scarce and nonexistent for most language pairs.  SimAlign (\cite{jalili-sabet-etal-2020-simalign}) leverages the contextual embeddings from state-of-the-art multilingual language models such as mBERT (\cite{devlin-etal-2019-bert}). As depicted in Figure \ref{fig:chap3_simalign}, SimAlign identifies alignments between words in parallel sentences based on the similarity of their contextual embeddings. This method allows for more accurate alignments that reflect the contextual use of words within specific sentences. Unlike statistically based word aligners that rely on statistical correlations in large corpora, SimAlign benefits from the deep linguistic and semantic understanding embedded in pre-trained language models. SimAlign offers improvements in alignment quality, especially for languages with complex morphological structures or less parallel data available for training. SimAlign is an unsupervised method, requiring no training data to compute word alignments. Still, SimAlign achieves better alignment accuracy than previous statistical and neural word alignment models. For those language pairs in which parallel data is available, AWESOME (\cite{dou-neubig-2021-word}) improves on this idea by fine-tuning multilingual pre-trained language models on unlabeled parallel text. The main idea is that unsupervised training objectives over the parallel corpus improve the alignment quality of the models.

\begin{figure}
    \centering
    \includegraphics[width=8cm]{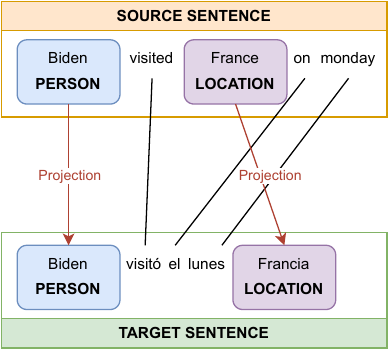}
    \caption{Illustration of annotation projection using word-alignments}
    \label{fig:chap3_wordalignments}
\end{figure}

The word alignments generated by the previously described methods are used for annotation projection. Words in the target sentence are labeled with the same category as the aligned words in the source sentence. This process is depicted in Figure \ref{fig:chap3_wordalignments}. 
Traditionally, most research in this area has focused on automatically annotating the English version of a multi-parallel corpus and then projecting these annotations to all other languages using statistical word alignments, as shown in the works of  \citet{yarowsky-etal-2001-inducing,hwa2005bootstrapping,Ehrmann} and \cite{fu-etal-2011-generating}. \citet{wang-manning-2014-cross} introduces a refinement to this approach by projecting model expectations instead of direct labels, enabling the transfer of model uncertainty across languages and potentially reducing the risk of error propagation. Nevertheless, inaccuracies in word alignment computation remain a significant issue, often resulting in incorrect annotation projections and the generation of noisy data.

To address this problem, \citet{ni-etal-2017-weakly} propose a heuristic scheme for selecting high-quality projection-labeled data from the noisy dataset. This scheme also includes projecting word embeddings from the target language into the source language, allowing the application of the source-language sequence labeling system to the target language without the need for re-training.  \citet{agerri-etal-2018-building} automatically annotate parallel data for multiple source languages and project the labeled data to a single target language. This method demonstrates that leveraging multiple sources can significantly enhance the quality of the projections. 

Instead of relying on automatics labels for the source part of a parallel corpus, \citet{tiedemann-etal-2014-treebank,fei-etal-2020-cross} use Machine Translation to automatically translate the sentences of a gold-labeled dataset to the target languages. The translated data is subsequently annotated by projecting the gold labels from the source dataset onto it. \citet{tiedemann-etal-2014-treebank} make use of GIZA++ for word alignments, whereas \citet{fei-etal-2020-cross} utilize the word alignment probabilities calculated with FastAlign, alongside the part-of-speech (POS) tag distributions of the source and target words, to enhance the precision of annotation projection.

In contrast to the methods previous mentioned,  \cite{shah2010synergy} implement a Translate-Test strategy in which input sentences are first translated into the source language using Google Translate. These sentences are then annotated by the model, and the annotations are projected onto the target sentences using alignments computed by GIZA++.

\subsubsection{Other annotation projection methods}

With the recent advancements in supervised machine translation, a new line of research has emerged which aims to replace word alignments in favour of directly using Machine Translation models.  

\begin{figure}[htb]
    \centering
    \includegraphics[width=\textwidth]{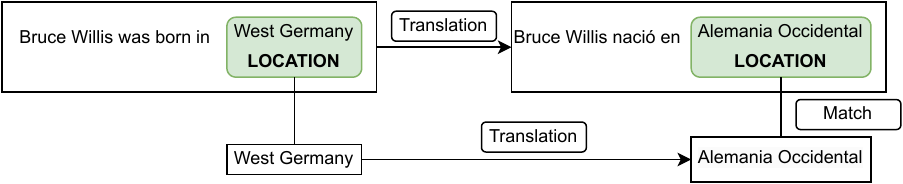}
    \caption{Illustration of annotation projection using Machine Translation. Individually labeled sequences are translated separately from the rest of the sentence. The translations of these sequences are then matched with the translations produced by translating the entire sentence.}
    \label{fig:chap3_translate_match}
\end{figure}

\citet{DBLP:conf/emnlp/JainPL19} introduce a "translate-match" methodology, which is illustrated in Figure~\ref{fig:chap3_translate_match}. In this approach the complete sentence, including labeled spans or entities, is first translated into the target language. Simultaneously, the labeled spans are translated independently of the full sentence. These individually translated spans are then matched with the corresponding spans in the translated sentence. However, this matching process does not guarantee that labeled spans will retain consistent translations when isolated from the sentence context, so the authors generate multiple translation candidates for each span and select the best match based on orthographic and phonetic similarities.

\begin{figure}[hbp]
    \centering
    \includegraphics[width=\textwidth]{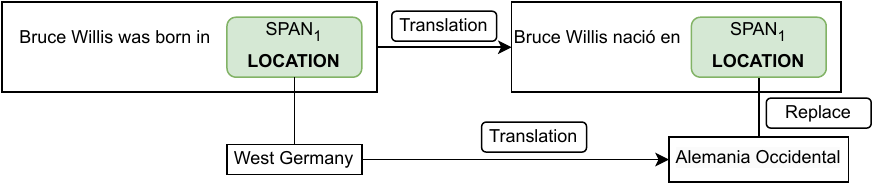}
    \caption{Illustration of annotation projection using Machine Translation and placeholders. Labeled sequences are replaced by a placeholder. The sentence with placeholders and the labeled sequences are translated independently. After translation, the placeholders are replaced with the corresponding labeled sequence translation.}
    \label{fig:chap3_translate_replace}
\end{figure}

To improve the matching process \cite{zhou-etal-2022-conner} propose to replace the labeled sequences in the source sentence with a placeholder. The sentence with placeholders and the labeled sequences are translated independently. Finally, the placeholders in the translated sentence are replaced with the corresponding labeled sequence translation. This process is illustrated in Figure \ref{fig:chap3_translate_replace}. They found that the translation model preserves the placeholders. Although this technique effectively addresses the matching challenge, it may introduce translation artifacts. These artifacts arise because the translation model does not process the entire context of the source sentence, potentially diminishing the overall quality of the translation.

\begin{figure}[ht]
    \centering
    \includegraphics[width=\textwidth]{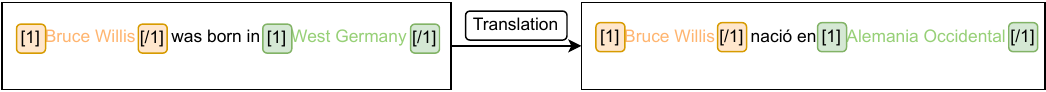}
    \caption{Illustration of the mark-then-translate approach. Markers are introduced around the labeled sequences. The sentence and the labeled spans are translated together.}
    \label{fig:chap3_translate_markers}
\end{figure}

Instead of translating the labeled sequences independently from the source sentence, \cite{daza-frank-2019-translate} and latter \cite{chen-etal-2023-frustratingly} introduce markers in the source sentence as depicted in Figure \ref{fig:chap3_translate_markers}. This mark-then-translate approach allows the model to jointly translate the source sentence and the labels. However, their studies reveal that the inclusion of markers can degrade the quality of the translation. Additionally, they encounter low projection rates, meaning that markers are frequently omitted in the translated output. To address this, they fine-tune the translation model using a synthetic dataset with named entity recognition annotations. Post-fine-tuning, the model not only preserves translation quality but it also surpasses the accuracy of word-alignment models in annotation projection tasks.

In a subsequent work \cite{DBLP:journals/corr/abs-2402-03131} enhances this method by implementing a constrained decoding algorithm, which ensures that the introduction of markers does not compromise the quality of the translation. In this improved approach, the training data in the high-resource language is first translated without markers. A second decoding phase then integrates the markers, with the constraint that the translation must align with the initial, marker-free output. This two-step process guarantees that the final translated sentence with markers remains consistent with what the model would have produced without them, thus preserving the translation quality.

Similar to the previous method, \cite{parekh2024contextual} also follows a two-step approach: first, translating the sentence without markers, and then performing the label translation. However, instead of using a Machine Translation system, they utilize an instruction-tuned large language model (Llama-2 \cite{DBLP:journals/corr/abs-2307-09288}) to perform the task in a few-shot setting with a few randomly selected examples. Although their method proves to be effective, most low-resource languages lack a high-quality instruction-tuned model, which limits the applicability of this approach.

\begin{figure}[htb]
    \centering
    \includegraphics[width=\textwidth]{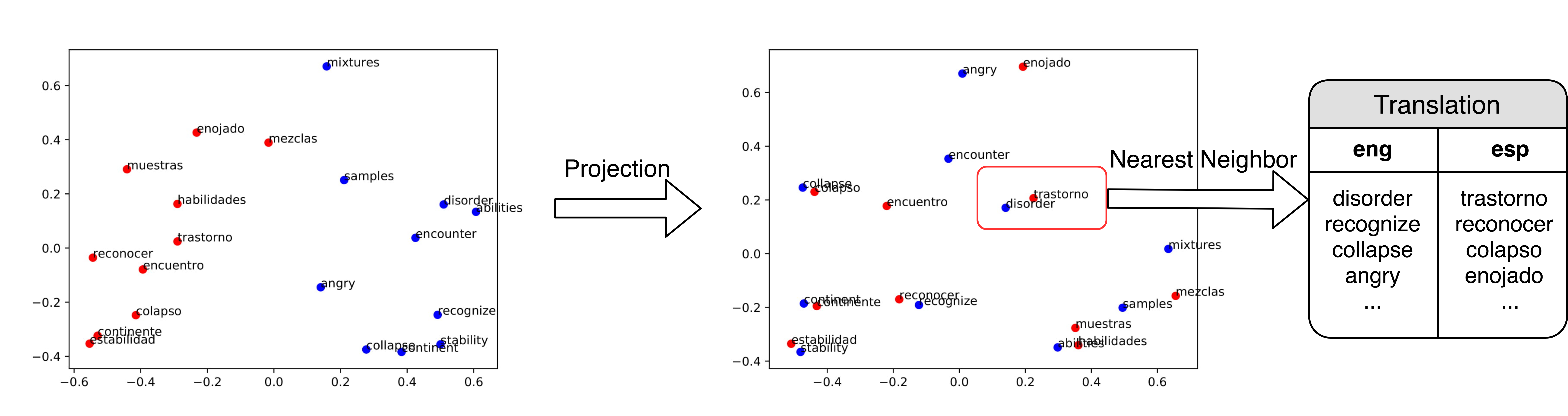}
    \caption{Illustration of bilingual dictionary generation. Monolingual embeddings are projected into a shared space in which a bilingual dictionary is computed by k-nearest-neighbor. Figure reproduced from \cite{xie-etal-2018-neural}.}
    \label{fig:chap3_xie2018}
\end{figure}

All the methods previously described presuppose the availability of a high-quality Machine Translation model and a sizeable parallel corpus containing both the source and target languages. However, this is not the case for all language pairs. For instance, such resources are nonexistent for translations between English and some very low-resource African languages. Taking this into account, an alternative research direction aims to facilitate data transfer between a high-resource language and a very low-resource language using only minimal resources, specifically a bilingual dictionary and unlabeled text in the target language.
\cite{xie-etal-2018-neural} first learns monolingual word embeddings (\cite{DBLP:journals/corr/abs-1301-3781,pennington-etal-2014-glove,DBLP:journals/tacl/BojanowskiGJM17}) for the source and the target language. As depicted in Figure \ref{fig:chap3_xie2018}, both embeddings are mapped into a bilingual vector space using a cross-lingual vector projection (\cite{zhang-etal-2016-ten,artetxe-etal-2016-learning,DBLP:conf/iclr/SmithTHH17}). A word translation dictionary is then computed by k-nearest-neighbor. The source sentence is translated word-by-word using this dictionary and the source label is copied for each corresponding word in the target sentence. Finally, a NER model is trained using the translated data. Building on this approach, \cite{DBLP:conf/acl/GuoR21} aim to refine the low-quality output that results from literal word-by-word translations. They employ a target language model and constrained beam search to produce text in the target language that exhibits a more natural and contextually appropriate word order. The constraints are designed to ensure the presence of entities in the generated text.

\subsection{Model-based transfer}

\begin{figure}
    \centering
    \includegraphics[width=0.5\textwidth]{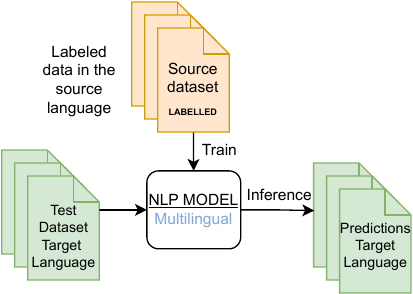}
    \caption{Illustration of the model-based coss-lingual transfer approach. A pre-trained multilingual model is finetuned with data in the source language and then applied without modification to label text in the target language.}
    \label{fig:chap3_zero}
\end{figure}

The model transfer approach involves leveraging multilingual models trained in high-resource languages to perform tasks in low-resource languages. In contrast, the data transfer paradigm focuses on manipulating the data to fit a monolingual model. Model transfer exploits the shared representation of languages in a pre-trained multilingual model. Thus, these models can be fine-tuned for a specific task in the source language and then applied without any modification to label text in any of the multiple languages the model supports. This approach is illustrated in Figure \ref{fig:chap3_zero}.

Some of the first attempts at model-based cross-lingual transfer (\cite{tackstrom-etal-2012-cross,kozhevnikov-titov-2014-cross,bharadwaj-etal-2016-phonologically,chaudhary-etal-2018-adapting}) leveraged the structural similarities between languages to facilitate tasks in languages with limited or non-existent training data. However, model-based transfer began to make significant progress (\cite{artetxe-schwenk-2019-massively,pires-etal-2019-multilingual}) following the introduction of Transformer-based (\cite{DBLP:conf/nips/VaswaniSPUJGKP17}) multilingual language models such as BERT (\cite{devlin-etal-2019-bert}) or XLM-RoBERTa (\cite{conneau-etal-2020-unsupervised}). These models were pre-trained using language modelling objectives on extensive datasets comprising over 100 languages. During this pre-training phase, the models acquired a shared representation for all included languages. Subsequently, these models can be fine-tuned on specific tasks with data from a source language and then applied to label data in any of the supported languages directly.

\selectlanguage{english}
\chapter[Data transfer vs Model transfer]{Data transfer vs Model transfer}
\label{ch:model-vs-data}

In this chapter we will conduct an in-depth study of the two main techniques used so far for cross-lingual zero-shot Sequence Labeling, focusing on either data or model transfer. We will apply state-of-the-art Machine Translation models, word alignments, and language models to assess the performance of various cross-lingual sequence labeling approaches. We will identify the advantages and shortcomings of each method, as well as the challenges faced by current techniques for cross-lingual zero-shot sequence labeling. These insights will serve as a foundation for our work in subsequent chapters.

\section{Motivation and contributions}
\label{sc4:intro}

The performance of supervised machine-learning methods for Natural Language Processing, including advanced 
deep-neural models (\cite{lample-etal-2016-neural,akbik-etal-2018-contextual,devlin-etal-2019-bert,conneau-etal-2020-unsupervised}),
heavily depends on the availability of manually annotated training data. 
In addition, supervised models show a significant decrease in
performance when evaluated in out-of-domain settings
(\cite{DBLP:conf/aaai/Liu0YDJCMF21}). This means that obtaining optimal results
would require to manually generate annotated data for each application
domain and language, an unfeasible task in terms of monetary cost and human
effort. 

zero-shot cross-lingual transfer approaches aim to apply the resources from a high-resource source language to low-resource target languages. In this chapter, we perform an in-depth study of the two main techniques employed so far for cross-lingual zero-shot sequence labeling, based either on data or model transfer. We implement both approaches using the latest advancements in machine translation, word aligners, and multilingual language models. We focus on two sequence labeling tasks, namely, Named Entity Recognition (NER) and Opinion Target Extraction (OTE). To this end, we present a data-based cross-lingual transfer approach consisting of translating gold-labeled data between English and seven other languages using state-of-the-art Machine Translation systems. Sequence labeling annotations are then automatically projected for every language pair using word alignments. We then compare the performance obtained for each of the target languages against the performance of the zero-shot cross-lingual method, consisting of fine-tuning the multilingual language models on the English gold data and generating the predictions in the required target languages.

The main contributions of this chapter are the following: 
\begin{itemize}
\item We empirically establish the required conditions for each of these two approaches, data-transfer and zero-shot model-based cross-lingual transfer, to outperform the other. In this sense, our experiments show that contrary to what previous research suggested (\cite{fei-etal-2020-cross,Li2021CrossLingualNE}), the zero-shot transfer approach is the most effective method when using high-capacity multilingual language models like XLM-R large. However, data transfer approaches remain valuable for models with weaker downstream cross-lingual performance. As there is no universally available multilingual pretrained model for every language and domain, data-based methods retain their relevance.
\item Our evaluation proves that despite high-quality machine translations and annotation projections, in the data transfer approach, we have identified issues like many-to-one translations or misalignments, which seem to account for the lower performance of data transfer methods compared to the model-based approach.
\item We perform an extensive error analysis which shows that using English gold-labeled data often produces a signal which, due to inherent differences in language usage, differs from the signal received when using gold-labeled data in the target language. This cultural misalignment problem hinders the performance of the zero-shot cross-lingual methods. The issue is more pronounced with low-capacity models that have lower generalization capabilities, whereas higher-capacity models are more successful at labeling words unseen in the training data that share a similar meaning with those seen during training.

\end{itemize}

\section{Methodology}
In this section, we describe our implementation of data-based and model-based transfer methods for Cross-Lingual Sequence Labeling. Our experiments follow a one-to-one framework, using English as the source language. Each model is evaluated in a single target language. We assume a scenario in which we have English gold-labeled train and development data. Furthermore, we also assume that the only gold-labeled data available for the target language is the evaluation set. 

\subsection{Data transfer}

In the data-transfer paradigm, described earlier in Chapter \ref{sc:transfer-methods}, the information extraction model always performs inference in the language it was trained in. This language could be either the source language, typically English, or the target language. We propose to use Machine Translation and annotation projection. We have implemented two distinct data-transfer strategies: Translate-Train and Translate-Test.

\begin{figure}
\centering
\begin{subfigure}{.45\textwidth}
  \centering
  \includegraphics[width=\linewidth]{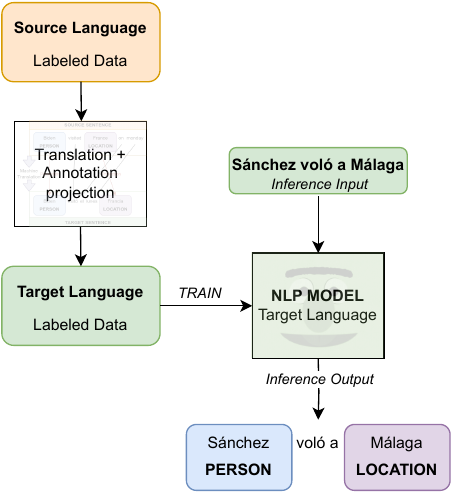}
  \caption{Translate-Train approach: We automatically generate data for the target language by translating the gold-labelled English data. We use this data to train an NLP system in the target language. At inference, the model can be used to label inputs in the target language.}
  \label{fig:Translate_train}
\end{subfigure}\hfill
\begin{subfigure}{.45\textwidth}
  \centering
  \includegraphics[width=\linewidth]{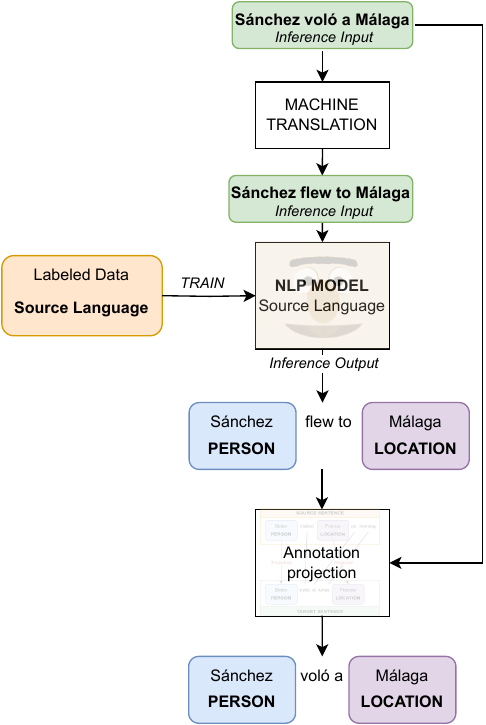}
  \caption{Translate-Test approach: We train a model with the gold labelled data in the source language. During the inference, the inputs are first translated into the source language. The outputs of the model are projected into the original target language sentence.}
  
  \label{fig:Translate_test}
\end{subfigure}
\caption{Illustration of the two data transfer approaches we have implemented. They are differentiated by the direction in which we translate the data. In both cases, English is the source language and Spanish is the target language.}
\label{fig:Translate_train_vs_translate_test}
\end{figure}

\subsubsection{Translate-Train}
In the Translate-Train approach, our goal is to generate data for the target language automatically, as depicted in Figure \ref{fig:Translate_train}. The process begins with translating English gold-labeled data into the target language using state-of-the-art Machine Translation models. Subsequently, we project the gold labels from the original English sentences onto the translated ones. This results in an automatically generated dataset in the target language. We then utilize this dataset to train an NLP model specifically for the target language. During inference, sentences in the target language are fed directly into the model for annotation. In this approach, the model is trained using ``silver'' data, the quality of which depends on the Machine Translation model's accuracy and the accuracy of the annotation projection algorithm. The data-transfer occurs during the model's training phase. Once trained, the model is capable of directly annotating sentences in the target language without requiring any additional steps

\subsubsection{Translate-Test}
The Translate-Test approach, depicted in Figure \ref{fig:Translate_test}, involves training an NLP model using English gold-labeled data. During the inference step, we first translate the input sentences in the target language into English. The model then generates annotations for these translated sentences. Subsequently, we project back these annotations onto the original sentences in the target language. Unlike the Translate-Train approach, where the model is trained with '``silver'' data, in the Translate-Test, the model is trained with gold data. However, during inference, the input must be first translated into English, and the model's output annotations must be projected back onto the original target language sentences. Therefore, as with Translate-Train, the quality of the predictions in Translate-Test also depends on the Machine Translation model's quality and the accuracy of the annotation projection algorithm.

\subsubsection{Annotation Projection Algorithm}

\begin{figure}[htb]
    \centering
    \includegraphics[width=8cm]{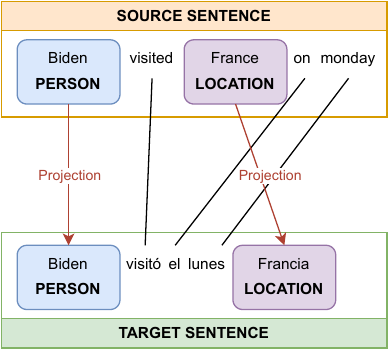}
    \caption{Illustration of the translation and annotation projection method for the Named Entity Recognition task.}
    \label{fig:annotation_projection}
\end{figure}

\begin{figure}
    \centering
    \includegraphics[width=\linewidth]{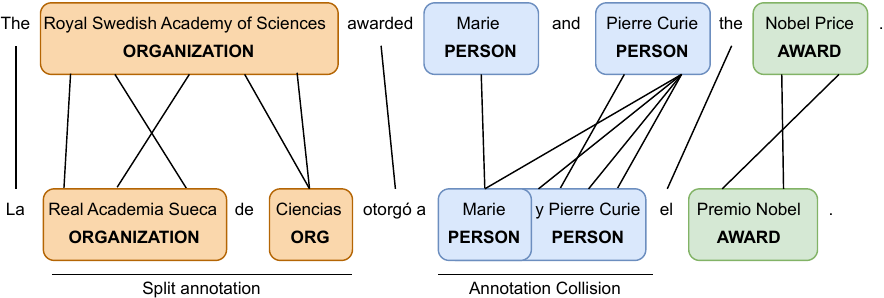}
    \caption{Illustration of the split annotation and annotation collision errors when projecting a sentence.}
    \label{fig:projection_errors}
\end{figure}

In both the Translate-Train and Translate-Test approaches, we project labels from the source language into the target language using automatic word alignments. Word alignments map words in a source language to their corresponding translations in a target language. Consider a source language sentence $x=\left\langle x_1,...,x_n  \right\rangle$ of length $n$, and its translation $y=\left\langle y_1,...,y_m \right\rangle$ in the target language with length $m$. We employ a word aligner to identify pairs $A=\left\{\left\langle x_{i}, y_{j}\right\rangle: x_{i} \in \mathbf{x}, y_{j} \in \mathbf{y}\right\}$, where each pair $\left\langle x_i,y_j\right\rangle$ indicates $y_j$ is the translation of $x_i$. For a sequence $s =\left\langle x_a,...,x_b  \right\rangle \in \mathbf{x}$ labeled with category $C$, we assign the same category $C$ to the sequence $t =\left\langle y_c,...,y_d  \right\rangle \in \mathbf{y}$, if every word $y_j$ in $t$ is aligned with at least one word $x_i$ in $s$: $\{ \forall y_j \in t \, \exists x_i \in  s : ( \left\langle x_i,y_j \right\rangle \in A )\}$. Essentially, if a word in the source sentence labeled with a category is aligned with a word in the target sentence, we label the target word with the same category. This method is illustrated in Figure \ref{fig:projection_errors}.

When projecting annotations, we encounter two primary challenges: \textit{Split Annotations} and \textit{Annotation Collisions}. Split Annotations occur when a labeled sequence in the source sentence is divided into multiple sequences in the target sentence, often due to missed alignments by the word alignment algorithm. In such cases, we merge the target sequences if they are separated by only one word. If multiple sequences remain, we retain the longest sequence and discard the others. Annotation Collisions arise when a word in the target sentence aligns with two different labeled sequences in the source sentence, typically resulting from incorrect alignments. This causes a word, or sequence of words, to have overlapping label categories in the target sentence. If the conflicting sequences belong to the same category, we merge them into a single label in the target sentence. If they are of different categories, we choose the longer target sequence and discard the other one. Additionally, if a punctuation symbol in the target sentence aligns with a labeled word in the source sentence, we disregard this alignment. These scenarios are illustrated in Figure \ref{fig:annotation_projection}. For example, in the projection of ``Royal Swedish Academy of Science'' to ``Real Academia Sueca de Ciencias'', the word ``of'' is incorrectly aligned, leading to a projection gap. Since this gap involves only one word, we merge the sequences, labeling ``Real Academia Sueca de Ciencias'' as a single organizational entity. In the case of ``Marie'' and ``Pierre Curie''', incorrect alignment of ``Curie'' with both names results in overlapping person spans in the target sentence. As they are of the same category (Person), we merge them into a single Person label. Although this approach might not be perfect, it helps minimize errors caused by incorrect alignments.

\subsection{Model transfer}

\begin{figure}[htb]
    \centering
    \includegraphics[width=8cm]{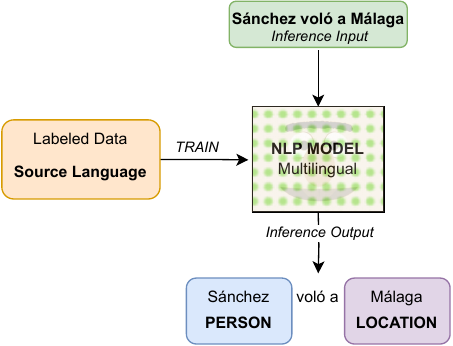}
    \caption{Illustration of model transfer approach. A multilingual model is trained in the source language (English). The model is then used to label sentences in the target language (Spanish).}
    \label{fig:model_transfer}
\end{figure}

Previously described in Chapter \ref{sc:transfer-methods}, the model transfer approach leverages multilingual language models \cite{devlin-etal-2019-bert, conneau-etal-2020-unsupervised}. These models are pre-trained on extensive unlabeled text corpora in multiple languages. As depicted in Figure \ref{fig:model_transfer}, we fine-tune a multilingual language model with the gold-labeled data available in English. Once the training is complete, the model is capable of directly labeling text in any of the languages included in its pre-training phase. We will refer to this approach to as \textit{zero-shot} cross-lingual sequence labeling.

\section{Experimental Setup}
\label{sc4:experimental-setup}

In this section, we will describe the experimental framework for this Chapter. 

\subsection{Datasets}
\label{sec:chap4_datasets}

\begin{figure}[htb]
    \centering
    \includegraphics[width=\linewidth]{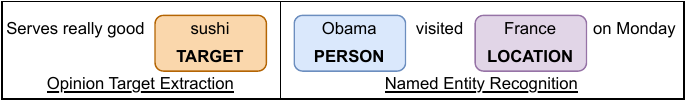}
    \caption{Illustration of the sequence labeling tasks used in the experiments in this chapter.}
    \label{fig:chapter4_tasks}
    \vspace{-0.5cm}
\end{figure}

We conducted experiments in two sequence labeling tasks, namely, Opinion Target Extraction and Named Entity Recognition. The tasks are illustrated in Figure \ref{fig:chapter4_tasks}. 
We list the number of examples in each dataset in table \ref{tab:Chapter4_dataset_size}. 

\paragraph{Opinion Target Extraction (OTE):} Given a review, the task is to detect the linguistic expression used to refer to the reviewed entity. For instance, in the sentence \textit{Serves really good sushi}, the word \textit{sushi} is the opinion target because it is the entity being discussed. We use the English SemEval 2016 Aspect Based Sentiment Analysys (ABSA) dataset (\cite{pontiki-etal-2016-semeval}). This dataset comprises user reviews from the restaurant domain. We experiment with the English, Spanish, Dutch, French, Russian and Turkish datasets from the restaurant domain. 

\paragraph{Named Entity Recognition (NER):} Given a text, the task is to detect named entities and classify them according to some pre-defined categories. For Spanish and Dutch we use the CoNLL-2002 datasets (\cite{DBLP:conf/conll/Sang02}). For English and German we use the CoNLL-2003 datasets (\cite{DBLP:conf/conll/SangM03}) and for Italian, we use the Evalita 2009 data (\cite{speranza2009named}). We map the Geo-Political Entities from Evalita 2009 to {\it location} labels to make them compatible with the CoNLL data. This dataset contains labeled sentences from news articles.

\begin{table}[htb]
\centering
\adjustbox{max width=\textwidth}{
\begin{tabular}{@{}lllccc@{}}
\toprule
Language & Dataset (Citation) & Train & Dev & Test & Labels \\ \midrule
\multicolumn{6}{l}{Opinion Target Extraction} \\ \midrule
English & SemEval 2016 ABSA (\cite{pontiki-etal-2016-semeval}) & 2000 & - & 676 & \multicolumn{1}{c}{\multirow{6}{*}{(1) Target}} \\
Spanish & SemEval 2016 ABSA (\cite{pontiki-etal-2016-semeval}) & 2070 & - & 881 & \\
French & SemEval 2016 ABSA (\cite{pontiki-etal-2016-semeval}) & 1665 & - & 668 & \\
Dutch & SemEval 2016 ABSA (\cite{pontiki-etal-2016-semeval}) & 1722 & - & 575 &  \\
Russian & SemEval 2016 ABSA (\cite{pontiki-etal-2016-semeval}) & 3655 & - & 1209 & \\
Turkish & SemEval 2016 ABSA (\cite{pontiki-etal-2016-semeval}) & 1232 & - & 144 & \\ \midrule
\multicolumn{6}{l}{Named Entity Recognition} \\ \midrule
English & CoNLL-2003 (\cite{DBLP:conf/conll/SangM03}) & 14987 & 3466 & 3684 & \multirow{5}{*}{\begin{tabular}[c]{@{}l@{}}(4) Person, \\ Location, \\ Organization, \\ Miscellaneous\end{tabular}} \\
Spanish & CoNLL-2002 (\cite{DBLP:conf/conll/Sang02}) & 6871 & 1914 & 1516 & \\
German & CoNLL-2003 (\cite{DBLP:conf/conll/SangM03}) & 12705 & 3068 & 3160 & \\
Dutch & CoNLL-2002 (\cite{DBLP:conf/conll/Sang02}) & 15806 & 2895 & 5195 & \\
Italian & Evalita 2009 (\cite{speranza2009named}) & 11227 & - & 4136 & \\ \bottomrule
\end{tabular}
}
\caption{Number of sentences for each dataset split.}
\label{tab:Chapter4_dataset_size}
\end{table}

\subsection{Machine Translation}

We tested different open-source and commercial Machine Translation systems. Including DeepL \footnote{\url{https://www.deepl.com/es/translator}}, OpusMT (\cite{tiedemann-thottingal-2020-opus}), M2M100 (1.2B, \cite{JMLR:v22:20-1307}) and mBART (mbart-large-50, \cite{liu-etal-2020-multilingual-denoising,DBLP:journals/corr/abs-2008-00401}). A qualitative analysis performed during the projection of the OTE labels established that DeepL produced more fluent translations. Thus we decided to use DeepL (web version during the second half of 2021) to perform the Machine Translation for our data-based cross-lingual transfer experiments. Turkish was not supported by DeepL at the time, so we used M2M100 1.2B instead. Experiments that compare the different Machine Translation models are discussed in Section \ref{chap4:sec_transaltionbenchmark}.

\subsection{Word Alignments}

To compute word alignments, we use the AWESOME system (\cite{dou-neubig-2021-word}). AWESOME leverages multilingual pre-trained Language Models and fine-tunes them on parallel corpora. Unsupervised training objectives over the parallel corpus improve the alignment quality of the models. The authors claim that AWESOME works better with multilingual-BERT \cite{devlin-etal-2019-bert} as the backbone, so we follow their advice. We also experiment with GIZA++ (\cite{och-ney-2003-systematic-giza}) and FastAlign (\cite{dyer-etal-2013-simple-fastalign}), which are models based on statistical machine translation. Additionally, we explore SimAlign (\cite{jalili-sabet-etal-2020-simalign}), which, similar to AWESOME, leverages multilingual pre-trained models, although in a fully unsupervised manner. All the systems are extensively described in Chapter \ref{ch:2_word_alignments}. As demonstrated in Section \ref{chap4:sec_alignmentbenchmark}, AWESOME produced the highest F1 scores when comparing the model projections to manually annotated projections.

\subsection{Sequence labeling Models}

We use three state-of-the-art multilingual pre-trained language models for sequence labeling: multilingual BERT (mBERT) (\cite{devlin-etal-2019-bert})
and XLM-RoBERTa (XLM-R) base and large (\cite{conneau-etal-2020-unsupervised}). 

The primary distinctions among these models lie in their parameter count and the size of their pretraining corpora. mBERT has 110 million parameters and was trained on Wikipedias\footnote{\url{https://www.wikipedia.org/}} of 104 languages. In contrast, XLM-RoBERTa-base contains 250 million parameters, and XLM-RoBERTa-large has 560 million parameters. Both versions of XLM-RoBERTa were trained using a corpus of 340 billion tokens extracted from 100 languages sourced from CommonCrawl\footnote{\url{https://commoncrawl.org/}}. This corpus, significantly larger as noted in \citet{conneau-etal-2020-unsupervised} encompasses a far broader range of data compared to that used for mBERT, covering various diverse domains. We categorize models with a lower parameter count and those pre-trained on smaller datasets, such as mBERT, as low-capacity language models. Conversely, we classify larger models trained on more extensive corpora, like XLM-RoBERTa-large, as high-capacity language models (\cite{aharoni-etal-2019-massively}).

\begin{figure}
    \centering
    \includegraphics[width=\linewidth, keepaspectratio]{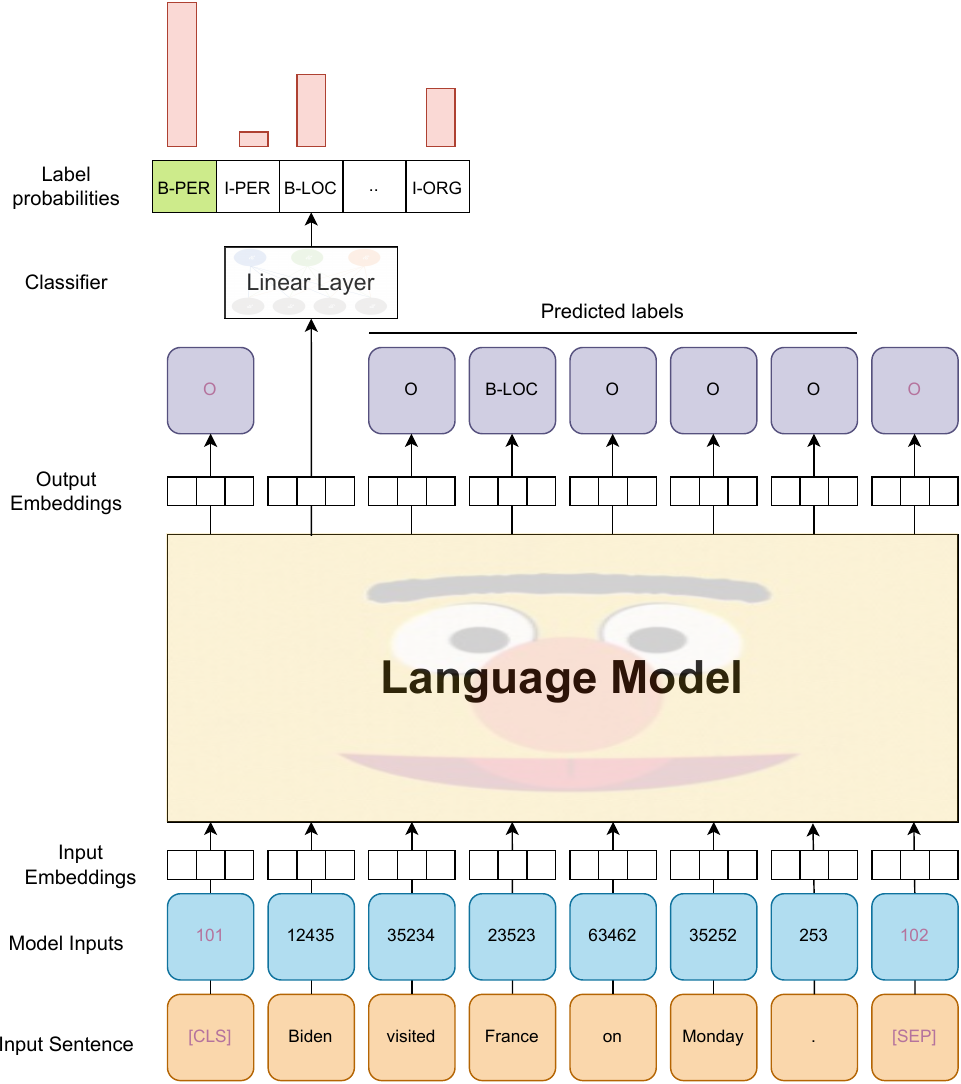}
    \caption{Implementation of the sequence labeling model. An encoder-based Language Model is fed the input sequence. The output representations are used by a token classification linear layer to predict the labels.}
    \label{fig:chapter4_tokenclassification}
\end{figure}

As depicted in Figure \ref{fig:chapter4_tokenclassification}, for both language models, we add a token classification layer (a linear layer) on top of each token's representation. This layer computes the probability distribution of the labels for each token. If a word is split into multiple sub-tokens, we use the representation of the first sub-token as input for the classifier. During training, we fine-tune the complete parameters of the model along with the token classification layer. We utilize the sequence labeling implementation from the Hugging Face open-source library (\cite{wolf-etal-2020-huggingface-transformers}). As listed in Section \ref{sec:chap4_datasets}, due to the significant difference in the total number of training examples between the OTE and NER datasets, we employ slightly different hyperparameter settings for each task. For OTE we use a batch size of $32$, $5e-5$ learning rate, and we train the model for $10$ epochs and $128$ maximum sequence length. Since only a train and test splits are available for the OTE task, we use the train set as both, train and development data. For NER we use a batch size of $32$, $2e-5$ learning rate, and we train the model for $4$ epochs and 256 maximum sequence length. For both tasks, we use the ``BILOU'' encoding scheme (\cite{ratinov-roth-2009-design}). 

We report the F1 score, which is the standard metric in sequence labeling tasks. F1 scores and standard deviation scores are reported by averaging the results of five runs with different random seeds. 

\section{Experimental Results}
\label{sec:chap4_experiments}
In this section we compare the \textsc{Translate-Train} and \textsc{Translate-Test} data-transfer approaches with the \textsc{zero-shot} model-transfer approach. As an upper bound, we also train the language models on gold-labeled data in the target languages. This upper bound, which we refer to as \textsc{gold}, is intended to assess the performance of the cross-lingual transfer approaches with respect to an ideal setting.

\subsection{Opinion Target Extraction}
Opinion Target Extraction (OTE) results are reported in Table \ref{tab:chap4_OTE_f1score}. When using mBERT, a low-capacity model, the zero-shot model transfer approach obtains better results for Spanish and French. However, for Dutch, Russian, and Turkish, the data-transfer approaches are superior. However, the overall picture changes as we increase the model capacity. When using XLM-RoBERTa (XLM-R) base, the zero-shot baseline is much closer to the gold upper bound than that of mBERT. This demonstrates that XLM-R has better multilingual transfer learning capabilities for these tasks. In fact, the zero-shot transfer outperforms the Translate-Train and Translate-Test for every language except Turkish. The XLM-R base results on gold-labeled data are also substantially better than those of mBERT, which once again demonstrates the better proficiency of XLM-R in these tasks. To summarise, XLM-R large offers the best cross-lingual transfer performance, as the zero-shot transfer is clearly superior for every language, including Turkish.

A common trend for all three models in the OTE benchmarks is that the Translate-Train approach consistently performs better than the Translate-Test approach. As expected, all the approaches achieve a performance significantly lower than the gold upper bound.

\begin{table}[htbp]
  \centering
  \small
\adjustbox{max width=.8\textwidth}{
\begin{tabular}{@{}lgccc@{}}
\toprule
\multicolumn{5}{c}{mBERT} \\
\midrule
Language  &  Gold  &  Zero-shot  &  Trans-Train  &  Trans-Test   \\
\midrule
English  & 76.2${\scriptscriptstyle\pm}$\tiny{0.9} &  -  &  -  &  -   \\
Spanish  & 75.2${\scriptscriptstyle\pm}$\tiny{0.5} & \textbf{68.4}${\scriptscriptstyle\pm}$\tiny{0.6} & 67.9${\scriptscriptstyle\pm}$\tiny{0.8} & 62.2${\scriptscriptstyle\pm}$\tiny{1.2} \\
French  & 74.0${\scriptscriptstyle\pm}$\tiny{1.1} & \textbf{62.7}${\scriptscriptstyle\pm}$\tiny{1.2} & 59.7${\scriptscriptstyle\pm}$\tiny{1.2} & 57.6${\scriptscriptstyle\pm}$\tiny{1.1} \\
Dutch  & 69.7${\scriptscriptstyle\pm}$\tiny{0.9} & 61.7${\scriptscriptstyle\pm}$\tiny{0.8} & 64.3${\scriptscriptstyle\pm}$\tiny{1.5} & \textbf{67.0}${\scriptscriptstyle\pm}$\tiny{0.8} \\
Russian  & 72.5${\scriptscriptstyle\pm}$\tiny{0.5} & 53.8${\scriptscriptstyle\pm}$\tiny{2.2} & \textbf{62.9}${\scriptscriptstyle\pm}$\tiny{0.6} & 59.7${\scriptscriptstyle\pm}$\tiny{0.4} \\
Turkish  & 62.0${\scriptscriptstyle\pm}$\tiny{1.2} & 45.3${\scriptscriptstyle\pm}$\tiny{4.0} & \textbf{45.7}${\scriptscriptstyle\pm}$\tiny{2.3} & 35.5${\scriptscriptstyle\pm}$\tiny{2.4} \\
\midrule
\multicolumn{5}{c}{XLM-R base}\\
\midrule
English  & 84.4${\scriptscriptstyle\pm}$\tiny{0.9} &  -  &  -  &  -   \\
Spanish  & 81.1${\scriptscriptstyle\pm}$\tiny{0.7} & \textbf{78.2}${\scriptscriptstyle\pm}$\tiny{0.4} & 72.5${\scriptscriptstyle\pm}$\tiny{0.7} & 62.9${\scriptscriptstyle\pm}$\tiny{0.9} \\
French  & 80.2${\scriptscriptstyle\pm}$\tiny{0.6} & \textbf{72.7}${\scriptscriptstyle\pm}$\tiny{0.3} & 64.7${\scriptscriptstyle\pm}$\tiny{0.8} & 60.0${\scriptscriptstyle\pm}$\tiny{0.6} \\
Dutch  & 80.8${\scriptscriptstyle\pm}$\tiny{1.7} & \textbf{75.5}${\scriptscriptstyle\pm}$\tiny{0.8} & 70.0${\scriptscriptstyle\pm}$\tiny{1.6} & 71.0${\scriptscriptstyle\pm}$\tiny{1.5} \\
Russian  & 81.5${\scriptscriptstyle\pm}$\tiny{0.3} & \textbf{74.9}${\scriptscriptstyle\pm}$\tiny{0.9} & 69.5${\scriptscriptstyle\pm}$\tiny{0.3} & 62.2${\scriptscriptstyle\pm}$\tiny{1.6} \\
Turkish  & 69.0${\scriptscriptstyle\pm}$\tiny{1.1} & 58.1${\scriptscriptstyle\pm}$\tiny{3.5} & \textbf{58.9}${\scriptscriptstyle\pm}$\tiny{1.8} & 36.4${\scriptscriptstyle\pm}$\tiny{1.8} \\
\midrule
\multicolumn{5}{c}{XLM-R large} \\
\midrule
English  & 86.4${\scriptscriptstyle\pm}$\tiny{1.1} &  -  &  -  &  -   \\
Spanish  & 83.6${\scriptscriptstyle\pm}$\tiny{0.1} & \textbf{79.3}${\scriptscriptstyle\pm}$\tiny{0.8} & 73.7${\scriptscriptstyle\pm}$\tiny{1.1} & 64.0${\scriptscriptstyle\pm}$\tiny{1.4} \\
French  & 82.2${\scriptscriptstyle\pm}$\tiny{0.6} & \textbf{74.6}${\scriptscriptstyle\pm}$\tiny{1.7} & 66.1${\scriptscriptstyle\pm}$\tiny{0.6} & 60.7${\scriptscriptstyle\pm}$\tiny{0.6} \\
Dutch  & 80.4${\scriptscriptstyle\pm}$\tiny{2.1} & \textbf{77.7}${\scriptscriptstyle\pm}$\tiny{1.9} & 74.0${\scriptscriptstyle\pm}$\tiny{1.3} & 72.9${\scriptscriptstyle\pm}$\tiny{1.8} \\
Russian  & 82.8${\scriptscriptstyle\pm}$\tiny{0.4} & \textbf{76.8}${\scriptscriptstyle\pm}$\tiny{1.3} & 69.3${\scriptscriptstyle\pm}$\tiny{2.3} & 62.2${\scriptscriptstyle\pm}$\tiny{1.3} \\
Turkish  & 72.3${\scriptscriptstyle\pm}$\tiny{2.4} & \textbf{62.4}${\scriptscriptstyle\pm}$\tiny{1.0} & 57.8${\scriptscriptstyle\pm}$\tiny{2.4} & 33.7${\scriptscriptstyle\pm}$\tiny{0.9} \\
\bottomrule
\end{tabular}
}
  \caption{OTE F1 scores with models of different capacities in the SemEval 2016 ABSA (\cite{pontiki-etal-2016-semeval}) dataset.}
  \label{tab:chap4_OTE_f1score}
\end{table}

\subsection{Named Entity Recognition}

The NER results presented in Table \ref{tab:chap4_zerovsSota} show a number of different patterns. First, the zero-shot approach using mBERT outperforms the data-based cross-lingual transfer methods for the majority of languages. Second, the Translate-Test approach is consistently superior to the Translate-Train approach. Third, the mBERT performance on gold data is similar to that of XLM-R base. Finally, the data transfer approaches achieve the best performance when using XLM-R base for German and Italian.

The only, result that remains consistent from the OTE tasks is that the zero-shot approach using XLM-R large achieves the best results for all languages.

\begin{table}[htb]
  \centering
  \small
\adjustbox{max width=0.8\textwidth}{
\begin{tabular}{@{}lgccc@{}}
\toprule
\multicolumn{5}{c}{mBERT}\\
\midrule
 Language &  Gold  &  Zero-shot  &  Trans-Train  &  Trans-Test   \\
\midrule
English  & 90.7${\scriptscriptstyle\pm}$\tiny{0.3} &  -  &  -  &  -  \\
Spanish  & 87.4${\scriptscriptstyle\pm}$\tiny{0.4} & \textbf{74.6}${\scriptscriptstyle\pm}$\tiny{0.4} & 69.5${\scriptscriptstyle\pm}$\tiny{0.4} & 70.8${\scriptscriptstyle\pm}$\tiny{0.6} \\
German  & 82.0${\scriptscriptstyle\pm}$\tiny{0.4} & \textbf{71.0}${\scriptscriptstyle\pm}$\tiny{0.9} & 70.1${\scriptscriptstyle\pm}$\tiny{0.3} & 70.6${\scriptscriptstyle\pm}$\tiny{0.5} \\
Dutch  & 90.8${\scriptscriptstyle\pm}$\tiny{0.4} & \textbf{78.5}${\scriptscriptstyle\pm}$\tiny{0.5} & 74.4${\scriptscriptstyle\pm}$\tiny{0.6} & 75.4${\scriptscriptstyle\pm}$\tiny{0.8} \\
Italian  & 84.7${\scriptscriptstyle\pm}$\tiny{0.3} & 68.2${\scriptscriptstyle\pm}$\tiny{0.5} & 68.7${\scriptscriptstyle\pm}$\tiny{0.5} & \textbf{70.7}${\scriptscriptstyle\pm}$\tiny{0.3} \\
\midrule
\multicolumn{5}{c}{XLM-R base}\\
\midrule
English  & 90.4${\scriptscriptstyle\pm}$\tiny{0.2} &  -  &  -  &  -   \\
Spanish  & 87.7${\scriptscriptstyle\pm}$\tiny{0.2} & \textbf{75.0}${\scriptscriptstyle\pm}$\tiny{0.4} & 70.1${\scriptscriptstyle\pm}$\tiny{0.6} & 72.5${\scriptscriptstyle\pm}$\tiny{0.2} \\
German  & 83.1${\scriptscriptstyle\pm}$\tiny{0.3} & 67.9${\scriptscriptstyle\pm}$\tiny{0.5} & \textbf{70.5}${\scriptscriptstyle\pm}$\tiny{0.5} & 70.1${\scriptscriptstyle\pm}$\tiny{0.8} \\
Dutch  & 89.8${\scriptscriptstyle\pm}$\tiny{0.2} & \textbf{78.1}${\scriptscriptstyle\pm}$\tiny{0.6} & 73.3${\scriptscriptstyle\pm}$\tiny{0.9} & 74.7${\scriptscriptstyle\pm}$\tiny{0.4} \\
Italian  & 84.3${\scriptscriptstyle\pm}$\tiny{0.3} & 71.2${\scriptscriptstyle\pm}$\tiny{0.5} & 71.1${\scriptscriptstyle\pm}$\tiny{0.4} & \textbf{71.7}${\scriptscriptstyle\pm}$\tiny{0.3} \\
\midrule
\multicolumn{5}{c}{XLM-R large}\\
\midrule
English  & 92.4${\scriptscriptstyle\pm}$\tiny{0.1} &  -  &  -  &  -   \\
Spanish  & 88.9${\scriptscriptstyle\pm}$\tiny{0.2} & \textbf{79.5}${\scriptscriptstyle\pm}$\tiny{1.0} & 70.9${\scriptscriptstyle\pm}$\tiny{0.6} & 74.0${\scriptscriptstyle\pm}$\tiny{0.5} \\
German  & 85.1${\scriptscriptstyle\pm}$\tiny{0.6} & \textbf{74.5}${\scriptscriptstyle\pm}$\tiny{0.7} & 73.7${\scriptscriptstyle\pm}$\tiny{0.5} & 72.9${\scriptscriptstyle\pm}$\tiny{0.3} \\
Dutch  & 92.9${\scriptscriptstyle\pm}$\tiny{0.7} & \textbf{82.3}${\scriptscriptstyle\pm}$\tiny{0.6} & 77.5${\scriptscriptstyle\pm}$\tiny{0.9} & 77.2${\scriptscriptstyle\pm}$\tiny{0.6} \\
Italian  & 87.5${\scriptscriptstyle\pm}$\tiny{0.2} & \textbf{76.0}${\scriptscriptstyle\pm}$\tiny{0.5} & 73.7${\scriptscriptstyle\pm}$\tiny{0.4} & 73.5${\scriptscriptstyle\pm}$\tiny{0.6} \\
\bottomrule
\end{tabular}
}
  \caption{NER F1 scores with models of different capacities in the CoNLL-2002 (\cite{DBLP:conf/conll/Sang02}) and CoNLL-2003 (\cite{DBLP:conf/conll/SangM03}) datasets.}
  \label{tab:chap4_NER_f1score}
\end{table}

We also compare our data-transfer and model-transfer implementations with previous research that leverages parallel data and/or annotation projection methods on the NER CoNLL 2002-2003 data. The results are listed in Table \ref{tab:chap4_zerovsSota}. XLM-R large in a zero-shot setting not only outperforms our data-transfer implementation, but it is also superior to previous data-transfer approaches. The only exception is the result obtained by \cite{Li2021CrossLingualNE} for German. 

Our data transfer approaches, although not achieving the best results for every language, are competitive with previously proposed methods. It is important to note that while we only leverage translations of the NER data, previous research uses other resources such as the automatic annotation of large parallel corpora.

\begin{table}[tbp]
    \centering

\adjustbox{max width=\textwidth}{
\begin{tabular}{@{}lll|ccc@{}}
\toprule
Method & Model & Approach & Spanish & German & Dutch \\
\midrule

\cite{dou-neubig-2021-word} & mBERT  & Translate train & 64.3 & - & - \\
\cite{DBLP:conf/emnlp/JainPL19} & BiLSTM + CRF  &  Translate train & 73.5 & 61.5 & 69.9 \\
\cite{DBLP:conf/acl/GuoR21}  & BiLSTM + CRF & Translate train & 77.9 & 71.4 & 80.6 \\
\cite{Li2021CrossLingualNE} &  XLM-R large & Translate train &   78.9 & \textbf{76.9} & 79.7 \\
\midrule
Ours & XLM-R base  & Translate train  & 70.1 & 70.5 & 73.3 \\
Ours & XLM-R base  & Translate test  & 72.5 & 70.1 & 74.7 \\
Ours & XLM-R large  & Translate train  & 70.9 & 73.7 & 77.5 \\
Ours & XLM-R large  & Translate test  & 74.0 & 72.9 & 77.2 \\
\midrule
Ours & mBERT  & zero-shot & 74.6 & 71.0 & 78.5 \\
Ours & XLM-R base  & zero-shot & 75.0 & 67.9 & 78.1 \\
Ours & XLM-R large  & zero-shot & \textbf{79.5} & 74.5 & \textbf{82.3} \\

\bottomrule
\end{tabular}}
    \caption{Comparison between the previous research methods that leverage projections, the zero-shot baselines and our annotation projections in the CoNLL-2002 (\cite{DBLP:conf/conll/Sang02}) and CoNLL-2003 (\cite{DBLP:conf/conll/SangM03}) datasets. F1 score reported}
    \label{tab:chap4_zerovsSota}
\end{table}

\subsection{Discussion}

Previous research has demonstrated (\cite{pires-etal-2019-multilingual, wu-dredze-2020-languages}) that mBERT's performance in cross-lingual transfer learning greatly differs depending on whether the source and target languages are topologically similar or not, with the former being the scenario in which the model performs best. Secondly, the monolingual performance of mBERT, as well as its cross-lingual transfer performance, is much better for high-resource languages than for low-resource languages. This is consistent with our results for mBERT in the NER and OTE tasks. The zero-shot transfer works best for French and Spanish, which are somewhat topologically similar to English. However, the performance of the zero-shot approach is worse than the Translate-Train and Translate-Test approaches for Russian and Turkish.

\begin{figure}[htb]
    \centering
    \includegraphics[width=0.6\linewidth]{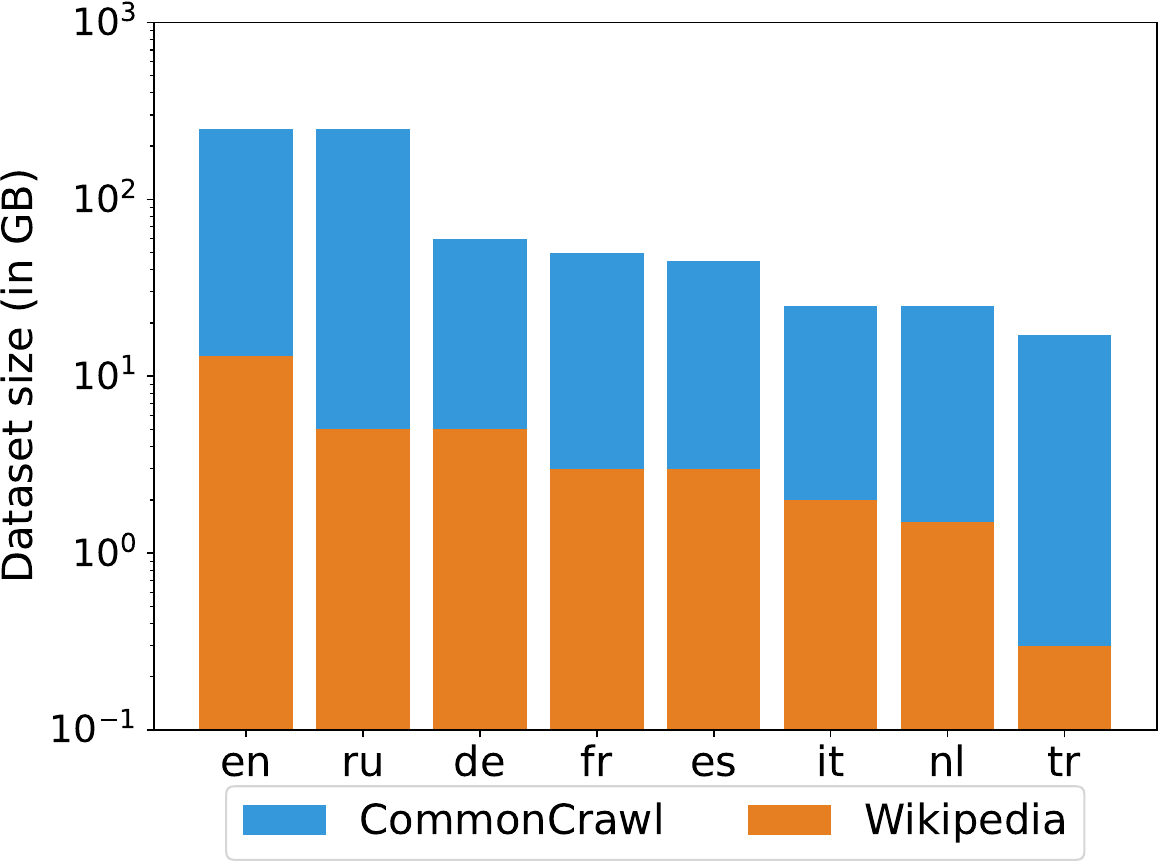}
    \caption{Amount of data in GiB (log-scale) for the languages we use in our experiments in Wiki-100 corpus used for training mBERT and the CC-100 used for XLM-R. The full figure can be found in \cite{conneau-etal-2020-unsupervised}}
    \label{fig:chapter4_pretraining_size}
\end{figure}

While mBERT was trained using Wikipedia data for 104 languages (\cite{devlin-etal-2019-bert}), XLM-R (both base and large) has been trained using CommonCrawl (\cite{conneau-etal-2020-unsupervised}), a much larger multilingual corpus with a variety of texts extracted from the web. As illustrated in Figure \ref{fig:chapter4_pretraining_size}, XLM-R was trained with orders of magnitude more data for Russian and Turkish compared to mBERT, thus acquiring higher proficiency in those languages. This results in the zero-shot approach using XLM-R achieving better performance in Russian and Turkish than the data-transfer approaches. The CommonCrawl dataset also contains a more diverse variety of texts sampled from the Web, perhaps including texts of similar domains to those in the OTE datasets. This may account for the large differences in OTE performance between XLM-R and mBERT. In this sense, the similar performance between mBERT and XLM-R base for NER may be partially attributed to the fact that the CoNLL and Evalita datasets consist of news stories in which most of the labeled entities may appear in the Wikipedia texts used to pre-train mBERT.

Our results suggest that the performance of model-transfer and data-transfer approaches relies on the model's proficiency in the target language and data domain.  If a high-capacity model, such as XLM-R large, with strong proficiency in the target language and domain is available, the zero-shot cross-lingual transfer method achieves the best results. However, for languages and domains where the model lacks sufficient proficiency, data-based cross-lingual transfer (Translate-Train and Translate-Test) approaches remain mainly useful. While XLM-R large in a zero-shot cross-lingual setting achieves the best results for every task and language in the experiments performed in this section, in the next chapters we will demonstrate that XLM-R large is not proficient in every language and domain, for example, in African languages (\cite{adelani-etal-2022-masakhaner}). Thus, developing both better model-transfer and data-transfer approaches is of great relevance.

\section{Error Analysis}
The experiments described in Section \ref{sec:chap4_experiments} showed that when using XLM-R large, the zero-shot approach outperforms the Translate-Train and Translate-Test approaches. The effectiveness of the Translate-Train and Translate-Test approaches depends on the quality of the Machine Translation and annotation projection models. In this section, we will assess the performance of both Machine Translation and annotation projection to better understand if the data-transfer methods are hindered by the performance of any of these steps. We will also conduct an error analysis to gain a deeper understanding of the shortcomings of Translate-Train and Translate-Test in comparison to the zero-shot cross-lingual transfer.

\subsection{Downstream evaluation of Machine Translation Models}
\label{chap4:sec_transaltionbenchmark}

To assess the impact of the Machine Translation system employed, we translated the English OTE gold-labeled data using four distinct translation systems. In all experiments, we utilized AWESOME as the word aligner for annotation projection. We fine-tuned XLM-R large with each set of generated training data and assessed its performance against the gold-labeled test set for each target language. Based on the results provided in Table \ref{tab:chap4_translators}, we can conclude that there are no significant differences in the final F1 scores when employing different translation systems. For each language, a different model demonstrates the best performance, although all systems exhibit similar average performance across all languages. The exception is Turkish,  a language in which MarianMT underperforms.

\begin{table}[htbp]
  \centering
  \small
\adjustbox{max width=.8\textwidth}{
\begin{tabular}{@{}ccccc@{}}
\toprule
Language &  MarianMT  &  Mbart  &  M2M100  &  DeepL  \\
\midrule
Spanish  & \textbf{75.6}${\scriptscriptstyle\pm}$\tiny{0.8} & 75.3${\scriptscriptstyle\pm}$\tiny{0.7} & 74.2${\scriptscriptstyle\pm}$\tiny{0.8} & 73.7${\scriptscriptstyle\pm}$\tiny{1.1} \\
French  & 64.5${\scriptscriptstyle\pm}$\tiny{1.6} & \textbf{66.4}${\scriptscriptstyle\pm}$\tiny{1.1} & 64.9${\scriptscriptstyle\pm}$\tiny{1.3} & 66.1${\scriptscriptstyle\pm}$\tiny{0.6} \\
Dutch  & 70.0${\scriptscriptstyle\pm}$\tiny{2.0} & 68.8${\scriptscriptstyle\pm}$\tiny{4.0} & 70.1${\scriptscriptstyle\pm}$\tiny{3.1} & \textbf{74.0}${\scriptscriptstyle\pm}$\tiny{1.3} \\
Russian  & 66.6${\scriptscriptstyle\pm}$\tiny{4.4} & \textbf{69.7}${\scriptscriptstyle\pm}$\tiny{1.4} & \textbf{69.7}${\scriptscriptstyle\pm}$\tiny{0.7} & 69.3${\scriptscriptstyle\pm}$\tiny{2.3} \\
Turkish  & 49.5${\scriptscriptstyle\pm}$\tiny{2.9} & 56.1${\scriptscriptstyle\pm}$\tiny{5.2} & \textbf{57.8}${\scriptscriptstyle\pm}$\tiny{2.4} &  -  \\
\bottomrule
\end{tabular}}
  \caption{OTE F1 score in the SemEval 2016 ABSA \cite{pontiki-etal-2016-semeval} dataset of different XLM-R large models trained using data generated with different translation systems.}
  \label{tab:chap4_translators}
\end{table}

\subsection{Evaluating the Projection Method}
\label{chap4:sec_alignmentbenchmark}

The performance of the data transfer method relies on the quality of the annotation projections. If the annotation projection algorithm does not produce highly accurate projections, it will generate noisy data that would hinder both the Translate-Train and Translate-Test approaches. In this section, we will evaluate the performance of various word alignment systems and compute the performance of the Translate-Test approach using gold annotation projections. To achieve this, we enlisted human annotators who manually projected labels from the English OTE gold-labeled data onto automatic translations in Spanish, French, and Russian. The machine translations were generated using DeepL for Spanish, French, and Russian, and M2M100 for Turkish. The annotators are NLP PhD candidates with either native or proficient skills in both English and the target language. For this experiment, we developed an application to assist during the annotation process. The annotator views the sentence in English, with a highlighted target, and must select the corresponding target in a translated sentence. The complete guidelines and the application code provided to the annotators are available in GitHub \footnote{\url{https://github.com/ikergarcia1996/Annotation-Projection-App}}.

\begin{table}[htbp]
  \centering
  \small
\adjustbox{max width=.8\textwidth}{
    \begin{tabular}{@{}ccccc@{}}
\toprule
Language & GIZA++ & FastAlign & SimAlign & AWESOME \\
\midrule
Spanish & 77.0 & 75.0 & 86.7 & \textbf{91.5} \\
French & 73.3 & 72.9 & 86.3 & \textbf{91.3} \\
Russian & 72.4 & 76.9 & 87.7 & \textbf{93.4} \\
Turkish & 64.0 & 68.4 & 81.9 & \textbf{88.5} \\
\bottomrule
\end{tabular}}
  \caption{OTE F1 score in the  SemEval 2016 ABSA \cite{pontiki-etal-2016-semeval} dataset between the human annotation projections vs the automatic projections generated using different alignment models.}
  \label{tab:chap4_alignmentsbenchmark}
\end{table}

First, we compare the projection of annotations automatically generated by different word alignment methods with those provided by human annotators. Table \ref{tab:chap4_alignmentsbenchmark} shows that language model-based methods (SimAlign and AWESOME) outperform statistically based methods (GIZA++ and FastAlign) by a wide margin. AWESOME consistently outperforms SimAlign for every language. It is worth mentioning that SimAlign uses a fully unsupervised algorithm, while AWESOME requires fine-tuning a Language Model; therefore, the cost in compute resources is significantly higher for AWESOME.

\begin{table}[tbp]
  \centering
  \small
\adjustbox{max width=.8\textwidth}{
\begin{tabular}{@{}cccc@{}}
\toprule
Language &  Translate Train  &  Translate Train (Manual) & Zero-shot \\
\midrule
Spanish  & 73.7${\scriptscriptstyle\pm}$\tiny{1.1} & 75.1${\scriptscriptstyle\pm}$\tiny{1.2} & \textbf{79.3}${\scriptscriptstyle\pm}$\tiny{0.8} \\
French  & 66.1${\scriptscriptstyle\pm}$\tiny{0.6} & 67.9${\scriptscriptstyle\pm}$\tiny{1.0}  & \textbf{74.6}${\scriptscriptstyle\pm}$\tiny{1.7} \\
Russian  & 69.3${\scriptscriptstyle\pm}$\tiny{1.3} & 69.4${\scriptscriptstyle\pm}$\tiny{2.1} & \textbf{76.8}${\scriptscriptstyle\pm}$\tiny{1.3} \\
Turkish  & 57.8${\scriptscriptstyle\pm}$\tiny{2.4} & 50.6${\scriptscriptstyle\pm}$\tiny{1.4} & \textbf{62.4}${\scriptscriptstyle\pm}$\tiny{1.0} \\
\bottomrule
\end{tabular}
}
  \caption{XLM-R large OTE F1 score in the  SemEval 2016 ABSA \cite{pontiki-etal-2016-semeval} dataset when training with automatically and manually projected datasets}
  \label{tab:chap4_ManualvsAutomaticProjection}
\end{table}

The performance of the AWESOME system confirms that it is possible to generate high-quality annotations that closely resemble those generated by human experts. However, the model does make some mistakes. To understand the performance implications of the mistakes produced by AWESOME, we compare the Translate-Train approach when using data projected with AWESOME and when using data manually projected by human experts. To achieve this, we fine-tuned an XLM-R large model on both datasets. Table \ref{tab:chap4_ManualvsAutomaticProjection} shows that when training with manually projected data, the results are slightly better, except for Turkish, which again acts as an outlier. Nevertheless, the results when training with manually projected data are still inferior to the zero-shot model-transfer approach. From the results, we can conclude that the projection mistakes produced by AWESOME do not have a significant impact on the performance of the data-based cross-lingual transfer approach, although there is still room for improvement. More importantly, the data-transfer approach is not inferior to the model-based transfer approach due to the errors produced by the annotation projection step.

\subsection{Categorization of mistakes}
We have evaluated the quality of the Machine Translation models and the annotation projection algorithms. We found both to be of high quality and not responsible for the data-transfer method performing worse than the zero-shot transfer approach. In this section we will categorize the errors produced by each approach and compare them. Table \ref{tab:chap4_Mistakes} displays the most frequent false negatives and positives for which there is a significant discrepancy in frequency between methods.

There are a few challenging words that all the systems struggle with. For example, as previously reported by \cite{AGERRI201985}, the words ``comida'' (food) and ``restaurante'' (restaurant) are highly ambiguous in the ABSA task. Both appear labeled as target and unlabeled frequently. As expected, models struggle with these words. In addition, we have identified four main sources of errors:
\begin{table}[htbp]
    \centering
    \small
    
\adjustbox{max width=.9\textwidth}{
\begin{tabular}{@{}cc|ccc@{}}
\toprule
English Word & Spanish Word & English Gold & Spanish Gold & Spanish Translate \\
\midrule
Service & Servicio & 153 & 229 & 133 \\
Treatment & Trato & 0 & 54 & 0 \\
Attention & Atención & 2 & 35 & 0 \\
\midrule
Place & Sitio & 120 & 41 & 2 \\
Place & Lugar & 120 & 19 & 91 \\
\bottomrule
\end{tabular}}
    \caption{Number of times words appear as target words in the SemEval 2016 ABSA (\cite{pontiki-etal-2016-semeval}) train dataset.}
    \label{tab:chap4_Wordcount}
    \vspace{-0.5cm}
\end{table}

\begin{table}[tbp]
    \centering
    \small
\adjustbox{max width=\textwidth}{
\begin{tabular}{lgggcccgggcccg}
\toprule
& \multicolumn{3}{c}{GOLD} & \multicolumn{3}{c}{Zero-shot} & \multicolumn{3}{c}{Tr-Train} & \multicolumn{3}{c}{Tr-Test} & \multicolumn{1}{c}{Total} \\
&  B & Xb & Xl & B & Xb & Xl & B & Xb & Xl & B & Xb & Xl & \multicolumn{1}{c}{} \\
\midrule
& \multicolumn{13}{c}{OTE False Negatives} \\
\midrule
comida & 3 & 3 & 2 & 6 & 2 & 1 & 4 & 1 & 1 & 1 & 9 & 5 & 98 \\
restaurante & 7 & 5 & 7 & 9 & 5 & 6 & 7 & 6 & 6 & 7 & 12 & 10 & 43 \\
servicio & 2 & 2 & 2 & 2 & 1 & 1 & 2 & 0 & 1 & 1 & 1 & 2 & 85 \\
trato & 1 & 1 & 0 & 5 & 6 & 1 & 14 & 10 & 5 & 6 & 8 & 6 & 19 \\
atención & 2 & 3 & 3 & 8 & 2 & 3 & 7 & 1 & 3 & 7 & 7 & 7 & 13 \\
lugar & 0 & 0 & 0 & 2 & 0 & 0 & 1 & 0 & 0 & 0 & 1 & 0 & 12 \\
sitio & 1 & 0 & 0 & 5 & 1 & 1 & 3 & 3 & 3 & 2 & 1 & 1 & 14 \\
\midrule
& \multicolumn{13}{c}{NER False Negatives} \\
\midrule
de & 32 & 29 & 33 & 45 & 51 & 90 & 233 & 252 & 264 & 148 & 146 & 167 & 450 \\
la & 4 & 5 & 3 & 10 & 12 & 16 & 63 & 62 & 62 & 45 & 44 & 45 & 174 \\
Gobierno & 0 & 0 & 0 & 17 & 53 & 64 & 72 & 70 & 75 & 30 & 45 & 67 & 80 \\
Estado & 0 & 0 & 0 & 4 & 4 & 8 & 9 & 8 & 9 & 6 & 6 & 8 & 10 \\
Administación & 0 & 0 & 0 & 4 & 8 & 11 & 10 & 11 & 11 & 5 & 5 & 7 & 11 \\
Economía & 0 & 0 & 0 & 2 & 6 & 2 & 7 & 8 & 8 & 5 & 6 & 8 & 8 \\
Plan & 0 & 0 & 0 & 1 & 2 & 2 & 3 & 5 & 5 & 1 & 4 & 7 & 8 \\
Junta & 0 & 0 & 0 & 0 & 0 & 0 & 4 & 10 & 8 & 2 & 3 & 5 & 24 \\
Hacienda & 0 & 0 & 0 & 1 & 3 & 0 & 4 & 4 & 4 & 4 & 3 & 4 & 5 \\
\midrule
& \multicolumn{13}{c}{NER False Positives} \\
\midrule
español & 0 & 0 & 0 & 16 & 16 & 2 & 16 & 16 & 12 & 13 & 14 & 15 & 0 \\
catalán & 0 & 0 & 0 & 8 & 8 & 5 & 7 & 7 & 8 & 8 & 8 & 8 & 0 \\
\bottomrule
\end{tabular}
}
    \caption{Most common false negatives and positives were there is a big mismatch between methods and the total number of labelled appearances of the word in the test data. B is the acronym for mBERT, Xb for XLM-R base and Xl for XLM-R large.}
    \label{tab:chap4_Mistakes}
\end{table}
\paragraph{Many-to-one translation:} Multiple words can share the same sense in a given context and be used interchangeably. This is the case with opinion targets in the ABSA task, such as ``trato'', ``atención'', and ``servicio'' in Spanish. In the context of restaurant reviews, they all convey the same meaning as the English word ``service.'' There are 160 sentences in the English gold-labeled data containing the word ``service''; in 153 of them, ``service'' is labeled as a target. Machine Translation systems, such as DeepL in our experiments, systematically translate it as ``servicio.'' However, as shown in Table \ref{tab:chap4_Wordcount}, in the Spanish gold-labeled data, ``service'' is also commonly referred to as ``trato'' or ``atención'', not just ``servicio.'' Therefore, the training dataset translated and projected from English into Spanish, as demonstrated in previous sections, encompasses high-quality translations and annotation projections. Still, this dataset is not a good reflection of reviews written by native Spanish speakers. As shown in Table \ref{tab:chap4_Wordcount}, the translated dataset does not contain any occurrences of ``trato'' and ``atención'', which often appear in the gold-labeled Spanish test data. In fact, both the zero-shot and the data-based cross-lingual transfer approaches fail to correctly label these words as demonstrated in Table \ref{tab:chap4_Mistakes}. Interestingly, the zero-shot approach using XLM-R large correctly classifies ``trato'' (only failing to label 1 of the 19 occurrences). 

In the case of the Translate-Train approach, as we use a model of increased capacity, the number of false negatives decreases. This shows that the issue is more pronounced with low-capacity models that have lower generalization capabilities, while higher-capacity models are more successful at labeling words unseen in the training data that share a similar meaning with those in the training data. 

The Translate-Test approach does not overcome this issue, as when translating data from the target to the source language, the Machine Translation model does not systematically translate ``trato'' and ``atención'' as ``service.'' Instead, these words are usually translated as ``treatment'' and ``attention'', respectively. In this case, the Machine Translation system fails to understand the context in which the word is used, opting for literal translations. ``Treatment'' and ``attention'' are not commonly used in the restaurant review domain in the same context as ``service.'' In fact, there is no occurrence of the word ``treatment'' in the English gold-labeled data and there are only two occurrences of the word ``attention'', which contrasts with the 153 occurrences of ``service.'' Once again, the model struggles with labeling these words.

There are more examples of many-to-one translations, such as the word ``place'', which in Spanish can be most frequently translated as ``lugar'' or ``sitio.'' However, DeepL almost always translates it as ``lugar'', resulting in ``sitio'' being absent in the automatically generated training data, despite being more frequent than ``lugar'' in the gold-labeled data. In this particular case, the Translate-Test approach is not subject to the problem as both ``lugar'' and ``sitio'' are translated into ``place'' in the Spanish-to-English translation direction.

\paragraph{Cultural alignment:} There is a group of words related to Spanish Government names which are not commonly used in the same contexts in English, constituting a significant portion of the false negatives in the NER datasets listed in Table \ref{tab:chap4_Mistakes}. For example, the word ``Economía'' refers to the ``Ministry of Economy'' or ``Ministerio de Economía'' in Spanish. ``Junta'' denotes a ``local government'' administering a specific region in Spain. ``Plan'' is often used to denote specific ``projects founded by the goverment''. While these words frequently appear in the Spanish data as part of named entities, this is not the case in the English data, where it is more customary to use terms like ``Treasury Department'' (or variations thereof), correctly translated into Spanish by DeepL as ``Departamento del Tesoro''. Thus, during fine-tuning on the translated data, the model does not receive any signal to learn that ``Economy'' may be part of an organization entity. For other terms like ``Plan'' and ``Junta'', the English gold-labelled data lacks references, as these entities are specific to Spanish administration. While the data transfer approach misses many of these named entities, the zero-shot approach is more successful. Training the model with translated data or performing inference with data from other languages translated into English seems to complicate the transfer between different domains. In any case, there is still a significant margin for improvement. The gold standard, trained with gold-labelled data in Spanish, correctly labelled all these entities.

\paragraph{Errors induced by incorrect or missing alignments:} We found that for NER, articles and prepositions (i.e. ``de'', ``la'') are among the words with higher false positive rate for the Translate-Test and Translate-Train approaches. Examining the annotation projection reveals that word aligners struggle to correctly align articles in complex multi-word named entities specially when a one-to-many or many-to-one alignment is required. For example, the word aligners we tested failed to correctly align ``of'' with ``de la'' in the following example: ``Consejo General de la Arquitectura Técnica de España'' (General Council of Technical Architecture of Spain).

\paragraph{Errors induced by annotation inconsistencies:} Finally, another issue is the differences across the original gold-labelled annotations. ``Gobierno'' (Government) and ``Estado'' (State) are labelled as organizations in the Spanish gold-labelled data, but they are not considered to be entities in the English gold-labelled data. The opposite occurs with demonym words. They are labelled as miscellaneous entities in the English data but in Spanish they are not annotated. Cross-lingual models are likely to fail labeling these cases.

Summarizing, we observe that using English gold-labeled data often produces a signal which, due to inherent differences in language usage, differs from the signal received when using gold-labelled data in the target language. This accounts for the substantial performance disparity between all the cross-lingual sequence labeling approaches and the gold standard trained with gold-labelled data in the target language. The zero-shot cross-lingual transfer approach, when employing high-capacity language models, achieves the best transfer performance across languages. This appears to be due to this method being less impacted by issues related to many-to-one translations and cultural alignment. These issues together with miss-alignments seems to be the most common reason for the larger number of false positive and negatives of the data-based cross-lingual transfer method with respect to the zero-shot technique.

\section{Conclusions}

In this chapter we have performed an in-depth and comprehensive evaluation of model-based and data-based zero-shot cross-lingual sequence labeling on two different tasks. A detailed error analysis demonstrates that cross-lingual transfer is hindered by the differences in the cultural behaviour of the source and target language in use. The issue is more pronounced with low-capacity models that have lower generalization capabilities, whereas higher-capacity models are more successful at labeling words unseen in the training data that share a similar meaning with those in the training data. This suggests that developing models with enhanced generalization capabilities could bridge this cultural gap.

Our findings indicate that the zero-shot transfer approach is the most effective method when using high-capacity multilingual language models like XLM-R large. However, data-based cross-lingual transfer approaches remain valuable for models with weaker downstream cross-lingual performance. As there is no universally available multilingual pretrained model for every language and domain, data-based methods retain their relevance. Despite the availability of high-quality machine translations and annotation projections, we have identified issues like many-to-one translations or misalignments, which seem to account for the lower performance of data-based cross-lingual transfer methods compared to the model-based approach.

Our results establish that there is still room for improving the cross-lingual performance of zero-shot sequence labeling. In the following chapters, we will focus on two areas. First, we aim to develop a more effective annotation projection method that enables data-based approaches to achieve comparable or superior performance to the model-based approach. Second, we will work on improving the generalization capabilities of sequence labeling models to address the cultural alignment and many-to-one translation issues identified in this study.

\selectlanguage{english}
\chapter[Improving Data Transfer]{Improving Data Transfer}
\label{ch:data-transfer}

In this chapter we will introduce a novel approach to annotation projection that is based on large pre-trained text-to-text language models and state-of-the-art Machine Translation technology. Our algorithm, named T-Projection, significantly outperforms previous methods of annotation projection by a wide margin. Thanks to this new approach, we have achieved the best results to date for zero-shot cross-lingual transfer between English and eight different low-resource African languages.

\section{Motivation and contributions}

In the previous chapter we demonstrated that model-based zero-shot cross-lingual transfer outperforms data-based approaches. However, model-based zero-shot transfer requires a pre-trained model with high proficiency in both the language and domain of application. Such models are not available for every language and domain, which is the case for the eight different low-resource African languages that will be the subject of study in this chapter. Therefore, data-transfer methods still hold value. The inferior performance of data-transfer methods compared with model-transfer in the previous chapter can be attributed to two main factors. First, Machine Translation of English gold-labeled data often produces a signal which, due to inherent differences in language usage, differs from the signal received when using gold-labeled data in the target language. While both model-transfer and data-transfer approaches are sensitive to this issue, it has a greater impact on the data-transfer approach. This phenomenon can be mitigated by improving the generalization capabilities of sequence labeling models. We will focus on improving generalization in the following chapters. The second factor involves the mis-alignments produced by current annotation projection methods, which use word alignment algorithms. In this chapter, we will focus on this second issue.

The majority of previous published work on annotation projection explore the application of different word-alignments. However, as demonstrated in Chapter \ref{ch:model-vs-data}, word alignments often produce partial, incorrect or missing annotation projections. This is because word alignments are computed on a word-by-word basis by leveraging word co-occurrences or similarity between vector representations.  It should be noted that this method does not take into consideration the labeled spans or categories to be projected. 

In this chapter we present \textsc{T-Projection}, a novel approach to automatically project sequence labelling across languages. We split the annotation projection into two steps. First, we use mT5 (\cite{mt5}) text-to-text model to generate a set of projection candidates in the target sentence for each labeled category in the source sentence. This step exploits the labeled categories and the cross-lingual capabilities of large pre-trained multilingual language models. Second, we rank the candidates based on the probability of being generated as a translation of the source spans. We use the M2M100
(\cite{JMLR:v22:20-1307}) and NLLB200 (\cite{DBLP:journals/corr/abs-2207-04672}) state-of-the-art MT models to compute the translation probabilities
(\cite{DBLP:journals/corr/abs-2204-13692}).

The main contributions of this chapter are the following: 
\begin{itemize}
\item We have developed a new annotation projection method, T-Projection. We compare the label projections generated by various systems with manually projected annotations on three different tasks, Opinion Target Extraction (OTE), Named Entity Recognition (NER) and Argument Mining (AM), and five different target languages (French, German, Italian, Russian and Spanish. On average, T-Projection improves the current state-of-the-art annotation projection methods by more than 8 points in F1 score.
\item We performed a real-world NER task evaluation involving eight low-resource African languages. In this downstream evaluation, T-Projection outperforms other annotation projection methods by 3.6 points in F1 score. 
\end{itemize}

\begin{figure}[t]
\centering
\includegraphics[width=0.5\textwidth]{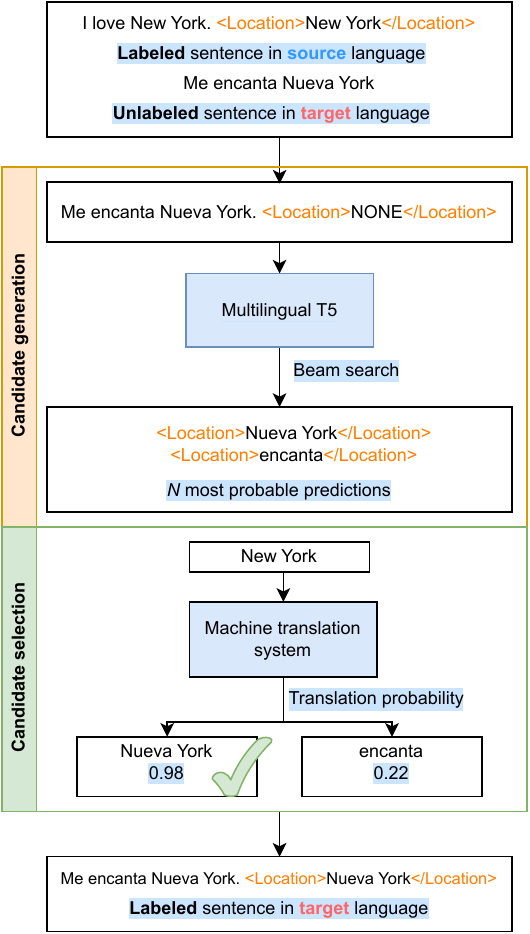}
\caption{T-Projection two-step method to project sequence
labeling annotations across languages.}
\label{fig:Tprojection}
\end{figure}

\section{T-Projection}
\label{sc4:Tprojection}
T-Projection assumes that we have a set of source sentences with sequences of words labeled with a category. Additionally, there is a parallel version of these sentences in a target language, though these translations are not labeled. T-Projection addresses the challenge of transferring the labels from the source sentences to the target sentences.

T-Projection implements two main steps, which are illustrated in Figure \ref{fig:Tprojection}. First, a set of projection candidates in the target sentence are generated for each labeled sequence in the source sentence. Second, each projection candidate is ranked using a Machine Translation model. More specifically, candidates are scored based on the probability of being generated as a translation of the source-labeled sequences. 

While the \emph{candidate generation} step exploits the labeled spans and their categories in the source sentence as well as the zero-shot cross-lingual capabilities of large pre-trained multilingual language models, the \emph{candidate selection} step
applies state-of-the-art MT technology to find those 
projection candidates that constitute the best translation for each source labeled span. These two steps are described in detail in the following two subsections. 

\subsection{Candidate Generation}\label{sc5:candidate-generation}

\begin{figure}[t]
\centering
\includegraphics[width=0.99\textwidth]{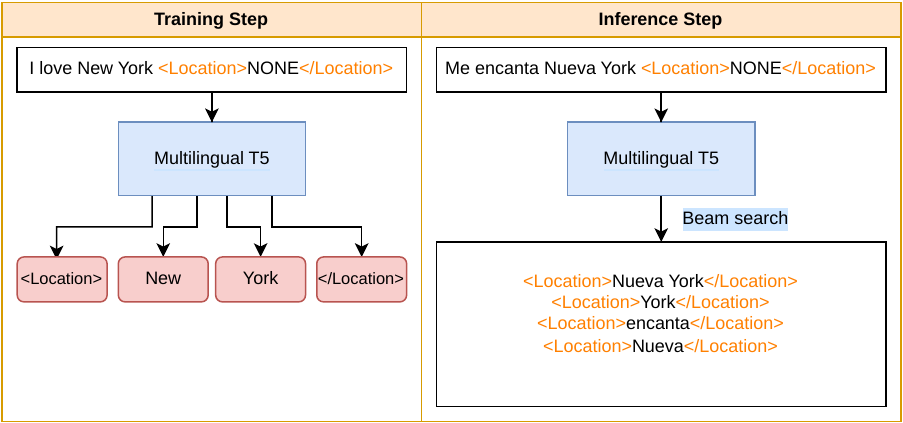}
\caption{Illustration of the candidate generation step. For each label, we generate a set of probable candidates.}
\label{fig:CandiateGen}
\end{figure} 

When projecting labeled sequences from a source dataset into its parallel target dataset, we expect both the source and the target to contain the same number of sequences, each labeled with the same category. For example, consider the English source sentence \textit{``<Person>Obama</Person> went to <Location>New York</Location>''} and its parallel, unlabeled Spanish target sentence \textit{``Obama fue a Nueva York''} We would expect the target sentence also to identify the same two entities (person and location). We propose a projection candidate generation step based on this premise. 

We finetune the text-to-text mT5 \cite{mt5} model using a HTML-tag-style prompt template
(\cite{huang-etal-2022-multilingual-generative}). As illustrated by Figure \ref{fig:CandiateGen}, we build the inputs for the model by concatenating the unlabeled sentence followed by a list of tags (\textit{``<Category>None</Category''}) with the category of each labeled span that we expect to find in the sentence, and the value ``None'' If two or more spans share the same category, we append the tag as many times as there are expected spans in that category.

Unlike the approach of \citet{huang-etal-2022-multilingual-generative}, we do not encode the tags for each category as special tokens. Instead, we verbalize the categories (i.e., PER $\,\to\,$ Person) and use the token representations already existing in the model. We anticipate that thanks to the language modeling pretraining, T5 will have a good semantic representation of sequence labeling types such as Person, Location, Claim, etc.

As Figure \ref{fig:CandiateGen} illustrates, we fine-tune mT5 with the labeled source dataset. We train the model to replace the token \emph{None} with the sequence of words in the input sentence that corresponds to that category.

At inference, we label the target sentences, which are parallel translations of the source sentences. As previously explained, we expect to identify the same number of labeled spans and categories as in the source sentence. Therefore, we use the labels from the corresponding source sentence to construct the prompts. In other words, our goal is to label parallel translations of the sentences used for training. We leverage the zero-shot cross-lingual capabilities of mT5 to project the labels from the source to the target sentence. The output tokens are generated in an autoregressive manner. We employ beam search decoding with 100 beams to generate 100 candidates for each input tag. The decision to generate 100 candidates was informed by a preliminary analysis of the performance of the candidate generation stage. Experiments in Section \ref{sec5:Howmany} demonstrate that this number was an overestimation, and generating 10 to 25 candidates is optimal.

\section{Candidate Selection}\label{sec5:candidate-selection}

\begin{figure}[t]
\centering
\includegraphics[width=0.60\textwidth]{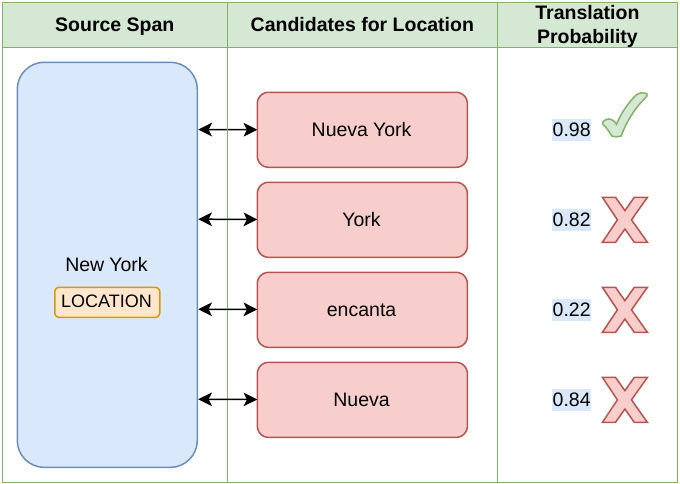}
\caption{Candidate selection: candidates are scored based on the probability of being generated as a translation of the source labeled sequences.}
\label{fig:CandiateSelection}
\end{figure}

In the previous step, we generated up to 100 candidate projections for each labeled span in the source sentence. In the candidate selection step, our goal is to identify the best projection candidate in the target sentence for each labeled span in the source sentence. As depicted in Figure \ref{fig:CandiateSelection}, all generated candidates are first grouped by category. For instance, if the previous step produced multiple spans with the same category (e.g., two \textit{locations} in a sentence), all such candidates are included in a single set. Additionally, candidates that are not subsequences of the input sentence are filtered out.

For each labeled span in the source sentence, we rank all the projection candidates that share the same category as the source span using their translation probabilities (also known as translation equivalence). These probabilities are obtained by applying the pretrained M2M100 (\cite{JMLR:v22:20-1307}) or NLLB200 (\cite{DBLP:journals/corr/abs-2207-04672}) MT models and the \textit{NMTScore} library\footnote{\url{https://github.com/ZurichNLP/nmtscore}} (\cite{DBLP:journals/corr/abs-2204-13692}). Given the source span $\mathbf{A}$ and the candidate $\mathbf{B}$, the translation probability is computed as follows (\cite{DBLP:journals/corr/abs-2204-13692}):

\begin{center}
$p_{\theta_a}(A \mid B):=\left[\prod_{i=0}^{|A|} p_{\theta_a}\left(A^i \mid B, A^{<i}\right)\right]^{\frac{1}{|A|}}$
\end{center}

\noindent The translation probability is normalized:

\begin{center}
$sim(A \mid B) = \frac{p_{\theta_a}(A \mid B)}{p_{\theta_a}(A \mid A)}$
\end{center}

Since translation probability can vary depending on the translation direction, the scores are symmetrized by calculating the scores for both translation directions and averaging them:

\begin{center}
$sim(A, B)=\frac{1}{2} sim(A \mid B)+\frac{1}{2} sim(B \mid A)$
\end{center}

Finally, for each labeled span in the source sentence, we select the candidate in the target sentence with the highest translation probability. Once a candidate has been chosen, that candidate and any others that overlap with it are removed from the set of possible candidates. This prevents the assignment of the same candidate in the target sentence to multiple spans in the source sentence.

\section{Experimental Setup}\label{sec5:Methodology}

To evaluate our method we perform both intrinsic and extrinsic evaluations.

\paragraph{Intrinsic Evaluation:} We selected several datasets that have been manually projected from English into various target languages. These manual annotations serve as the gold standard for evaluating and comparing T-Projection against previous state-of-the-art label projection models. Results are reported using the F1-score, a standard metric for sequence labeling (\cite{DBLP:conf/conll/Sang02}). The intrinsic evaluation focuses on measuring the annotation projection accuracy of the models, isolated from other factors such as the quality of the translation models or any other steps in the pipeline.

\paragraph{Extrinsic evaluation:} In this evaluation we assess the capability of T-Projection to automatically generate training data for sequence labeling tasks, NER in this particular case. The process begins by utilizing the Machine Translation system NLLB200 (\cite{DBLP:journals/corr/abs-2207-04672}) to translate data from English into 8 low-resource African languages. We then project the labels from English onto the respective target languages. The automatically generated datasets are then employed to train NER models, which are evaluated using a relatively small manually annotated test set. The same procedure is performed with other state-of-the-art label projection models. The comparison of the results obtained is reported in terms of F1-score.

\subsection{Datasets}\label{sec5:datasets}
\begin{figure}[htb]
    \centering
    \includegraphics[width=0.8\linewidth]{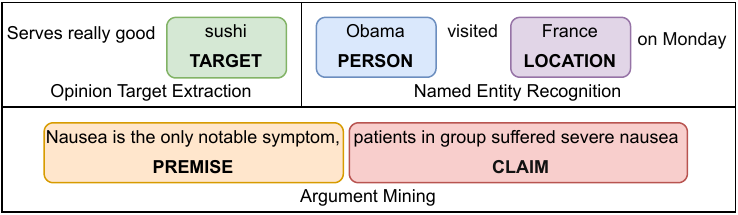}
    \caption{Sequence labeling tasks in our experiments}
    \label{fig:TasksChap5}
\end{figure}

The datasets used correspond to three sequence labeling tasks which are illustrated by Figure \ref{fig:TasksChap5}. The number of examples and labels are listed in Table \ref{tab5:DatasetLen}.

\begin{table}[htb]
    \centering
 \adjustbox{max width=0.8\linewidth}{
\begin{tabular}{@{}lccc@{}}
\toprule
Task & Split & \multicolumn{1}{c}{Sentence No} & \multicolumn{1}{c}{Labels}\\ \midrule
\multicolumn{4}{c}{ABSA} \\ \midrule
ABSA (\cite{pontiki-etal-2016-semeval}) & Train & 2000 & \multicolumn{1}{c}{\multirow{2}{*}{(1) Target}} \\
ABSA (\cite{pontiki-etal-2016-semeval}) & Test & 676 & \\ \midrule
\multicolumn{4}{c}{NER} \\  \midrule
Europarl (\cite{agerri-etal-2018-building}) & Test & 799 & \multirow{12}{*}{\begin{tabular}[c]{@{}l@{}}(3) Person, \\ Location, \\ Organization, \end{tabular}} \\
CoNLL03 (\cite{DBLP:conf/conll/SangM03}) & Train & 14987 & \\
CoNLL03 (\cite{DBLP:conf/conll/SangM03}) & Dev & 3466 & \\
CoNLL03 (\cite{DBLP:conf/conll/SangM03}) & Test & 3684 & \\ 
MasakhaNER2.0 (\cite{adelani-etal-2022-masakhaner}) & Test (hau) & 1632 & \\
MasakhaNER2.0 (\cite{adelani-etal-2022-masakhaner}) & Test (ibo) & 2180 & \\
MasakhaNER2.0 (\cite{adelani-etal-2022-masakhaner}) & Test (sna) & 1772 & \\
MasakhaNER2.0 (\cite{adelani-etal-2022-masakhaner}) & Test (swa) & 1882 & \\
MasakhaNER2.0 (\cite{adelani-etal-2022-masakhaner}) & Test (xho) & 1632 & \\
MasakhaNER2.0 (\cite{adelani-etal-2022-masakhaner}) & Test (yor) & 1963 & \\
MasakhaNER2.0 (\cite{adelani-etal-2022-masakhaner}) & Test (nya) & 1784 & \\
MasakhaNER2.0 (\cite{adelani-etal-2022-masakhaner}) & Test (zul) & 1669 & \\ \midrule
\multicolumn{4}{c}{AM} \\ \midrule
AbsRCT Neoplasm (\cite{DBLP:conf/ecai/0002CV20}) & Train & 4404 & \multirow{3}{*}{\begin{tabular}[c]{@{}l@{}}(2) Claim, \\ Premise \end{tabular}} \\
AbsRCT Neoplasm (\cite{DBLP:conf/ecai/0002CV20}) & Dev & 679 & \\
AbsRCT Neoplasm (\cite{DBLP:conf/ecai/0002CV20}) & Test & 1251 & \\
\bottomrule
\end{tabular}
}

    \caption{Size (Number of sentences) of the dataset we use to train and evaluate our systems.}
    \label{tab5:DatasetLen}
\end{table}

\paragraph{Opinion Target Extraction (OTE)} Given a review, the task of Opinion Target Extraction (OTE) is to identify the linguistic expressions that refer to the reviewed entity. For instance, in the sentence \textit{Serves really good sushi}, the word \textit{sushi} is the opinion target because it is the entity being discussed. We utilize the English SemEval 2016 Aspect Based Sentiment Analysis (ABSA) datasets (\cite{pontiki-etal-2016-semeval}). This dataset contains user reviews from the Restaurant domain. Additionally, for evaluation purposes, we employ parallel datasets in Spanish, French, and Russian, which were generated through Machine Translation and manual label projection in Chapter \ref{ch:model-vs-data}. 

\paragraph{Named Entity Recognition (NER)} The NER task involves detecting named entities and classifying them according to predefined categories. We use a parallel NER dataset in English, Spanish, German, and Italian (\cite{agerri-etal-2018-building}), based on Europarl data (transcriptions of discussions from the European Parliament) (\cite{DBLP:conf/mtsummit/Koehn05}). 
These transcriptions are parallel in multiple languages and were annotated following the CoNLL 2003 guidelines (\cite{DBLP:conf/conll/SangM03}). For the extrinsic evaluation, we use MasakhaNER 2.0 (\cite{adelani-etal-2022-masakhaner}), a human-annotated NER dataset for 20 African languages.

\paragraph{Argument Mining (AM)} The AbstRCT English dataset includes annotations for two types of argument components, Claims and Premises, in medical and scientific texts collected from the MEDLINE database (\cite{DBLP:conf/ecai/0002CV20}). A \textit{Claim} is a concluding statement made by the author about the study's outcome, such as an assertion of a diagnosis or a treatment in the medical domain. A \textit{Premise} is an observation or measurement (ground truth) that supports or challenges another argument component, usually a claim. Premises are considered observed facts and are credible without further evidence. For evaluation, we used the Spanish parallel counterpart, generated following an adapted version of the method described in Chapter \ref{ch:model-vs-data}. The labeled sequences in the AM task consist of very long spans of words, frequently encompassing full sentences. We use the Neoplasm split.

\subsection{Baselines}

\begin{wrapfigure}{r}{0.46\textwidth}
    \vspace{-0.95cm}
    \centering
    \includegraphics[width=\linewidth]{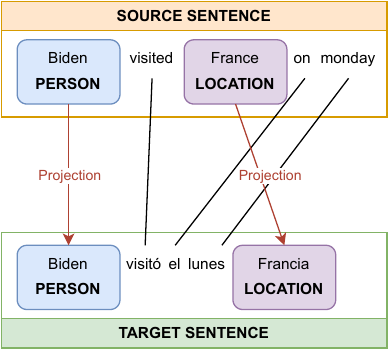}
    \caption{Illustration of the translation and annotation projection task using word-alignments.}
    \label{fig:projection_with_alignments}
    \vspace{-0.5cm}
\end{wrapfigure}

We use the same alignment systems and methodology described in Chapter \ref{ch:model-vs-data} as a baseline. This approach is illustrated in Figure \ref{fig:projection_with_alignments}. We compare T-Projection with two statistical systems, \textbf{Giza++} (\cite{och-ney-2003-systematic-giza}) and \textbf{FastAlign} (\cite{dyer-etal-2013-simple-fastalign}). These systems are widely used in the field and require very small computational resources. Additionally, we evaluate two current state-of-the-art Transformer-based word-alignment systems, \textbf{SimAlign} (\cite{jalili-sabet-etal-2020-simalign}) and \textbf{AWESOME} (\cite{dou-neubig-2021-word}), which leverage pre-trained multilingual language models to generate alignments. As recommended by the authors, we use multilingual BERT (mBERT) (\cite{devlin-etal-2019-bert}) as the backbone. We tested different models as backbones but observed no improvement in performance. 

\begin{figure}[htb]
    \centering
    \includegraphics[width=\textwidth]{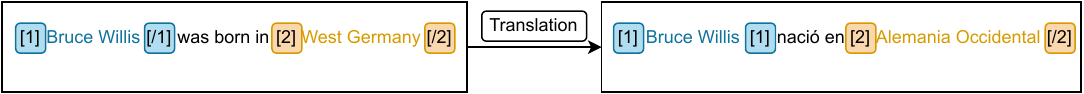}
    \caption{Illustration of the translation with markers approach. Markers are introduced around the labeled sequences. The sentence and the labeled spans are translated together.}
    \label{fig:projection_with_markers}
\end{figure}

We also experiment with \textbf{EasyProject} (\cite{chen-etal-2023-frustratingly}), a system that jointly performs translation and projection by inserting special markers around the labeled spans in the source sentence as depicted in Figure \ref{fig:projection_with_markers}. Additionally, we evaluate \textbf{CODEC} (\cite{DBLP:journals/corr/abs-2402-03131}), a subsequent work that divides the projection process into two distinct steps. In this improved approach, the training data in the high-resource language is first translated without markers. During a second decoding phase, the markers are integrated with the constraint that the translation must align with the initial, marker-free output. This two-step process ensures that the final translated sentence with markers remains consistent with what the model would have produced without them, thereby preserving the translation quality. As both these methods generate their own translations they are therefore not suitable for the intrinsic evaluation which is why we only used them for the extrinsic evaluation. 

\begin{figure}[htb]
    \centering
    \includegraphics[width=.8\linewidth]{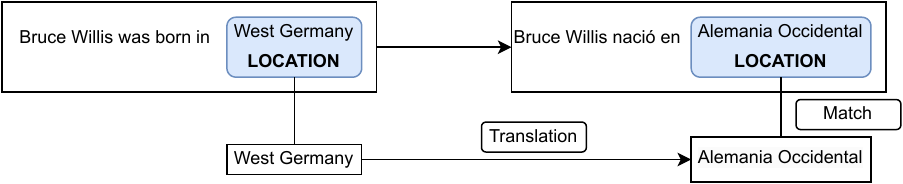}
    \caption{Illustration of the span translation annotation projection approach. The source labels are translated independently, and these translated spans are then matched with their counterparts in the target sentence.}
    \label{fig:projection_translate_match}
\end{figure}

In addition to the previous methods, we also implement two additional baselines inspired by previous works. In the first baseline, inspired by \citet{Li2021CrossLingualNE}, we use the \textbf{XLM-RoBERTa} model (\cite{conneau-etal-2020-unsupervised}) with 3 billion parameters (matching the parameter count of the mT5 model used in T-Projection) and add a token classification layer (linear layer) on top of each token representation. We train the model on the source labeled dataset and use it to predict entities in the translated target sentences.
The second baseline adopts a \textbf{span translation} approach inspired by \cite{DBLP:conf/emnlp/JainPL19} and \citet{DBLP:journals/corr/abs-2211-09394}. In this approach, we translate the labeled spans in the source sentence using the pretrained M2M100 model with 12 billion parameters and then match them with the corresponding spans in the target sentence. For example, as depicted in Figure \ref{fig:projection_translate_match} given the labeled source sentence ``\textit{<Person> Bruce Willis </person> was born in <Location> West Germany </Location>}'' and the target sentence ``\textit{Bruce Willis nació en Alemania Occidental}'', we translate the span \textit{West Germany} into the target language, resulting in \textit{Alemania Occidental}, which is then matched in the target sentence. We employ beam search to generate 100 possible translations and select the most probable one that matches the target sentence.

\subsection{Models Setup}

We use the 3 billion parameters pretrained mT5
(\cite{mt5}) for the \emph{candidate generation} step while \emph{candidates are selected} using the M2M100 12 billion parameter
Machine Translation model (\cite{JMLR:v22:20-1307}). In the case of MasakhaNER, since not all languages are included in M2M100, we resorted to NLLB200 (\cite{DBLP:journals/corr/abs-2207-04672}) 3 billion parameter model instead, which was also used by the EasyProject method (\cite{chen-etal-2023-frustratingly}). Both MT models demonstrate comparable performance. 

We train the HuggingFace's (\cite{DBLP:journals/corr/abs-1910-03771})  implementation of mT5  \footnote{\url{https://huggingface.co/google/mt5-xl}} (3 billion parameter model) in the candidate generation step using the following hyper-parameters: Batch size of 8, 0.0001 learning rate, 256 tokens sequence length, cosine scheduler with 500 warn up steps and no weight decay. We use AdaFactor (\cite{DBLP:conf/icml/ShazeerS18}) optimizer. We train the model for 10 epochs in the OTE task, and 4 epochs for the NER and AM tasks. 
In the candidate selection step, we also use HuggingFace's implementation of M2M100, and we use m2m100-12B-last-ckpt \footnote{\url{https://huggingface.co/facebook/m2m100-12B-last-ckpt}} checkpoint of M2M100 released by the authors. We use the direct-translation function of the NMTscore library to compute the translation probabilities. 
For MasakhaNER2.0 we use the training script and evaluation script developed by the authors \footnote{\url{https://github.com/masakhane-io/masakhane-ner/blob/main/MasakhaNER2.0/scripts/mdeberta.sh}} and the same hyper-parameter setup than \citet{chen-etal-2023-frustratingly}.

\section{Intrinsic Evaluation} \label{sec5:IntrinsicEvaluation}

In this section, we present a set of experiments to evaluate T-Projection in comparison to current state-of-the-art approaches for annotation projection. The intrinsic evaluation focuses on measuring the annotation projection accuracy of the models, isolated from other factors such as the quality of the translation models or any other steps in the pipeline. We also separately analyze the performance of the candidate generation and candidate selection steps.

For the OTE task, we train T-Projection and XLM-RoBERTa using the full English ABSA 2016 dataset. Additionally, we train the four word-alignment systems (excluding SimAlign, which is an unsupervised method) using the English data along with the respective translations as parallel corpora. We augment the parallel data with 50,000 random parallel sentences from ParaCrawl v8 (\cite{espla-etal-2019-paracrawl}). Models are evaluated based on the manual label projections described in Chapter \ref{ch:model-vs-data}.

As the Europarl-based NER dataset (\cite{agerri-etal-2018-building}) provides only test data for each language, T-Projection and XLM-RoBERTa are trained using the full English CoNLL 2003 dataset (\cite{DBLP:conf/conll/SangM03}) together with the labeled English Europarl test data. The word alignment models, in turn, are trained with the parallel sentences from the Europarl-based NER data plus 50,000 parallel sentences extracted from Europarl v8 (\cite{DBLP:conf/mtsummit/Koehn05}). We evaluate the models based on the manual annotations provided by \citet{agerri-etal-2018-building}.

For Argument Mining, we use the full Neoplasm data from the AbstRCT dataset to train T-Projection and XLM-RoBERTa, adding its Spanish translation as parallel corpora for the word alignment systems. As this is a medical text corpus, the parallel corpora are complemented with 50,000 parallel sentences from the WMT19 Biomedical Translation Task (\cite{bawden-etal-2019-findings}). We evaluate the models based on the manually projected labels by \citet{DBLP:journals/corr/abs-2301-10527}.

\subsection{Annotation Projection Quality}
\begin{table}[htb]
    \centering
\adjustbox{max width=\linewidth}{
\begin{tabular}{@{}l|ccc|ccc|c|c@{}}
\toprule
 & \multicolumn{3}{c}{OTE} & \multicolumn{3}{c}{NER} & \multicolumn{1}{c}{AM} &  \multicolumn{1}{c}{Avg} \\ \midrule
 &  ES &  FR & RU &  ES &  DE &  IT &  ES &  \\ \midrule
Giza++ (\cite{och-ney-2003-systematic-giza}) & 77.0 & 73.3 & 72.4 & 73.3 & 75.3 & 68.4 & 86.6 & 77.7 \\
FastAlign (\cite{dyer-etal-2013-simple-fastalign}) & 75.0 & 72.9 & 76.9 & 70.2 & 77.0 & 67.0 & 85.7 & 77.4 \\
SimAlign (\cite{jalili-sabet-etal-2020-simalign}) & 86.7 & 86.3 & 87.7 & 85.4 & 87.4 & 81.3 & 84.1 & 85.3 \\
AWESOME (\cite{dou-neubig-2021-word}) & 91.5 & 91.1 & 93.7 & 87.3 & 90.7 & 83.1 & 54.8 & 78.0 \\ \midrule
XLM-RoBERTa-xl (\cite{conneau-etal-2020-unsupervised}) & 80.2 & 76.2 & 74.5 & 73.9 & 68.3 & 73.9 & 66.5 & 71.8 \\
Span Translation & 66.5 & 46.3 & 58.7 & 68.8 & 63.5 & 69.2 & 21.6 & 48.7 \\ \midrule
T-Projection & \textbf{95.1} & \textbf{92.3} & \textbf{95.0} & \textbf{93.6} & \textbf{94.0} & \textbf{87.2} & \textbf{96.0} & \textbf{93.9} \\ \bottomrule
\end{tabular}
}
    \caption{F1 scores for annotation projection in the OTE, NER and Argument Mining tasks.}
    \label{tab5:IntrinsicResults}
\end{table}

Table \ref{tab5:IntrinsicResults} presents the results of the automatically projected datasets generated by each projection method, compared to the human-projected versions of those datasets. Systems using word alignments achieve consistently good results, particularly those utilizing language models such as SimAlign and AWESOME. Specifically, AWESOME performs well in OTE and NER but poorly in AM. Manual inspection reveals that AWESOME struggles to align articles and prepositions within long sequences. The statistical-based models, Giza++ and FastAlign, achieve competitive performance considering their very low computational resource requirements (they do not require a GPU) compared to the Transformer-based approaches.

XLM-RoBERTa-xl demonstrates strong zero-shot cross-lingual performance. However, the quality of the generated datasets is lower than those produced by the word-alignment systems. The results of the Span Translation approach are disappointing, particularly for the long sequences in the AM task. Translating the labeled spans independently often results in translations that cannot be located in the target sentence.

Our T-Projection method achieves the best results for every task and language. In OTE, it outperforms all other methods by more than 2 points in F1 score averaged across the three languages. This indicates that T-Projection robustly projects labeled spans into machine-translated data. The NER evaluation is slightly different because the parallel data was translated by human experts. In this context, T-Projection significantly improves AWESOME's results by 4.7 points, marking a substantial improvement in the quality of the generated datasets. Despite the word alignment systems being trained with Europarl domain-specific data and most of the training data for T-Projection coming from the CoNLL-2003 dataset (news domain) plus a few annotated sentences (699) from Europarl, T-Projection still achieves the best results in NER label projection. This suggests that our system is effective even in out-of-domain settings. 

Furthermore, T-Projection achieves the highest overall scores in Argument Mining, demonstrating its exceptional ability to project long sequences. T-Projection outperforms the second-best model by 9.4 points in F1 score, with a by 96.0 points in F1-scores, indicating near-perfect projection of all examples in the dataset.

When considering the average performance across the three tasks and five languages, T-Projection improves the F1 score by 8.6 points compared to the second-best system, SimAlign. These results represent a significant advancement over all previous annotation projection approaches. To the best of our knowledge, these are by a wide margin the best annotation projection results published for sequence labeling.

\subsection{The Role of the Candidates} \label{sec5:RoleCandidates}

\begin{table}[htb]
    \centering
    \adjustbox{max width=\linewidth}{
        \begin{tabular}{@{}l|ccc|ccc|c|c@{}}
            \toprule
             & \multicolumn{3}{c|}{OTE} & \multicolumn{3}{c|}{NER} & \multicolumn{1}{c|}{AM} & \multicolumn{1}{c}{Avg} \\ \midrule

             & ES & FR & RU & ES & DE & IT & ES & \\ \midrule
            T-Projection & 95.1 & 92.3 & 95.0 & 93.6 & 94.0 & 87.2 & 96.0 & 93.9 \\ \midrule
            Ngrams + Candidate Selection & 89.7 & 86.1 & 93.8 & 83.8 & 79.3 & 73.3 & 73.5 & 80.7 \\ 
            mT5 + Most Probable Candidate & 83.7 & 87.2 & 85.3 & 79.5 & 82.8 & 72.3 & 90.9 & 84.8 \\ 
            mT5 + Upper bound & 98.6 & 97.0 & 97.9 & 98.0 & 98.5 & 94.0 & 99.3 & 98.0 \\
            \bottomrule
        \end{tabular}
    }
    \caption{F1 scores for different candidate generation and candidate selection methods.}
    \label{tab5:CandidateResults}
\end{table}

We perform a set of experiments to measure the relevance and performance of the \emph{candidate generation} and \emph{candidate selection} steps. First, we replace mT5 as the candidate generation model with an n-gram-based approach. We extract all the n-grams with sizes ranging from 1 to the sentence length (e.g., \textit{``Serve'', ``really'', ``good'', ``sushi'', ``Serves really'' ... ``Serves really good sushi''}) and consider them as candidates. Then we rank the candidates using the translation probabilities obtained by the M2M100 model. As shown in Table \ref{tab5:CandidateResults}, the n-gram-based approach's performance is significantly lower than T-Projection in all tasks and languages. This indicates that the mT5 model is crucial for generating relevant candidates. The n-gram approach generates a large number of very similar candidates, which makes it difficult for the M2M100 model to select the correct one. These experiments demonstrate the importance of the mT5 model in generating relevant candidates.

We also replace the \emph{candidate selection} method with the \emph{most probable candidate}. That is, we select the most probable candidate generated by mT5 for each labeled span in the source sentence, thus we do not consider the translation probabilities, as only one candidate is generated for each labeled span. This approach achieves competitive results with the word alignment systems in Table \ref{tab5:IntrinsicResults}, but it is outperformed by T-Projection by an average of 9.2 points in F1-Score.

This ablation study demonstrates that although each step of T-Projection in isolation is able to achieve competitive results on its own, it is the combination of both steps that allows T-Projection to achieve very high performance in the intrinsic evaluation.

Finally, we define an upper bound for the \emph{candidate selection} step. This upper bound is defined by always selecting the correct candidate among the generated candidates. If the correct candidate is not generated by mT5, we select the most probable candidate. This upper bound achieves an average F1 score of 98.0. This result confirms that with a very high probability, the correct candidate is among the candidates generated by mT5.

\subsection{How many candidates are necessary?} \label{sec5:Howmany}

\begin{figure}[htb]
    \centering
    \includegraphics[width=0.80\linewidth]{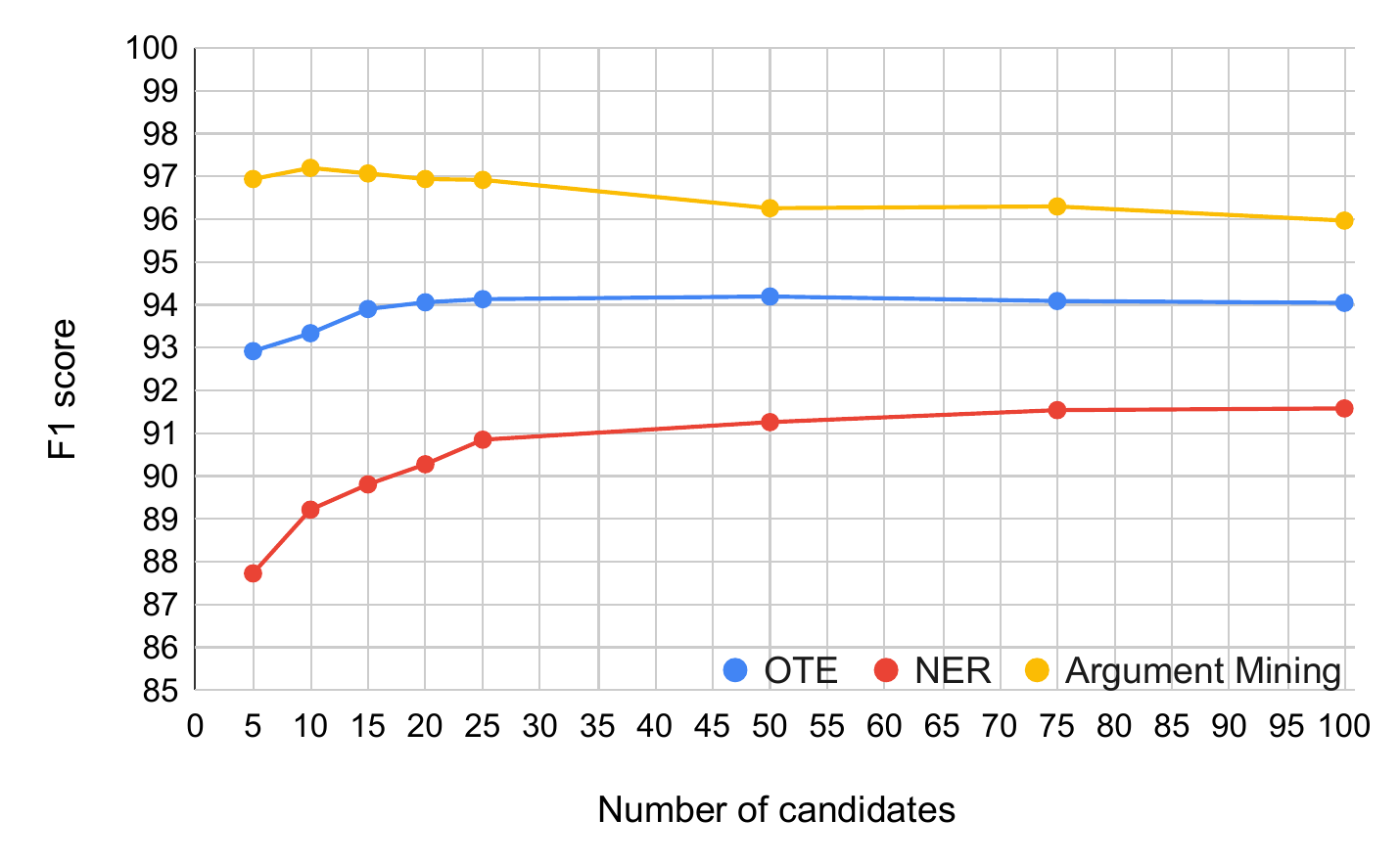}
    \caption{F1 score when generating a different number of candidates.}
    \label{fig5:CandidateNo}
\end{figure}

Generating candidates is computationally expensive. The number of FLOPs, memory usage, and inference time increase linearly with the number of candidates. Thus, generating 20 candidates is twice as expensive as generating 10 candidates. Additionally, we must consider the extra cost of computing similarity scores for each candidate. Therefore, we performed an experiment to determine the optimal number of candidates to generate. As shown in Figure \ref{fig5:CandidateNo}, the performance of T-Projection increases with the number of candidates. However, the performance improvement diminishes as the number of candidates grows. For OTE and NER, the improvement is negligible after generating 25 candidates. For AM, generating more than 10 candidates is, in fact, counterproductive. While the reported results in the paper were obtained by generating 100 candidates, a decision informed by preliminary studies using the upper bound described in Section \ref{sec5:RoleCandidates}, the results in Figure \ref{fig5:CandidateNo} suggest that generating 10 to 25 candidates is optimal.

\begin{figure}[htb]
    \centering
    \includegraphics[width=0.80\linewidth]{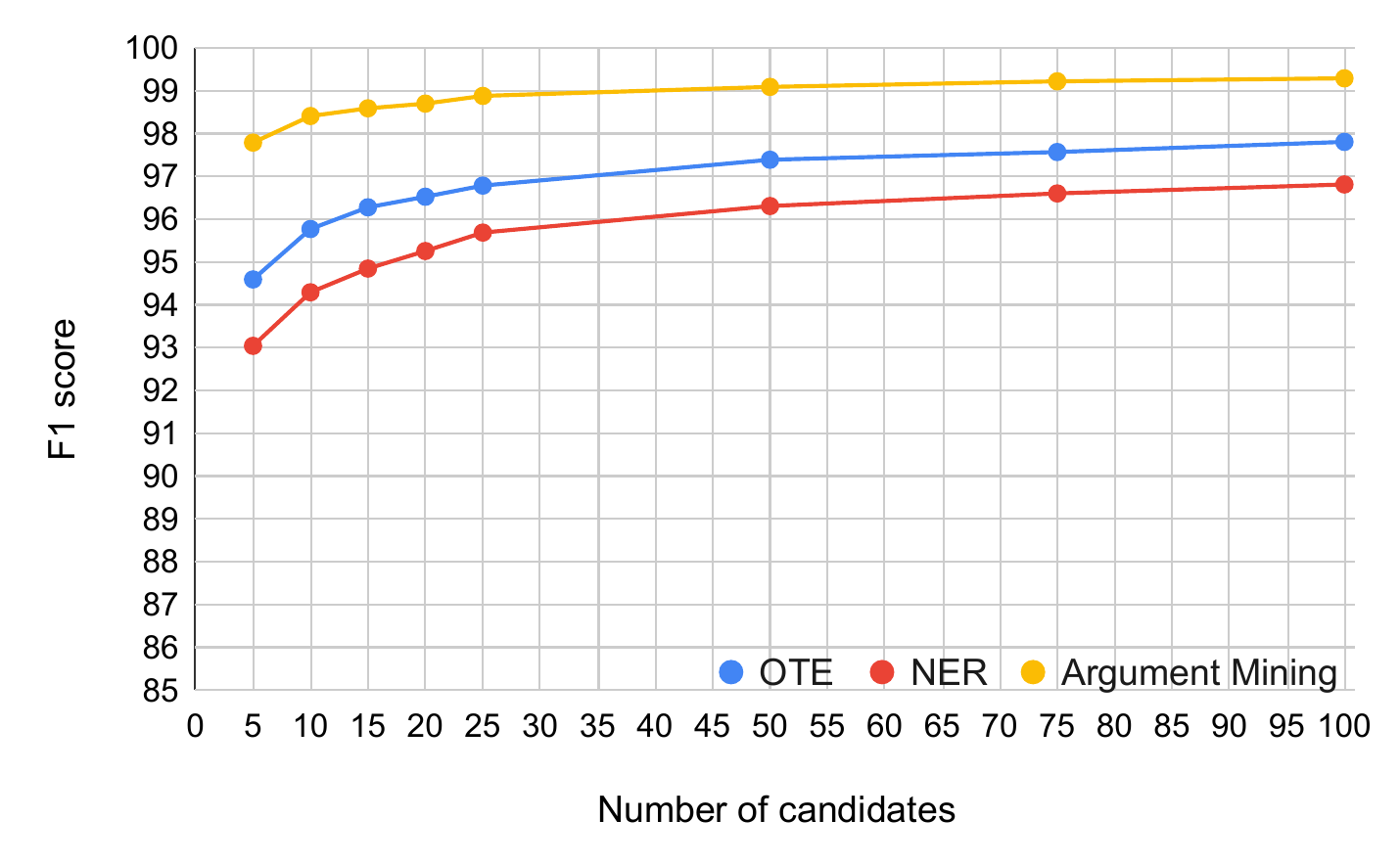}
    \caption{Number of times the correct candidate is among the top-k candidates generated by mT5.}
    \label{fig5:CandidateNoUpperbound}
\end{figure}

Generating a larger number of candidates is not beneficial for the performance of T-Projection. To further understand why this happens, we computed the F1 score of T-Projection following the upper bound described in Section \ref{sec5:RoleCandidates} for different number of candidates. This is, we compute how many times the correct candidate is among the top-k candidates generated by mT5. As shown in Figure \ref{fig5:CandidateNoUpperbound}, as we increase the number of candidates, the probability of the correct candidate being among the top-k candidates also increases. However, the improvement is not linear. It becomes less significant after generating 25 candidates. The performance improvement from going from 25 to 100 candidates is only 1 point in F1 score for OTE and NER and less than 0.5 points for AM. Interestingly, for AM, the best candidate is among the top-25 candidates 99\% of the time, which explains the great performance of T-Projection in this task. However, if the best candidate is found with a very high probability among the top-25 candidates, what are the candidates beyond the top-25? By examining the candidates generated by mT5 in the NER tasks, we found that only a few of the generated candidates are valid. The remaining ones are hallucinated spans that do not exist in the sentence and are therefore filtered out. These hallucinated spans are usually variations of the correct candidate. We found that there are fewer than 20 valid candidates per sentence, with an average of 5.95 valid candidates per sentence for NER. Thus, generating more than 25 candidates is not beneficial, but it also does not introduce noise that severely hinders the performance of T-Projection, as the extra candidates are usually hallucinated spans that are filtered out in the candidate selection step, rather than different n-grams from the target sentence.

\subsection{Model size and performance}

\begin{table}[htb]
    \centering
 \adjustbox{max width=0.8\linewidth}{
\begin{tabular}{@{}llccccc@{}}
\toprule
& Model & \#Params & OTE & NER & AM & Average \\ \midrule
\multirow{3}{*}{ MT Size } & m2m100 & 418M & 92.3 & 91.7 & 95.5 & 93.1 \\
& m2m100 & 1.2B & 94.0 & \textbf{92.0} & 95.8 & \textbf{93.9} \\
& m2m100 & 12B & 94.1 & 91.6 & 96.0 & \textbf{93.9} \\ 
& Prism & 745M & 94.1 & 90.9 & \textbf{96.3} & 92.7 \\
& nllb200 & 3B & \textbf{94.2} & 91.0 & 93.0 & 93.0 \\ 
\midrule
\multirow{4}{*}{ mT5 size } & mT5-small & 60M & 36.4 & 66.3 & 00.0 & 34.2 \\
& mT5-base & 220M & 72.8 & 86.2 & 33.6 & 64.2 \\
& mT5-large & 738M & 90.9 & 90.1 & 65.3 & 82.1 \\
& mT5-xl & 3B & \textbf{94.1} & \textbf{91.6} & \textbf{96.0} & \textbf{93.9} \\
\bottomrule
\end{tabular}
}
    \caption{F1 scores of T-Projection when using translation and mT5 models of different size}
    \label{tab5:ModelSize}
\end{table} 

We analyze the performance of T-Projection using an mT5 model and a translation system with different numbers of parameters. Additionally, we evaluate T-Projection with various Machine Translation models, namely, NLLB200 (\cite{DBLP:journals/corr/abs-2207-04672}) and PRISM (\cite{DBLP:conf/emnlp/ThompsonP20}). Table \ref{tab5:ModelSize} demonstrates that the Machine Translation system and its parameter count do not significantly impact T-Projection's performance.

However, the size of the mT5 model has a substantial impact on the system's final performance. While switching from a 3B to a 738M parameter mT5 model yields competitive results for OTE and NER, this is not the case for AM. The overall trend indicates that decreasing the number of parameters results in decreased performance. In summary, to achieve competitive performance across all tasks, T-Projection requires an mT5 model with 3B parameters, although a 738M parameter model remains competitive for OTE and NER.

\section{Extrinsic Evaluation}

\begin{table}[htb]
    \adjustbox{max width=\linewidth}{
   \begin{tabular}{@{}lllcccccc@{}}
   \toprule
   Language & No. of & \multicolumn{1}{c}{Language} & \multicolumn{1}{c}{Finetune} & \multicolumn{1}{c}{AWESOME} & \multicolumn{1}{c}{EasyProject} & \multicolumn{1}{c}{CODEC} & \multicolumn{1}{c}{T-Projection} & \multicolumn{1}{c}{T-Projection} \\
   & Speakers & \multicolumn{1}{c}{family} &  English & +English & +English &  &  & +English \\ \midrule
   Hausa & 63M & Afro-Asiatic /Chadic & 71.7 & \textbf{72.7} & 72.2 & 72.4 & \textbf{72.7} & 72.0 \\
   Igbo & 27M & NC / Volta-Niger & 59.3 & 63.5 & 65.6 & 70.9 & 71.4 & \textbf{71.6} \\
   Chichewa & 14M & English-Creole & \textbf{79.5} & 75.1 & 75.3 & 76.8 & 77.2 & 77.8 \\
   chiShona & 12M & NC / Bantu & 35.2 & 69.5 & 55.9 & 72.4 & \textbf{74.9} & 74.3 \\
   Kiswahili & 98M & NC / Bantu & \textbf{87.7} & 82.4 & 83.6 & 83.1 & 84.5 & 84.1 \\
   isiXhosa & 9M & NC / Bantu & 24.0 & 61.7 & 71.1 & 70.4  & \textbf{72.3} & 71.7 \\
   Yoruba & 42M & NC / Volta-Niger & 36.0 & 38.1 & 36.8 & 41.4 & \textbf{42.7} & 42.1 \\
   isiZulu & 27M & NC / Bantu & 43.9 & 68.9 & 73.0 & \textbf{74.8} & 66.7 & 64.9 \\ \midrule
   AVG &  &  & 54.7 & 66.5 & 66.7 & \textbf{70.3} & \textbf{70.3} & 69.8 \\ \bottomrule
   \end{tabular}
   }
   \caption{F1 scores on MasakhaNER2.0 for mDebertaV3 trained with projected annotations from different systems. "+EN" denotes concatenation of the automatically generated target language dataset with the source English dataset.}
   \label{tab5:MasakhaNER2}
   \end{table}

In this section, we evaluate T-Projection in a real-world low-resource scenario, namely, Named Entity Recognition for African languages. We compare the results obtained by training on NER datasets automatically generated by T-Projection with those automatically projected using three state-of-the-art label projection systems: AWESOME (the second-best NER system in Table \ref{tab5:IntrinsicResults}), EasyProject, and CODEC. We use the exact same settings as \citet{chen-etal-2023-frustratingly} and \cite{DBLP:journals/corr/abs-2402-03131}. For each target language in MasakhaNER2.0, we first translate the English CoNLL dataset using the NLLB-200 3 billion parameter model. Next, we project the English labels into the target language. It should be noted that EasyProject performs both of these processes in a single step. Subsequently, we train an mDebertaV3 (\cite{DBLP:conf/iclr/HeLGC21/deberta}) model using the automatically generated datasets for each target language. Finally, this model is evaluated on the gold MasakhaNER2.0 test data. We only evaluate the eight languages in MasakhaNER2.0 supported by mT5. We focus on named entities referring to Person, Location, and Organization. We also evaluate the zero-shot model-transfer approach presented in Chapter \ref{ch:model-vs-data}. That is, we train the mDebertaV3 model with the original English CoNLL data and evaluate it on the MasakhaNER2.0 test sets.

Table~\ref{tab5:MasakhaNER2} presents the results of the evaluated models on the gold MasakhaNER 2.0 test sets. For T-Projection, we present the results of training with the automatically generated data for the target language only, and also by adding the original English CoNLL data concatenated with the automatically generated data for each target language. Regarding other systems, we only show the former results, as it was the only metric reported by previous work. In order to train and evaluate the NER models, we apply the same hyperparameter settings and code as the authors of EasyProject.

\subsection{T-Projection vs other annotation projection systems}

The results show that T-Projection achieves superior performance for seven out of the eight languages compared to other annotation projection systems. Our model demonstrates a more pronounced performance difference in agglutinative languages such as Igbo and Shona. As outlined in Section \ref{sec5:IntrinsicEvaluation}, our model produces superior alignments compared to AWESOME. Furthermore, we found that EasyProject, which utilizes markers for simultaneous translation and projection, introduces translation artifacts that hinder the performance of the downstream model. These artifacts are particularly noticeable in agglutinative languages, as EasyProject tends to separate words. For instance, in the case of Shona, consider the English sentence \textit{``[Germany]'s representative to the [European Union]'s veterinary committee [Werner Zwingmann]''}. Our system produces the Shona sentence \textit{``Mumiriri [weGermany] kukomiti yemhuka [yeEuropean Union] [Werner Zwingmann]''}, while EasyProject produces \textit{``Mumiriri we [Germany] ku [European Union] komiti yezvokurapa mhuka [Werner Zwingmann]''}. When training mDeberta-v3 with T-Projection's generated data, which preserves the agglutinated words, we achieve better results compared to EasyProject which introduces artifacts by separating agglutinated words during translation and projection. CODEC improves over EasyProject, the two step approach, in which first the translation is performed and then the markers are added, helps to preserve the translation quality. However, the performance of CODEC is still lower than T-Projection for every language except for Zulu. 

\subsection{T-Projection vs Model-transfer}

Table \ref{tab5:MasakhaNER2} presents the results of training the multilingual model mDebertaV3 with the original English CoNLL data and conducting zero-shot evaluation on the MasakhaNER2.0 test sets. The findings indicate that T-Projection outperforms this baseline in six out of eight languages, with an average improvement of 15.6 F1 points. This contrasts with the results from Chapter \ref{ch:model-vs-data}, where model-based transfer learning surpassed data-based approaches. The discrepancy can be attributed to two factors. 
First, the data generated by T-Projection is of higher quality than the data generated by the word alignment systems used in Chapter \ref{ch:model-vs-data}. On average, T-Projection is 8.6 F1 score points better than the best word-alignment system in intrinsic evaluation and 3.8 F1 score points superior in extrinsic evaluation.

Second, the zero-shot evaluation in Chapter \ref{ch:model-vs-data} was conducted from English to other high-resource languages, whereas in this chapter, we perform cross-lingual evaluation into African low-resource languages. These low-resource languages have significantly different morphology and syntax compared to English, and multilingual models typically have lower proficiency in these languages. As demonstrated in Chapter \ref{ch:model-vs-data}, model-based transfer performance requires a multilingual model with high proficiency in both the source and target languages.

These results demonstrate that data-based transfer approaches, such as T\-Projection, can be highly effective for performing Natural Language Processing tasks in low-resource languages, especially in the absence of a high-proficiency multilingual model.

In contrast to previous work, our experiments revealed that concatenating English and translated data did not yield better results, likely due to the superior quality of the data generated by T-Projection. To the best of our knowledge, these are the best zero-shot results achieved for MasakhaNER2.0, highlighting the significant benefits of T-Projection for NLP tasks in low-resource languages.

\section{Conclusions}

In this section we have introduced T-Projection, a novel method for projecting labeled sequences across languages. T-Projection leverages the zero-shot cross-lingual capabilities of large pretrained multilingual language models to generate candidates for each labeled span in the source sentence. These candidates are then ranked using a Machine Translation model to select the best projection candidate for each labeled span in the target sentence.

We have demonstrated that T-Projection outperforms current state-of-the-art label projection systems in both intrinsic and extrinsic evaluations. In the intrinsic evaluation, T-Projection achieves the best results for every task and language, improving the F1 score by 8.6 points compared to the second-best system. In the extrinsic evaluation, T-Projection achieves superior performance in seven out of the eight languages in the MasakhaNER2.0 dataset. These results underscore the effectiveness of T-Projection in generating high-quality training data for sequence labeling tasks in low-resource languages.

Moreover, T-Projection surpasses model-based cross-lingual transfer in the extrinsic evaluation, specifically for named entity recognition in African low-resource languages. While model-based transfer learning outperformed data-based approaches in Chapter \ref{ch:model-vs-data}, it is less effective for cross-lingual transfer from English into the African languages tested in this chapter. In this scenario, T-Projection generates high-quality training data that significantly improves the performance of the downstream model. This demonstrates the potential of data-based transfer approaches for NLP tasks in low-resource languages.

\selectlanguage{english}
\chapter[Improving Model Transfer]{Improving Model Transfer}
\label{ch:model-transfer}

In this chapter we will focus on zero-shot model transfer. In Chapter \ref{ch:model-vs-data} we demonstrated that model transfer could be an effective and efficient approach for cross-lingual transfer when using a high-capacity model on the target language. Therefore, for this approach to be effective, it is crucial to use the most powerful models available. Currently, these models are the text-to-text Large Language Models (LLMs). However, using LLMs for zero-shot cross-lingual sequence labeling is not straightforward. In this chapter we will introduce a constrained decoding algorithm that effectively addresses this issue. A comprehensive empirical evaluation across multiple tasks and languages demonstrates that, when our method is applied to an LLM, it helps not only to improve over the unconstrained beam search baseline but also to outperform the zero-shot cross-lingual capabilities of encoder-only models, especially for languages that significantly differ from English.

\section{Motivation and contributions}
\label{sc6:intro}

In Chapter \ref{ch:model-vs-data}, we demonstrated that the performance of zero-shot cross-lingual transfer can be significantly enhanced by using a high-capacity model for the target language. However, we only employed encoder-only models such as XLM-RoBERTa-large (\cite{conneau-etal-2020-unsupervised}), which has 561 million parameters and was trained on approximately 295 billion tokens. However, as mentioned in Chapter \ref{sc:deep-learning-sota}, the most powerful models currently available are text-to-text Large Language Models (LLMs) like T5 (\cite{DBLP:journals/jmlr/RaffelSRLNMZLL20-T5}), LLaMA (\cite{llama3modelcard}), and GPT-4 (\cite{openai2024gpt4technicalreport}). These models have demonstrated superior capabilities in a wide range of NLP tasks, including the ability to solve tasks for which they were not explicitly trained. Consequently, efforts to scale NLP models have primarily focused on text generation models. As shown in Table \ref{tab:model-size}, the latest generation of LLMs have significantly more parameters and were trained on much larger datasets compared to XLM-RoBERTa.

\begin{table}[htb]
    \centering
    \small
    \adjustbox{max width=\textwidth}{
    \begin{tabular}{@{}lcccccc@{}}
    \toprule
     & XLM-RoBERTa  & XLM-RoBERTa-xxl    & mT5  & Llama2  & Gemma2  & LLama3  \\ 
     & \cite{conneau-etal-2020-unsupervised} & \cite{DBLP:journals/corr/abs-2105-00572} & \cite{mt5} & \cite{DBLP:journals/corr/abs-2307-09288} & \cite{DBLP:journals/corr/abs-2403-08295-gemma} & \cite{llama3modelcard} \\ \midrule
    Parameters  & 560M & 10.7B  & 11.3B & 70B & 27B & 405B \\
    Train Tokens & 296B & 296B & 1T & 2T & 8T & 17T \\ \bottomrule
    \end{tabular}
    }
    \caption{Size and training data of some relevant open source models.}
    \label{tab:model-size}
    \end{table}

LLM models have already been proven effective for Information Extraction and sequence labeling tasks in monolingual evaluations in English (\cite{DBLP:journals/corr/abs-2305-15444, sainz2024gollie}). However, their performance still lags behind encoder-only models in multilingual sequence labeling (\cite{DBLP:conf/semeval/FetahuKCRM23}). For low-resource languages, such as African languages, \cite{DBLP:conf/africanlp/OjoO23} demonstrated that most text-to-text LLMs for named entity recognition do not perform well at all when evaluated in a zero-shot setting. Their results are reproduced in Table \ref{tab6:howgoodafrican}. The table shows that the performance of LLMs is significantly lower than that of XLM-RoBERTa-large. This is a surprising result, given that many of the LLMs have been trained on much larger multilingual datasets and have significantly more parameters than XLM-RoBERTa-large. Thus, the question arises: why do LLMs perform poorly in zero-shot cross-lingual sequence labeling tasks?

\definecolor{Color}{gray}{0.9}
\begin{table}[htb]
    \begin{center}
    \resizebox{\textwidth}{!}{
    \begin{tabular}{llrrrrrrrrrrrrrrrrrrrr}
    \toprule
    \textbf{Model} &\textbf{Size} &\textbf{amh} &\textbf{bam}  &\textbf{bbj} &\textbf{ewe} &\textbf{hau} &\textbf{ibo} &\textbf{kin} &\textbf{lug} &\textbf{luo} &\textbf{mos} &\textbf{nya} &\textbf{pcm} &\textbf{sna} &\textbf{swa} &\textbf{tsn} &\textbf{twi} &\textbf{wol} &\textbf{xho} &\textbf{yor} &\textbf{zul} \\
    \midrule
    \multicolumn{2}{l}{\texttt{Fine-tune: SotA}} \\
    \rowcolor{Color}
    AfroXLMR-large & 550M & \textbf{78.0}  &\textbf{79.0} &\textbf{90.3} &75.2 &\textbf{85.4} &\textbf{88.9} &\textbf{86.8} &\textbf{88.9} &\textbf{75.3} &\textbf{73.5} &\textbf{92.4} &\textbf{90.0} &\textbf{96.1} &\textbf{92.7} &\textbf{88.9} &\textbf{79.2} &\textbf{83.8} &\textbf{89.2} &\textbf{67.9} &\textbf{90.6} \\
    \midrule
    \multicolumn{2}{l}{\texttt{Prompting of LLMs}} \\
    GPT-4 & - &28.5  &52.7 &50.3 &\textbf{75.6} &64.9 &56.0 &55.1 &73.3 &49.8 &60.2 &63.6 &64.7 &33.4 &71.5 &64.6 &58.6 &67.9 &28.4 &58.3 &34.9 \\
    AYA & - &14.1 &7.1 &20.0 &26.5 &34.5 &28.2 &30.8 &16.3 &12.7 &34.4 &21.7 &27.4 &13.4\definecolor{Color}{gray}{0.9} &35.6 &29.4 &18.9 &14.5 &4.2 &17.5 &11.4  \\
    mT0 & 13B &0.0 &0.0 &0.0 &0.1 &0.0 &0.0 &0.0 &0.0 &0.0 &0.0 &0.0 &0.0 &0.0 &0.0 &0.0 &0.0 &0.0 &0.0 &0.0 &0.0  \\
    mT0-MT & 13B &0.0 &0.0 &0.0 &0.1 &0.0 &0.0 &0.0 &0.0 &0.0 &0.0 &0.0 &0.0 &0.0 &0.0 &0.0 &0.0 &0.0 &0.0 &0.0 &0.0  \\
    LLaMa 2 & 13B &0.0 &13.8 &12.3 &25.1 &22.1 &22.0 &23.1 &27.5 &19.0 &11.0 &20.0 &27.5 &11.3 &25.8 &26.2 &20.7 &16.0 &8.1 &15.1 &9.0  \\
    \bottomrule
    \end{tabular}
    }
    \caption{Comparison of F1-score of various LLMs with that of the current state of the art result in Masakhaner 2.0. Table reproduced from \cite{DBLP:conf/africanlp/OjoO23}.
    }
    \label{tab6:howgoodafrican}
    \end{center}

    \end{table}

\begin{figure}[htb]
    \centering
    \includegraphics[width=\textwidth]{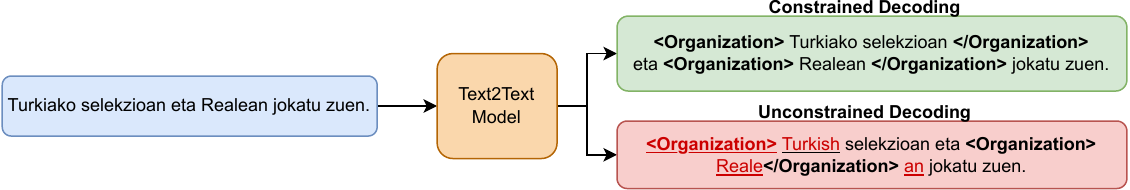}
    \caption{Comparison between a valid (top green) and invalid (bottom red) output structure to represent a Named Entity Recognition task. English translation: (They) played in Real and in the Turkish national team.}
    \label{fig6:constrained_unconstrained}
\end{figure}

In this chapter, we investigate the performance of LLMs in zero-shot cross-lingual sequence labeling tasks. Our contributions are as follows:

\paragraph{We identify the challenges faced by text-to-text models for zero-shot sequence labeling:} In this setting we must first establish a text-based input and output representation for the specific task. However, current text-to-text models are tailored for generating free-form text. As our experiments demonstrate, models fail to strictly adhere to the output structure. Moreover, as demonstrated by our experiments, text-to-text models often produce outputs mixing the source language and the target language, which compromises their performance. These issues are illustrated in Figure \ref{fig6:constrained_unconstrained}, where the incorrect output mixes English and Basque (Turkiako-Turkish) and incorrectly segments the organization entity ``Realean''.

\paragraph{We propose a constrained decoding algorithm for text-to-text models:} We introduce a constrained decoding algorithm that enforces the output structure of the target task. Our method can be seamlessly integrated with any text-to-text model without any significant increase in the decoding cost. Although constrained generation has been previously explored in a monolingual setting (\cite{liu-etal-2022-autoregressive}), we adapt and extend this approach for zero-shot cross-lingual IE. Our new decoding algorithm is evaluated on three popular IE tasks for 25 languages of varied morphological characteristics. Empirical results indicate that our method, when applied to an LLM such as mT0-XL (\cite{DBLP:conf/acl/MuennighoffWSRB23}), not only surpasses the unconstrained beam search baseline but also outperforms the zero-shot cross-lingual performance of encoder-only models. Our method is especially successful for languages that significantly differ from English.

To the best of our knowledge, our new technique achieves the best zero-shot model-based cross-lingual transfer results to date.

\section{Related Work}
In this section, we focus first on related work concerning large language models (LLMs) for sequence labeling. In the second part of the related work, we review prior research on constrained decoding.

\subsection{LLMs for sequence labeling}

The introduction of models like T5 (\cite{DBLP:journals/jmlr/RaffelSRLNMZLL20-T5}) and GPT (\cite{radford2019language}) revolutionized NLP by adopting a text-to-text approach, enabling models to handle a wide array of tasks with a single training objective. Consequently, all NLP tasks can be framed as text-to-text tasks, where the input is a description of the task or a prompt, and the output is the desired result (\cite{chung-flan-instruction-models}). Scaling these models in both the amount of training data and the number of parameters, has led to the development of state-of-the-art models, such as GPT-4 (\cite{openai2024gpt4technicalreport}), LLaMA (\cite{llama3modelcard}), and Mistral (\cite{jiang2023mistral7b}). These models achieve state-of-the-art results on a broad range of NLP tasks (\cite{DBLP:journals/csur/MinRSVNSAHR24}), including those they were not explicitly trained for (\cite{radford2019language}).

In the field of Information Extraction (IE), the text-to-text approach has also been explored. \cite{DBLP:conf/acl/0001LDXLHSW22} introduced a unified text-to-structure generation model capable of handling various IE tasks universally. \cite{DBLP:conf/aaai/Lou0DJLH0023} proposed converting IE tasks into a semantic matching problem, allowing their method to generalize to new domains and label ontologies not encountered during training. \cite{DBLP:journals/corr/abs-2304-08085} framed IE tasks as natural language descriptive instructions and trained a large language model (LLM) across a diverse range of IE tasks. In evaluations involving tasks with unseen label ontologies, their model outperformed other instruction-tuning methods. More recently, \cite{DBLP:conf/acl/BlevinsGZ23} and \cite{sainz2024gollie} proposed using complex instructions that include annotation guidelines, similar to the ones used by human annotators, to enhance the performance of LLMs in sequence labeling tasks. This approach has proven effective in achieving strong performance in classifying unseen categories in sequence labeling tasks in English.

While success has been achieved in labeling unseen categories in English, the supervised performance of LLMs, when training data is available, is still not superior to that of smaller encoder-only models (\cite{sainz2024gollie}). Additionally, recent shared tasks (\citep{DBLP:conf/semeval/FetahuKCRM23}) have shown that for languages other than English, encoder-only language models such as XLM-RoBERTa (\cite{conneau-etal-2020-unsupervised}) and mDEBERTA (\cite{DBLP:conf/iclr/HeLGC21/deberta}) remain the most effective models.

\subsection{Constrained decoding}

The formulation of information extraction tasks in a constrained text-to-text format has been previously explored (\cite{DBLP:conf/nips/VinyalsKKPSH15,DBLP:conf/acl/XiaoDG16,DBLP:conf/naacl/DyerKBS16}). However, it was with the emergence of large-scale text-to-text language models, that this approach garnered significant attention within the community. \citet{DBLP:conf/emnlp/LesterPHCB20} propose a Named Entity Recognition system that uses Viterbi decoding (\cite{viterbi}) with heuristically determined transition probabilities that prohibit illegal transitions. This achieves similar performance to the conditional random field (CRF) models (\cite{DBLP:conf/icml/LaffertyMP01}), but it is more computationally efficient. \citet{genre} and \citet{mgenre} propose a sequence-to-sequence system for Multilingual Entity Linking, which can generate entity names from left to right, token by token, in an autoregressive manner, conditioned by the context. To ensure that only valid entity identifiers are generated, they employ a prefix tree to enable constrained beam search. 

Closer to our work, which focuses on constraining large language models (LLMs) to adhere to a pre-defined output structure, \citet{lu-etal-2021-text2event} presents a constrained decoding algorithm that ensures the model adheres to a specified output structure during inference. Similarly, \citet{zheng-etal-2023-grammar} and \citet{DBLP:journals/corr/abs-2302-02275} propose constrained decoding algorithms that enhance semantic parsing. Instead of constraining the generation of output text, \citet{cui-etal-2021-template} perform Named Entity Recognition (NER) by computing the probability of a text span filling predefined structures. Rather than flattening the structured output into a sequence, \citet{liu-etal-2022-autoregressive} model the output as sequences of actions. These actions are predicted in an autoregressive manner using LLMs, and executing the actions generates the structured output. Their approach improves upon previous methods in NER, end-to-end relation extraction, and co-reference resolution. \cite{DBLP:conf/emnlp/GengJP023} demonstrate that grammar-constrained decoding (GDC) can significantly enhance the performance of large language models (LMs) across a variety of structured NLP tasks, such as information extraction, entity disambiguation, and constituency parsing, by ensuring outputs adhere to a given structure. GCD-enhanced LMs outperform both unconstrained LMs and task-specific finetuned models, particularly in scenarios with limited training data.

Although previous research has demonstrated the effectiveness of constrained decoding for information extraction,  most of it has focused on monolingual settings. Thus, \citet{DBLP:conf/acl/GuoR21} propose an algorithm that employs constrained decoding of text-to-text LLMs for zero-shot NER in low-resource languages. First, they translate labeled data in a word-by-word manner using a dictionary. Then, they construct target language text from the source-language named entities using a pretrained language model. They utilize constrained decoding to ensure the presence of entities in the generated text. This data-transfer method was later surpassed by model-based cross-lingual transfer methods as we demonstrate in Chapter \ref{ch:model-vs-data}.

To project labels across languages in sequence labeling tasks, \citet{DBLP:journals/corr/abs-2402-03131} introduce markers to the input text to represent the labeled sequences. They then translate the text into the target language, achieving both translation and annotation projection. To prevent translation artifacts caused by the markers, they propose a constrained decoding algorithm that ensures the output of the translation when markers are introduced, remains consistent with translations without markers. Although this method is effective, it is a data-based approach that requires training a new model on the projected data. 

\section{Approach}\label{sec6:Approach}

In this section, we describe our representation of a Sequence Labelling task by applying our new Constrained text-to-text approach. Our algorithm can be used for both encoder-decoder (\cite{DBLP:conf/nips/VaswaniSPUJGKP17}) and decoder-only (\cite{DBLP:conf/iclr/LiuSPGSKS18}) architectures, as well as any other auto-regressive architecture. 

\subsection{Input-Output Representation}
\begin{figure}[htb]
    \centering
    \includegraphics[width=\linewidth]{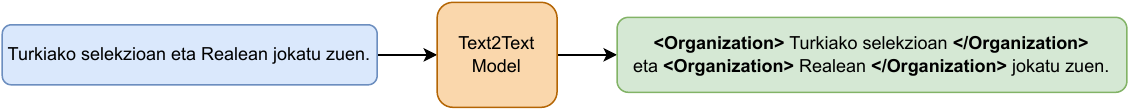}
    \caption{Text-to-Text representation of the Sequence Labeling task. Given an input sentence, the model must generate the same sentence annotated with html-style tags.}
    \label{fig6:constrained}
\end{figure}

The model is prompted with a sentence to label. The expected output is the same sentence annotated with HTML-style tags. An example is provided in Figure \ref{fig6:constrained}. The HTML tags for each task are added as special tokens to the model's vocabulary. Previous research (\cite{DBLP:conf/emnlp/0001NCHYS22}) found that different structures do not greatly impact the performance of the model so we use HTML-style tags because the format is easy for humans to read. Furthermore, LLMs, which have been trained on vast amounts of data from the Internet, are already familiar with this format, and implementing a constrained grammar for this structure is quite straightforward. In any case, our method can be adapted to any other task representation.
For encoder-decoder models, the unlabeled sentence is given as input into the encoder block, while the decoder block generates the labeled output. For encoder-only models, we use the token $\,\to\,$ during training as a separator between the unlabeled and labeled sentence. In the case of instruction-tuned models, instead of the separator, we use corresponding the chat-template to represent the unlabeled sentence as the user input and the labeled sentence as the chatbot response. We also experimented with generating only the labeled spans as output (i.e., \textit{<Person> Obama </Person> <Location> New York </Location>}), but we obtained worse results.

\subsection{Constrained decoding}
\begin{figure}[htb]
    \centering
    \includegraphics[width=\linewidth]{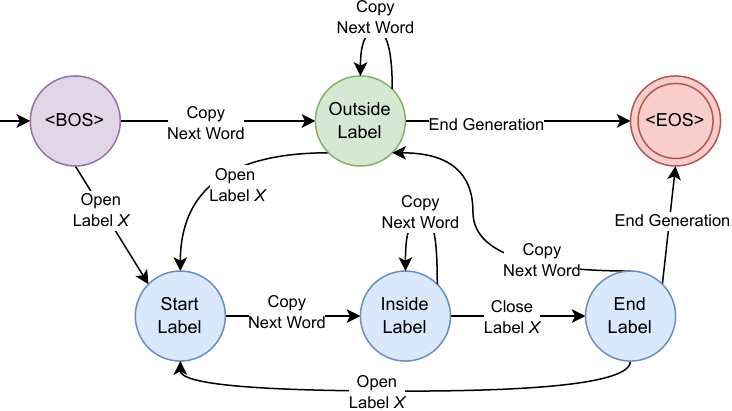}
    \caption{Our Constrained Decoding Algorithm is defined as a Finite State Automaton.}
    \label{fig6:automaton}
\end{figure}

The constrained decoding algorithm ensure that the output sequence contains the same words as the input sequence. This \textbf{prevents hallucinations}, which are very common when a model is trained in one language and then used to label sentences in another language. It also ensures that the output sequence is a valid HTML annotation, with no unclosed tags, empty tags, or other errors. This \textbf{prevents the generation of unparseable outputs}. We implement our constrained decoding algorithm using the Finite State Automaton described in Figure \ref{fig6:automaton}. At each stage, the model can generate only a set of valid tokens. This set includes copying the next word from the input (if the word is split by the tokenizer into multiple tokens, all of them are copied to prevent the splitting of words). It can also open an HTML tag, but only if no tag remains open, or close it, but only if we have already opened a tag and copied at least a word. The generation process ends when all the words in the input have been copied into the output and no tag remains open.

Given a sequence \((x_1, x_2, \dots, x_{t-1})\) that has been generated thus far and a set \(S_t\) of valid next tokens at step \(t\), the next token \(x_t\) is selected as:
\[ x_t = \arg \max_{x \in S_t} P(x|x_1, x_2, \ldots, x_{t-1}) \]
where \(P(x|x_1, x_2, \dots, x_{t-1})\) represents the conditional probability of token \(x\) given the prior tokens. Any token not in \(S_t\) is given a probability of zero, ensuring that the generated sequence adheres to the constraints. The probability for each token \(x_i \in S_t\) is computed using the softmax function applied to the model predictions:

\[
P(x_i|x_1, x_2, \ldots, x_{t-1}) = \frac{e^{x_i}}{\sum_j e^{x_j}}
\]

The probability of the generated sequence up to step \(T\) is computed as:

\[
P(x_{1:T}|\textrm{<bos>}) = \prod_{t=1}^{T} P(x_t|x_1, x_2, \ldots, x_{t-1})
\]

While most previous constrained decoding algorithms are limited to greedy decoding, we implement a \textbf{constrained beam search} approach. We keep track of the top \(k\) most probable sentences at each step \(t\), ensuring a broader exploration of the solution space and yielding higher-quality output sequences that adhere to the given constraints. Our constrained beam search approach adds very little overhead compared to the standard beam search decoding strategy. At each step, our only additional computation is to filter out invalid tokens from the beam. It's important to note that our constrained beam search decoding algorithm merely eliminates invalid sequences from the search space. Consequently, the constrained beam search will always yield an output that is at least as good as, if not superior to, unconstrained beam search.

\section{Experimental Setup}

\begin{figure}[htb]
    \centering
    \includegraphics[width=\linewidth]{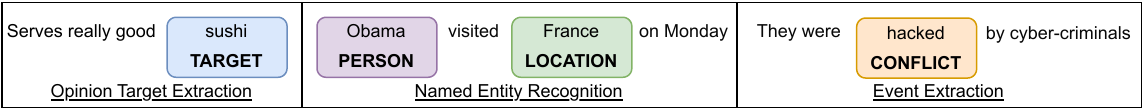}
    \caption{Information Extraction Tasks in our experiments}
    \label{fig6:tasks}
\end{figure}

The datasets used address three information extraction tasks which are illustrated by Figure \ref{fig6:tasks}.

\paragraph{Named Entity Recognition (NER):} This task consists of detecting
named entities and classifying them according to some pre-defined categories. We evaluate the models on MasakhaNER 2.0 (\cite{adelani-etal-2022-masakhaner}), a manually annotated NER dataset for 20 African languages. We train the models with the CoNLL03 (\cite{DBLP:conf/conll/SangM03}) English training split. We focus on named entities referring to Person, Location and Organization.

\paragraph{Opinion Target Extraction (OTE):} Given a review, the task is to detect the linguistic expression used to refer to the reviewed entity. We use the
English SemEval 2016 Aspect Based Sentiment Analysis (ABSA) datasets
(\cite{pontiki-etal-2016-semeval}). The English training split is used for fine-tuning; results are reported on the Spanish, French, Dutch, Russian and Turkish test sets.

\paragraph{Event Extraction (EE):} It consists of detecting and classifying event mentions according to some pre-defined class-inventory. We use the English ACE05 (\cite{ACE}) training split for training and the Chinese test split for evaluation. We also perform the Entity Mention Extraction task separately as an additional indicator of performance. 

\subsection{Language Models and baselines}

\paragraph{Baselines:} We assess the performance of our grammar-constrained beam search algorithm (\textbf{Cons}) against the unconstrained decoding baseline (\textbf{Base}). After fine-tuning, we test the same checkpoint using both constrained and unconstrained decoding. Additionally, our method is compared to popular encoder-only models, which currently set the benchmark for zero-shot cross-lingual transfer and have been widely adopted by the community. Thus, we evaluate mDeBERTa-v3 (\cite{DBLP:conf/iclr/HeLGC21/deberta}), an 86-million-parameter model, and GLOT500 (\cite{DBLP:conf/acl/ImaniLKSSKMSMYS23}), a 125-million-parameter model. Although we also experimented with XLM-RoBERTa (\cite{conneau-etal-2020-unsupervised}) models of various sizes, they consistently lagged behind mDeBERTa-v3 in performance. For MasakhaNER, we additionally compared with afro-xlmr-large (\cite{alabi-etal-2022-adapting}), a 355-million-parameter encoder-only model fine-tuned on African languages.

\paragraph{Text-to-text Models:} We experiment with three different encoder-decoder models: mT0-XL (\cite{DBLP:conf/acl/MuennighoffWSRB23}) 3.7 Billion parameter model. mT0-XL is an mT5 (\cite{mt5}) pretrained multilingual language model fine-tuned in the cross-lingual task mixture xP3. We also experimented with mT5 itself and Aya-101 (\cite{aya101}) an encoder-decoder model trained with instruction data in 101 languages. 

We also test multiple instruction tuned decoder-only models: Qwen2~(\cite{yang2024qwen2technicalreport}), gemma~(\cite{gemmateam2024gemmaopenmodelsbased}), LlaMA-3~(\cite{llama3modelcard}), Aya-23\\(\cite{aya23}) and Yi 1.5~(\cite{ai2024yiopenfoundationmodels}). These models have been trained on a wide range of tasks and languages, and have demonstrated strong multilingual capabilities.

\subsection{Training Setup}

All models were trained exclusively with English-labeled data and subsequently evaluated in the target languages. For the encoder-only models, we added a token classification layer (linear layer) on top of each token representation and trained them using the Cross-Entropy loss. The text-to-text models, were trained using the standard Next Token Prediction (NTP) loss. 
We finetune all the parameters of mT0 and mT5 using the Adafactor (\cite{DBLP:conf/icml/ShazeerS18}) optimizer. For the other text-to-text models, we found that the full-finetuning approach produces suboptimal results.
Therefore we use Low-Rank Adaptation (LoRA) (\cite{DBLP:journals/corr/abs-2106-09685}) to adapt the models to the target task. LoRA freezes the pre-trained model weights and injects trainable rank decomposition matrices into linear layers of the Transformer architecture.  We applied the LoRA to all linear Transformer block layers as recommended by \cite{qlora}. We use the AdamW  optimizer (\cite{DBLP:journals/corr/abs-1711-05101}). Preliminary experiments showed that LoRA produces a better performance than the full-finetuning approach for these models.  This has already been reported by previous research (\cite{sainz2024gollie}). We hypothesize that the large number of parameters in these models makes them prone to overfitting when finetuning all the hyperparameters on small datasets. For mT0 and mT5, we use a beam size of 4, while for larger models, we use a beam size of 1 as the computational cost of larger beams was prohibitive for us. In any case, we found that increasing the beam size did not significantly improve the performance of the models.

For both, encoder and text-to-text models we use the Huggingface open-source library (Apache-2.0 License) (\cite{DBLP:journals/corr/abs-1910-03771}).

\begin{table}[htb]
    \centering
    \small
    \adjustbox{max width=\textwidth}{
    \begin{tabular}{@{}lccc@{}}
    \toprule
     & Encoder Models & mT5/MT0 & Other text-to-text models \\ \midrule
    Finetuning & Full & Full & LoRA \\
    Batch Size & 32 & 16 & 32 \\
    Optimizer & AdamW & Adafactor & AdamW \\
    Learning Rate & $5e^{-5}$ & $1e^{-4}$ & $7e^{-5}$ \\
    Scheduler & Cosine & Cosine & Cosine \\
    Warnup steps & 0 & 500 & 500 \\
    Beams & - & 4 & 1 \\
    Sequence Length & 192 & 512 & 512 \\
    Preccision & FP16 & BF16 & BF16 \\
    Epochs (NER) & 20 & 20 & 10 \\
    Epochs (OTE) & 10 & 50 & - \\
    Epochs (ACE) & 20 & 45 & - \\ \bottomrule
    \end{tabular}}
    \caption{Hyperparameters used for fine-tuning the models.}
    \label{tab6:hparams}
    \end{table}

We evaluate the models at the end of several epochs on the validation set and select the best checkpoint based on the F1 score. The full training hyperparameters are provided in Table \ref{tab6:hparams}. These hyperparameters were chosen based on a hyperparameter search on the validation set. 

\subsection{Evaluation Metrics}

We evaluate the models using the standard F1-score metric for Sequence Labeling tasks (\cite{DBLP:conf/conll/SangM03}). For the text-to-text models, the output of the model is converted into an IOB2 format by splitting the output into words by whitespace. All the models are evaluated using the seqeval library (\cite{seqeval}).

\section{Experiments}
\label{sc6:experiments}

\subsection{Named Entity Recognition}
\label{sc6:ner}

In this section we will present and discuss the experiments in the Named Entity Recognition, Opinion Target Extraction and Event Extraction tasks. 

\begin{table}[htb]
    \centering
    \small
    \adjustbox{max width=\textwidth}{
    \begin{tabular}{@{}lcc|cc|cc|cc|ccc@{}}
    \toprule
     & \multicolumn{2}{c}{mT5-xl} & \multicolumn{2}{c}{mT0-xl} & \multicolumn{2}{c}{\cellcolor[HTML]{FFFFFF}{\color[HTML]{1F1F1F} aya-101}} & \multicolumn{2}{c}{\cellcolor[HTML]{FFFFFF}{\color[HTML]{1F1F1F} Yi-1.5-9B-Chat}} &  &  &  \\
    Lang & Base & Cons. & Base & Cons. & Base & Cons. & Base & Cons. & \multirow{-2}{*}{mDertaV3} & \multirow{-2}{*}{afro-xlmr-large} & \multirow{-2}{*}{GLOT500} \\ \midrule
    English & 93.4 & 93.7 & 93.2 & 93.3 & 93.2 & 93.4 & \cellcolor[HTML]{B7E1CD}94.5 & 94.3 & 93.4 & 93.4 & 92.3 \\ \midrule
    Bambara & 52.5 & 53.4 & 52.8 & 53.8 & 56.0 & \cellcolor[HTML]{B7E1CD}56.2 & 46.0 & 46.7 & 33.8 & 40.0 & 51.1 \\
    Ghomálá & 46.1 & 47.5 & 43.3 & 43.7 & 25.8 & 25.5 & 45.9 & \cellcolor[HTML]{B7E1CD}49.6 & 43.3 & 44.0 & 45.7 \\
    Éwé & 79.8 & 81.0 & 73.4 & 73.6 & 80.1 & \cellcolor[HTML]{B7E1CD}81.2 & 74.4 & 74.8 & 74.4 & 70.3 & 72.1 \\
    Fon & 52.0 & 55.4 & 68.0 & \cellcolor[HTML]{B7E1CD}69.7 & 44.5 & 45.3 & 47.4 & 52.2 & 49.2 & 49.8 & 56.7 \\
    Hausa & 71.3 & 73.8 & 70.0 & 71.9 & 67.9 & 70.1 & 62.0 & 61.3 & 70.7 & \cellcolor[HTML]{B7E1CD}74.1 & 67.2 \\
    Igbo & 72.6 & \cellcolor[HTML]{B7E1CD}77.2 & 55.9 & 61.0 & 53.5 & 54.1 & 55.5 & 57.3 & 58.8 & 72.5 & 62.1 \\
    Kinyarwanda & 71.9 & 73.1 & 71.9 & \cellcolor[HTML]{B7E1CD}74.3 & 67.3 & 69.2 & 48.7 & 53.3 & 65.7 & 67.9 & 66.1 \\
    Luganda & 81.9 & 82.3 & 79.0 & 79.5 & 82.8 & \cellcolor[HTML]{B7E1CD}83.1 & 66.9 & 73.6 & 73.0 & 77.9 & 79.2 \\
    Mossi & 52.5 & 53.7 & 55.4 & 55.7 & 56.7 & 56.7 & 54.4 & \cellcolor[HTML]{B7E1CD}57.1 & 44.6 & 45.7 & 51.4 \\
    Naija & 76.3 & \cellcolor[HTML]{B7E1CD}83.5 & 73.5 & 80.1 & 69.3 & 72.1 & 60.5 & 63.5 & 78.7 & 80.4 & 71.1 \\
    Chichewa & 77.7 & 78.8 & 76.5 & 76.7 & 79.9 & 80.2 & \cellcolor[HTML]{B7E1CD}80.5 & 80.4 & 73.7 & 79.6 & 76.6 \\
    chiShona & 35.2 & 48.2 & 24.3 & \cellcolor[HTML]{B7E1CD}54.0 & 35.2 & 42.8 & 25.8 & 43.0 & 35.8 & 35.2 & 39.8 \\
    Kiswahili & 86.4 & \cellcolor[HTML]{B7E1CD}89.6 & 85.7 & 88.0 & 83.3 & 84.7 & 73.1 & 71.4 & 86.7 & 88.2 & 84.0 \\
    Setswana & 81.0 & 81.3 & 72.3 & 73.5 & 82.1 & \cellcolor[HTML]{B7E1CD}82.6 & 60.4 & 64.0 & 63.1 & 73.3 & 66.8 \\
    Akan/Twi & 60.2 & 61.4 & 60.1 & 61.5 & 64.0 & \cellcolor[HTML]{B7E1CD}64.6 & 49.7 & 55.5 & 49.9 & 40.3 & 55.9 \\
    Wolof & 53.3 & 54.3 & 56.4 & 56.8 & 61.3 & \cellcolor[HTML]{B7E1CD}62.3 & 57.1 & 60.9 & 42.0 & 51.3 & 61.6 \\
    isiXhosa & 30.5 & 40.3 & 27.0 & \cellcolor[HTML]{B7E1CD}55.8 & 34.0 & 40.8 & 22.7 & 32.1 & 24.9 & 26.0 & 26.5 \\
    Yorùbá & 55.1 & \cellcolor[HTML]{B7E1CD}58.5 & 51.0 & 51.3 & 26.0 & 25.8 & 48.9 & 55.4 & 34.1 & 52.5 & 54.4 \\
    isiZulu & 49.4 & 54.9 & 39.2 & \cellcolor[HTML]{B7E1CD}66.7 & 40.5 & 45.0 & 23.5 & 33.7 & 44.7 & 47.1 & 43.3 \\ \midrule
    \textbf{Average MasakhaNER} & 62.4 & \cellcolor[HTML]{B7E1CD}65.7 & 59.8 & 65.7 & 58.4 & 60.1 & 52.8 & 57.1 & 55.1 & 58.7 & 59.6 \\ \bottomrule
    \end{tabular}
    }
    \caption{F1 scores in the Named Entity Recognition Task. Model are trained in English and evaluated in a set of African languages.}
    \label{tab6:NER}
    \end{table}

Table \ref{tab6:NER} presents the performance of our method compared to the baselines in the NER task. All models show comparable performance in English. However, when assessing zero-shot cross-lingual transfer, significant performance differences emerge.

In the zero-shot cross-lingual transfer setting, constrained decoding consistently outperforms unconstrained decoding. For some languages, such as Bambara, Ghomálá, and Éwé, both methods yield similar results. In contrast, other languages, including Shona, isiXhosa, and Zulu, exhibit marked performance improvements. These Southern Bantu languages have unique linguistic features: they capitalize proper names following the noun class prefix (e.g., kweZambia) and display highly inflected morphology (\cite{adelani-etal-2022-masakhaner}). These attributes challenge the cross-lingual transfer abilities of English fine-tuned NER models. Consequently, all baseline models, including the encoder-only variants, perform suboptimally in these languages and are clearly outperformed by our constrained decoding approach.

As demonstrated in Section \ref{sc6:ablation}, text-to-text models struggle with agglutinative languages, frequently mislabeling entities by arbitrarily splitting them into sub-words. Our constrained decoding corrects this by ensuring that the output sentence retains the original words from the input sentence. Overall, constrained decoding excels in the zero-shot cross-lingual setting for languages with highly inflected agglutinative morphology. Although the performance gap is less pronounced for language isolates like Bambara, Éwé, Fon, and Twi, it remains significant.

\begin{table}[htb]
    \centering
    \begin{tabular}{@{}lccc@{}}
    \toprule
    \multicolumn{1}{l}{Model} & Unconstrained & Constrained & \multicolumn{1}{c@{}}{Delta} \\ \midrule
    mT5-xl & 62.4 & 65.7 & \multicolumn{1}{c@{}}{+3.3} \\
    mT0-xl & 59.8 & 65.7 & \multicolumn{1}{c@{}}{+5.9} \\
    aya-101 & 58.4 & 60.1 & \multicolumn{1}{c@{}}{+1.7} \\
    Qwen2-7B-Instruct & 39.7 & 42.0 & \multicolumn{1}{c@{}}{+2.3} \\
    gemma-1.1-7b-it & 46.8 & 49.0 & \multicolumn{1}{c@{}}{+2.2} \\
    Llama-3-8B-Instruct & 51.2 & 52.7 & \multicolumn{1}{c@{}}{+1.6} \\
    aya-23-8B & 51.6 & 52.6 & \multicolumn{1}{c@{}}{+0.9} \\
    Yi-1.5-9B-Chat & 52.8 & 57.1 & \multicolumn{1}{c@{}}{+4.3} \\ \midrule
    GLOT500 & \multicolumn{2}{c}{59.6} & \\
    mDeBERTa-v3 & \multicolumn{2}{c}{55.1} & \\
    Davlan/afro-xlmr-large & \multicolumn{2}{c}{58.7} & \\ \bottomrule
    \end{tabular}
    \caption{Average F1 scores in the MasakhaNER dataset.}
    \label{tab6:ner_mini}
    \end{table}

Models exhibit varying performance across languages. For instance, aya-101 achieves the best performance for Éwé, Luganda, Setswana, Twi, and Wolof, while mT0 is superior for languages such as Fon, Kinyarwanda, chiShona, isiXhosa, and Zulu. We attribute this to the different training data used by the models. Nonetheless, we observe that mT5-xl and mT0-xl, combined with our constrained decoding algorithm, outperform encoder-only models by more than 5 points in F1 score on average. This represents a significant improvement over the previous state-of-the-art for zero-shot cross-lingual transfer in NER tasks.

In Table \ref{tab6:ner_mini} we present the average performance of different models on the MasakhaNER dataset. Qwen2, Gemma, and Aya-23 achieve suboptimal results compared to the other text-to-text models. This is likely due to these models being trained on a smaller number of high-resource languages, rendering them less proficient in African languages. However, the results demonstrate that constrained decoding is effective in improving the performance of all text-to-text models in zero-shot cross-lingual transfer.

\subsection{Opinion Target Extraction}
\label{sc6:ote}

\begin{table}[htb]
    \small
    \centering
    \adjustbox{max width=\linewidth}{%
    \begin{tabular}{@{}lcc|cc@{}}
    \toprule
     & \multicolumn{2}{c}{mT0-xl} &  &  \\
    Lang & Base & Cons & \multirow{-2}{*}{\begin{tabular}[c]{@{}c@{}}GLOT\\ 500\end{tabular}} & \multirow{-2}{*}{\begin{tabular}[c]{@{}c@{}}mDeBERTa\\ V3\end{tabular}} \\ \midrule
    English & 82.6 & \cellcolor[HTML]{B7E1CD}{84.8} & 82.6 & 83.6 \\ \midrule
    Spanish & 77.8 & \cellcolor[HTML]{B7E1CD}{79.4} & 69.4 & 78.0 \\
    French & 74.1 & 76.6 & 65.8 & \cellcolor[HTML]{B7E1CD}{76.9} \\
    Dutch & 74.1 & 77.1 & 66.5 & \cellcolor[HTML]{B7E1CD}{77.3} \\
    Russian & 71.1 & 75.7 & 69.2 & \cellcolor[HTML]{B7E1CD}{76.5} \\
    Turkish & 56.8 & \cellcolor[HTML]{B7E1CD}{57.7} & 50.4 & 56.4 \\ \midrule
    Average & 70.8 & \cellcolor[HTML]{B7E1CD}{73.3} & 64.3 & 73.0 \\ \bottomrule
    \end{tabular}
    }
    \caption{F1 scores in the Opinion Target Extraction Task.}
    \label{tab6:OTE}
    \end{table}

In the NER task, we experimented with cross-lingual transfer approaches using a set of low-resource African languages that significantly differ from English. For the Opinion Target Extraction task, we evaluated cross-lingual transfer performance into languages from the Indo-European language family. Due to the high computational cost of the text-to-text models, we only evaluated the best model from the previous task, mT0-XL.

As shown in Table \ref{tab6:OTE}, excluding Turkish (an agglutinative language), the performance decline in the target languages compared to English is less pronounced, suggesting a more seamless transfer. Even in this context, our constrained generation algorithm significantly surpasses the unconstrained generation. Finally, while mT0-XL and mDeBERTa-v3 show comparable performance, our approach demonstrates slightly higher average performance across the board.

\subsection{Event Extraction}
\label{sc6:ee}

\begin{table}[htb]
    \centering
    \small
    \adjustbox{max width=\linewidth}{%
    \begin{tabular}{@{}lcc|cc@{}}
    \toprule
     & \multicolumn{2}{c}{mT0-xl} &  &  \\
    Lang & Base & Cons & \multirow{-2}{*}{\begin{tabular}[c]{@{}c@{}}GLOT\\ 500\end{tabular}} & \multirow{-2}{*}{\begin{tabular}[c]{@{}c@{}}mDeBERTa\\ V3\end{tabular}} \\ \midrule
    English\textsubscript{Entity} & \cellcolor[HTML]{B7E1CD}{95.5} & \cellcolor[HTML]{B7E1CD}{95.5} & 94.5 & 95.3 \\
    Chinese\textsubscript{Entity} & 70.1 & \cellcolor[HTML]{B7E1CD}{73.3} & 34.1 & 54.2 \\ \midrule
    English\textsubscript{Trigger} & \cellcolor[HTML]{B7E1CD}{78.9} & \cellcolor[HTML]{B7E1CD}{78.9} & 74.1 & 78.0 \\
    Chinese\textsubscript{Trigger} & 49.6 & \cellcolor[HTML]{B7E1CD}{52.1} & 0.0 & 30.5 \\ \bottomrule
    \end{tabular}
    }
    \caption{F1 scores in the Event Extraction Task.}
    \label{tab6:EE}
    \end{table}

For the Event Extraction task we aim to perform zero-shot cross-lingual transfer from English to Chinese. This task is particularly challenging due to the vast linguistic and cultural differences between the two languages, including script type, syntax, semantics, and the use of tones in Chinese. As reported in Table \ref{tab6:EE}, both GLOT500 and mDEBERTa struggle with the transfer from English to Chinese, whereas mT0-XL achieves much better results. Consistent with previous evaluations, our constrained generation approach improves over the unconstrained generation method by approximately 3 points in F1 score.

\subsection{Model Transfer vs Data Transfer}
\label{sc6:transfer}

In this chapter we focus on improving the zero-shot model-transfer approach. However, constrained decoding can also be used in conjunction with data transfer. In this section we compare the performance of the constrained decoding algorithm when used in both zero-shot model-based transfer and data transfer settings using the MasakhaNER NER dataset. To this end, we use the automatically generated NER datasets for eight African languages from Chapter \ref{ch:data-transfer}. These datasets were generated by translating the CoNLL2003 (\cite{DBLP:conf/conll/SangM03}) English dataset into the target languages using NLLB200 (\cite{DBLP:journals/corr/abs-2207-04672}) and then projecting the labels using T-Projection. Using the same settings as for the zero-shot approach, we evaluate the performance of the constrained decoding algorithm when fine-tuning the models on the generated datasets. The results are presented in Table \ref{tab6:modelvsdata}. ``Zero'' refers to the models trained with English CoNLL 2003 data and evaluated in the target languages, while "Data" refers to the models fine-tuned on the translated CoNLL 2003 datasets. Both settings use the constrained decoding algorithm.

\begin{table}[htb]
    \centering
    \small
    \adjustbox{max width=\textwidth}{
        \begin{tabular}{@{}l|cc|cc|cc|cc|cc|cc@{}}
            \toprule
            \multicolumn{1}{l}{} & \multicolumn{2}{c}{mDebertaV3} & \multicolumn{2}{c}{mT5-xl}                   & \multicolumn{2}{c}{mT0-xl}               & \multicolumn{2}{c}{aya-101}              & \multicolumn{2}{c}{Llama-3-8B-Instruct} & \multicolumn{2}{c}{Yi-1.5-9B-Chat} \\
            Lang                 & Zero    & Data   & Zero                    & Data & Zero & Data                & Zero & Data                & Zero           & Data          & Zero      & Data     \\ \midrule
            Hausa                & 70.7         & 72.7            & \cellcolor[HTML]{B7E1CD}73.8 & 67.8          & 71.9      & 72.4                         & 70.1      & 72.6                         & 63.5                & 70.1                   & 61.3           & 62.4              \\
            Igbo                 & 58.8         & 71.4            & 77.2                         & 69.9          & 61.0      & 72.9                         & 54.1      & \cellcolor[HTML]{B7E1CD}82.7 & 54.7                & 73.7                   & 57.3           & 58.7              \\
            Chichewa             & 73.7         & 77.2            & 78.8                         & 51.6          & 76.7      & 76.8                         & 80.2      & \cellcolor[HTML]{B7E1CD}83.0 & 78.6                & 68.2                   & 80.4           & 52.4              \\
            chiShona             & 35.8         & 74.9            & 48.2                         & 75.0          & 54.0      & 74.1                         & 42.8      & \cellcolor[HTML]{B7E1CD}77.0 & 25.3                & 65.3                   & 43.0           & 46.2              \\
            Kiswahili            & 86.7         & 85.5            & \cellcolor[HTML]{B7E1CD}89.6 & 77.4          & 88.0      & 85.1                         & 84.7      & 85.4                         & 77.2                & 80.1                   & 71.4           & 65.0              \\
            isiXhosa             & 24.9         & 72.3            & 40.3                         & 53.6          & 55.8      & \cellcolor[HTML]{B7E1CD}74.8 & 40.8      & 74.5                         & 25.8                & 64.8                   & 32.1           & 46.0              \\
            Yorùbá               & 34.1         & 42.7            & 58.5                         & 37.1          & 51.3      & 46.7                         & 25.8      & \cellcolor[HTML]{B7E1CD}60.5 & 34.9                & 56.4                   & 55.4           & 37.6              \\
            isiZulu              & 44.7         & 66.7            & 54.9                         & 64.4          & 66.7      & \cellcolor[HTML]{B7E1CD}71.2 & 45.0      & 64.9                         & 22.5                & 58.0                   & 33.7           & 36.4              \\ \midrule
            Average              & 53.7         & 70.4            & 65.2                         & 62.1          & 65.7      & 71.8                         & 55.5      & \cellcolor[HTML]{B7E1CD}75.1 & 47.8                & 67.1                   & 54.3           & 50.6              \\ \bottomrule
            \end{tabular}
    }
    \caption{F1 Scores for Named Entity Recognition Task. ``Zero'' refers to the model trained in English and evaluated on a set of African languages. ``Data'' refers to the model trained on automatically translated and projected data using T-Projection for each language.}
    \label{tab6:modelvsdata}
    \end{table}

The results show that the zero-shot cross-lingual transfer performance when using text-to-text models such as mT5-xl or mT0-xl is significantly better than the zero-shot performance of mDeBERTa-v3, as we already demonstrated in Section \ref{sc6:ner}. However, mDeBERTa-v3 shows very competitive results in the data-transfer setting. For text-to-text models, the performance of the zero-shot and data-transfer approaches varies across languages. In languages where the model is proficient, such as Hausa or Igbo for mT5-xl, the zero-shot approach outperforms the data-transfer approach. However, in languages where the model is less proficient, such as isiXhosa or Zulu, the data-transfer approach is superior. In the case of aya-101 and LLama-3, which we reported to be less proficient in African languages in Section \ref{sc6:ner}, the data-transfer approach results in a significant performance improvement. In fact, aya-101 outperforms all other models in the data-transfer setting. This suggests that the constrained decoding algorithm can be used in conjunction with data-transfer methods to further improve the performance of models in low-resource languages. 

Similar to the insights from Chapter \ref{ch:model-vs-data}, the results suggest that when a model is proficient in both the source and target languages, model-based transfer is superior to data-based transfer. Thanks to the methodology developed in this chapter, we can now successfully leverage the power of text-to-text LLMs in a zero-shot setting to achieve superior zero-shot cross-lingual transfer results. However, when the model is less proficient in the target language, data-based transfer can be a better option. Data transfer also has the advantage of allowing the use of more efficient models. The results demonstrate that while mDeBERTa-v3 is not competitive in the zero-shot setting, it achieves similar results to the best-performing text-to-text models in the data-transfer setting, despite having fewer parameters and requiring less computational resources.

\section{Ablation Study}
\label{sc6:ablation}

In this section we aim to better understand why and in which scenarios constrained decoding performs better than unconstrained decoding. To achieve this, we identify the types of mistakes that unconstrained decoding makes which are subsequently fixed by constrained decoding. These errors can be grouped into three categories: inconsistent HTML markups, word hallucinations, and word splittings.

\begin{figure}[tb]
    \centering
    \includegraphics[width=0.8\linewidth]{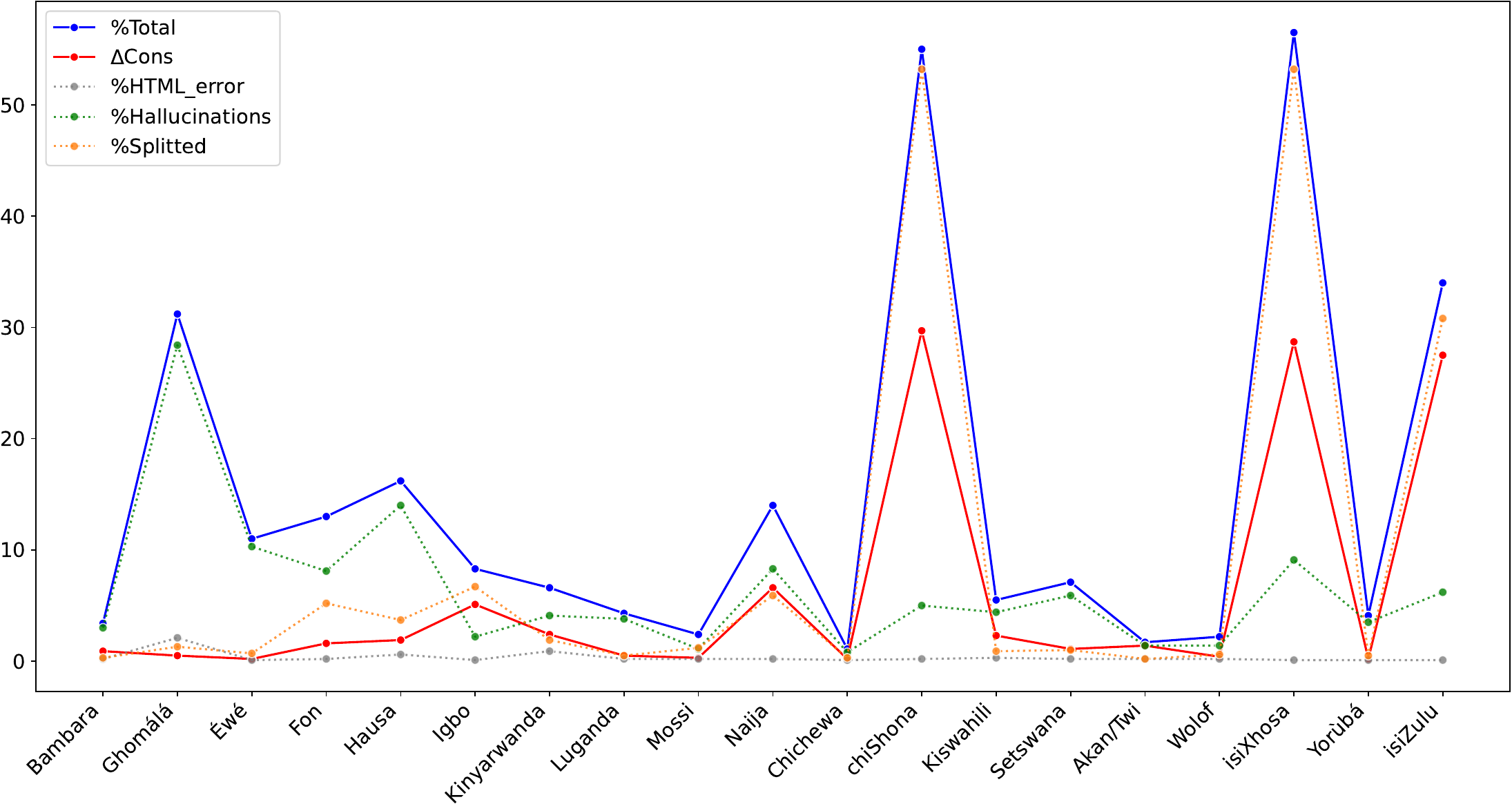}
    \caption{Percentage of hallucinated words compared to the performance delta between unconstrained and unconstrained beam search in MasakhaNER using mT0-XL.}
    \label{fig6:Hallucination}
\end{figure}

\paragraph{Inconsistent HTML markups:} The model occasionally generates HTML markup that cannot be parsed correctly, such as when a tag is opened but never closed. We found that this occurs in less than 1\% of the annotated sentences. Consequently, it has a negligible effect on the overall performance of the model.

\paragraph{Word hallucinations:} The model sometimes includes in the output a word that was not present in the input. This occurs because unconstrained generation often produces output that mixes English and the target language. For instance, given the sentence \textit{``Kaliforni sullã sẽn togse''}, mT0-XL, when using unconstrained decoding, produces \textit{``<Location> California </Location> sullã sẽn togse''}. In this instance, the model has translated \textit{``Kaliforni''} to \textit{``California''}. Furthermore, inadvertent translation is not the only cause of hallucinations in the output. Perhaps due to a limited understanding of the target language, the model often introduces typos (e.g., \textit{``okudlula''} incorrectly becomes \textit{``okudludlule''}). Interestingly, it even mixes African languages. For instance, given a Zulu sentence as input containing the word \textit{``Musawenkosi''} (God Bless You), the model outputs the very similar Chichewa word \textit{``Mumawenkosi''} (You are welcome).

\paragraph{Word Splittings:} They refer to instances where the model either divides a word into multiple subwords or, conversely, combines several words into a single one. This occurs because the model has been trained in English and, when tested on agglutinative languages, it attempts to mimic English morphology by arbitrarily splitting words. For instance, the sequence \textit{``<Location> waseThekwini </Location> <Person> uShauwn Mkhize </Person>''} becomes \textit{``wase <Location> Thekwini </Location> u <Person> Shauwn Mkhize </Person>''}. This behavior is interesting, as lemmatization is a component of many downstream Information Extraction applications. One could argue that this is the desired behavior. However, although accidental lemmatization was performed correctly in this particular example, this is not usually the case. For instance, in Basque (whose results are not reported here for brevity, although the models were tested in this language), as illustrated in Figure \ref{fig6:constrained_unconstrained}, the model incorrectly splits the term \textit{``Realean''} into \textit{``Reale''} and \textit{``an''}. However, \textit{``Reale''} does not represent the correct lemma, which would correspond to \textit{``Reala''}, the name of a football team. Therefore, the model seems to be arbitrarily splitting words to mimic English morphology.

\begin{figure}[tbp]
    \centering
    \includegraphics[width=\linewidth]{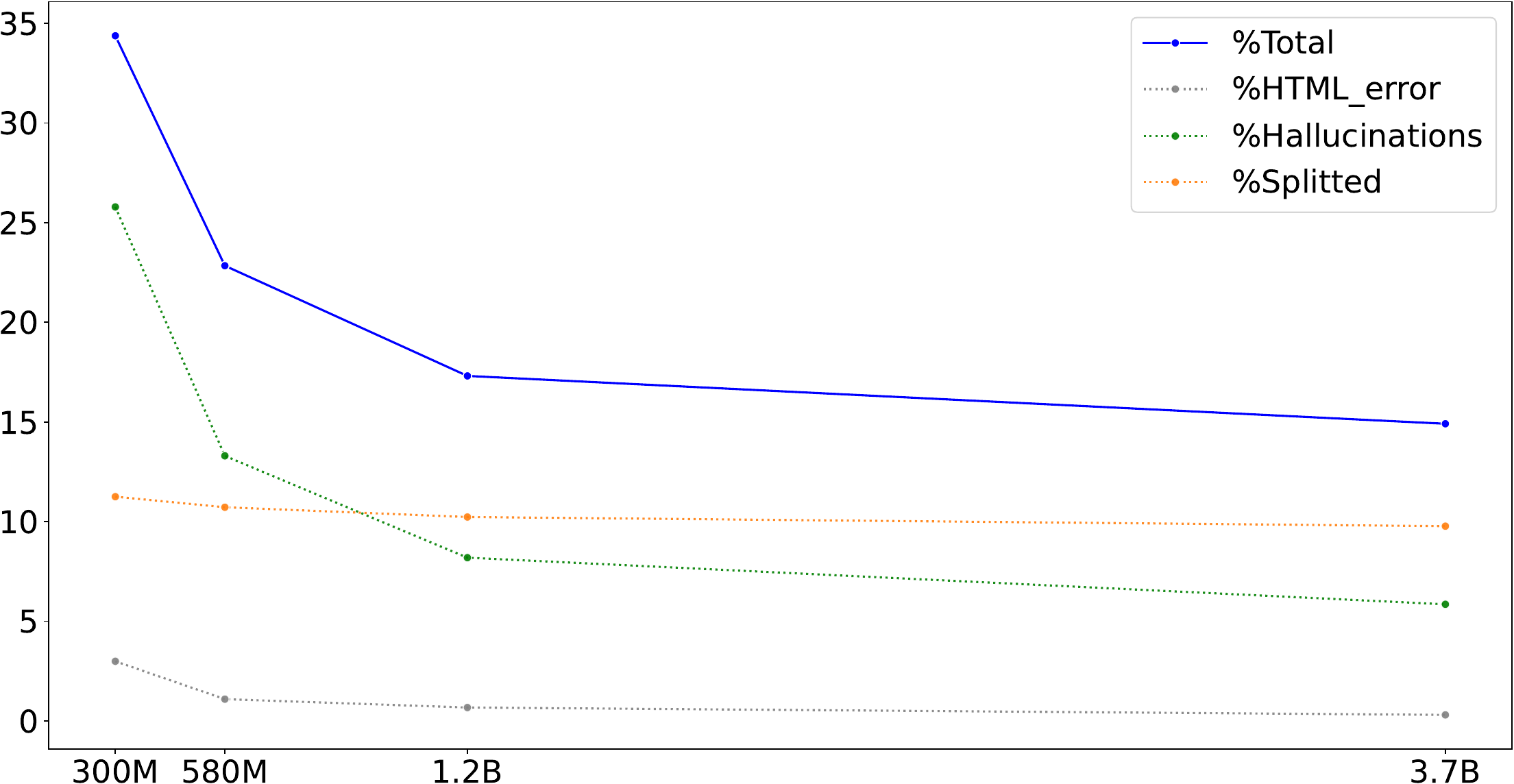}
    \caption{Average percentage of mistakes generared by Unconstrained Beam search in MasakhaNER using mT0 models of different sizes}
    \label{fig6:HallucinationvsParams}
\end{figure}

We calculate the percentage of sentences containing some of these errors for each language in the NER task when using mT0-XL with unconstrained generation. The results are depicted in Figure \ref{fig6:HallucinationvsParams}. Additionally, we compared the overall percentage of sentences containing any error with the performance difference between constrained and unconstrained generation. The larger the delta, the greater the performance improvement with constrained generation.

Figure \ref{fig6:HallucinationvsParams} indicates that word splitting and hallucinations correlate with the performance delta, suggesting that addressing these issues is key to the superiority of the constrained generation algorithm. It also underscores that unconstrained generation produces a substantial proportion of sentences with errors. In cases like chiShona and isiXhosa (discussed in Section \ref{sc6:ner}), this could affect over 50\% of the output sentences. It should be noted that word splitting has a more pronounced effect on the performance delta than hallucinations. This can be attributed to the standard sequence evaluation method used for these tasks. 

To convert the model's output into IOB2 encoding, we derive annotations such as \textit{"B-LOC O O O"} for the example \textit{``<Location> California </Location> sullã sẽn togse''}. This annotation remains accurate even if the model translates the entity into English. However, when the model splits or merges words, the IOB2 labeling is disrupted, rendering the sentence incorrect in the evaluation. Thus, although the evaluation method may gloss over hallucination errors, it is important to note that models generate a significant number of hallucinations when producing unconstrained predictions, potentially impacting the ultimate efficacy and applicability of IE systems.

We also evaluated the total number of mistakes generated by unconstrained beam search in the NER task with mT0 models of varying sizes. As illustrated in Figure \ref{fig6:HallucinationvsParams}, word splitting and inconsistent HTML markups remain consistent across models with different parameter sizes. However, the frequency of hallucinations decreases as the model size increases. This might be because models with more parameters have a more refined representation of individual languages and therefore mix languages less frequently.

\begin{figure}[tbp]
    \centering
    \includegraphics[width=\linewidth]{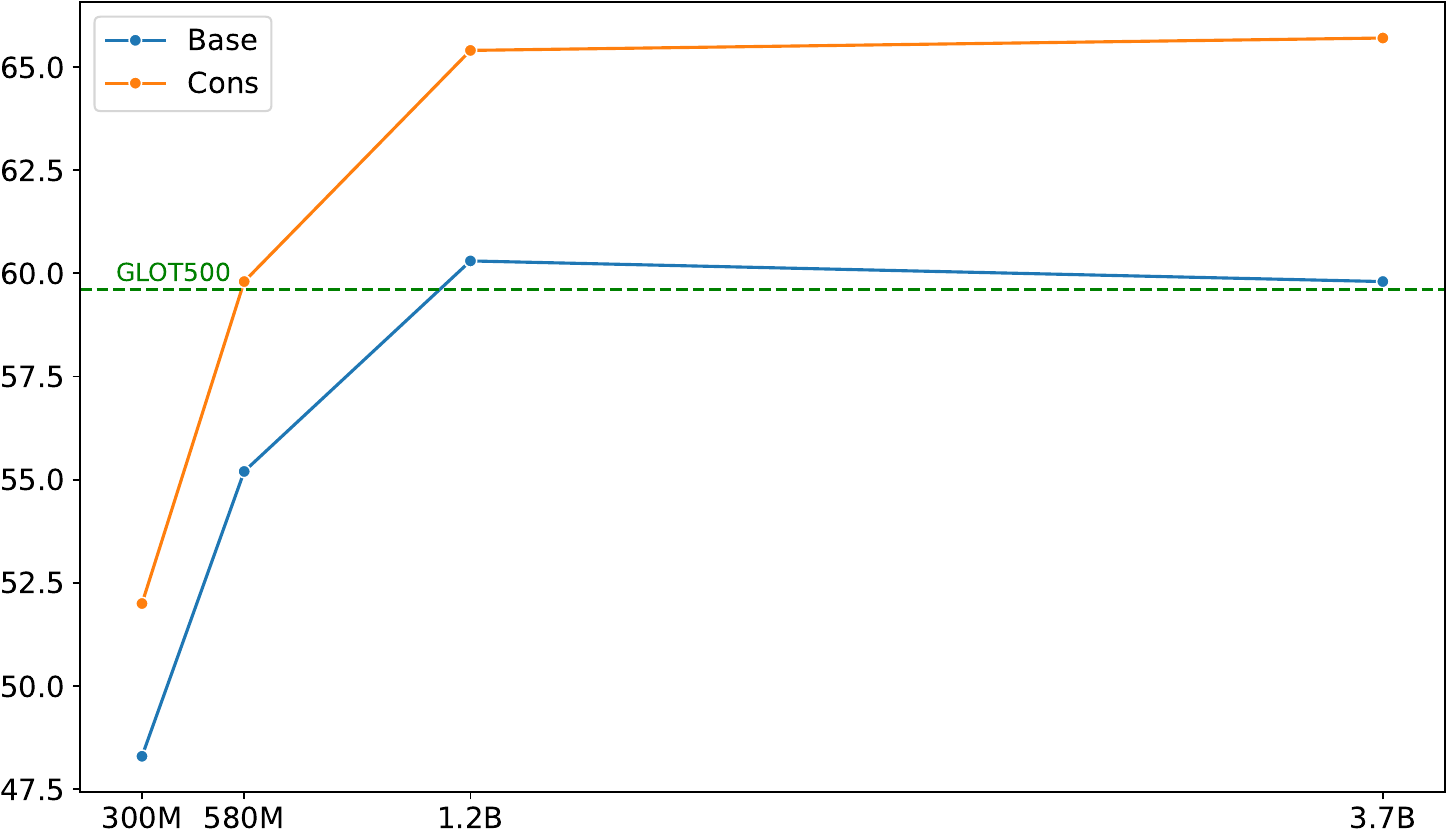}
    \caption{Average F1 score in MasakhaNER compared to the mT0 model size}
    \label{fig6:F1vsParams}
\end{figure}

Additionally, we assess the average F1 score in the NER task for mT0 models ranging from 300 million to 3.7 billion parameters. The results, presented in Figure \ref{fig6:F1vsParams}, show that as the mT0 model's parameter count increases, the F1 score improves, although we observe diminishing returns beyond 1.2 billion parameters. While our experiments utilize the 3.7 billion parameter mT0-XL, constrained generation surpasses both GLOT500 (a 125 million parameter model) and afro-xlmr-large (355 million parameters) when using an mT0 model with only 580 million parameters. This indicates that the superiority of our method over encoder-only models is not solely due to leveraging a larger model. Notably, with constrained generation, the 580 million parameter mT0 model achieves performance comparable to the 1.2 billion parameter model when the latter employs unconstrained generation. Therefore, constrained generation is also considerably more computationally efficient than its unconstrained counterpart.

\begin{figure}[t]
    \centering
    \includegraphics[width=0.8\linewidth]{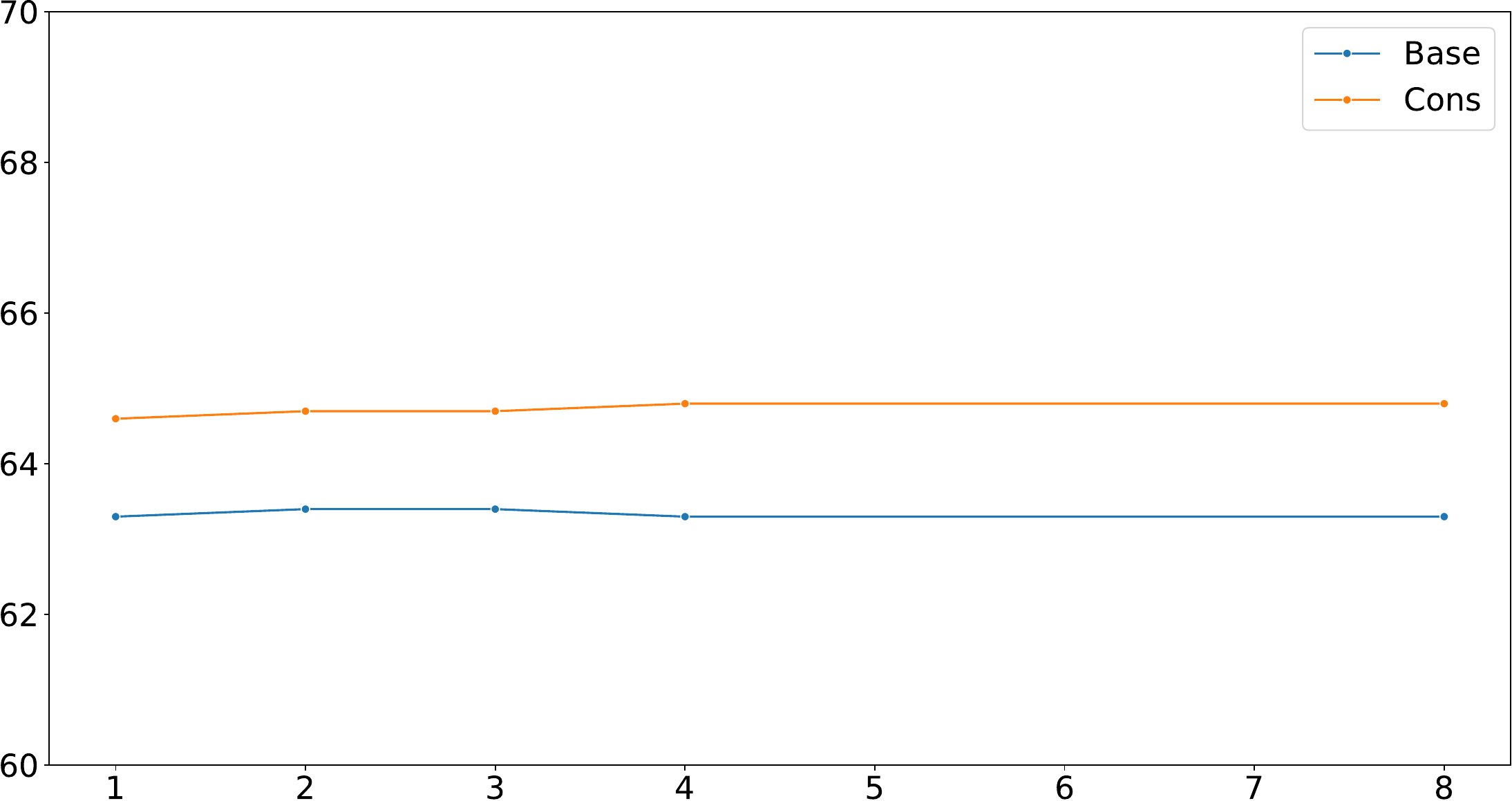}
    \caption{Average F1 score of mT0-XL in a subset of MasakhaNER compared to the number of beams used for decoding.}
    \label{fig6:BeamsF1}
\end{figure}

Finally, we evaluate the performance of mT0-XL using a varying number of beams. We assess the same checkpoint with beam search ranging from 1 to 8 beams. For these experiments, we utilize a subset of MasakhaNER2, which includes the following languages: Bambara, Ghomálá, Éwé, Fon, Hausa, Igbo, Kinyarwanda, Luganda, and Mossi. As illustrated in Figure \ref{fig6:BeamsF1}, increasing the number of beams has a negligible effect on performance. Considering that the computational cost and GPU memory requirements increase linearly with the number of beams, in this scenario, using a single beam (greedy decoding) offers the best performance-to-cost ratio. This is because the model is highly confident in its top prediction during each step of the decoding process, and introducing additional beams does not significantly diversify or improve the generated outputs.

\section{Conclusion}

In this Chapter we introduce a Constrained Beam Search Algorithm that can be seamlessly incorporated into any text-to-text LLM. We demonstrate that, compared to Unconstrained Beam Search, our algorithm significantly improves zero-shot cross-lingual performance across a broad range of IE tasks and languages. Through an extensive ablation study, we show that constrained generation effectively mitigates issues such as word-splitting and language mixing, which lead to typos and unintentional translations, errors commonly observed when applying text-to-text models to these tasks. Our approach allows the text-to-text mT0 language model to outperform encoder-only models, which had previously set the state-of-the-art standard for zero-shot cross-lingual IE. To the best of our knowledge, we present the best zero-shot cross-lingual results up to date. 

The method developed in this chapter enables model-based cross-lingual transfer for sequence labelling tasks with text-to-text models. This is a significant step forward in the field of zero-shot cross-lingual transfer, as it allows for the use of more powerful models that can handle a wide range of tasks. Considering the prevailing focus on text-to-text LLMs in current research, and the infrequent training of new encoder-only models, we believe that this represents significant progress in this research area. 
\selectlanguage{english}
\chapter[Medical MT5: Cross-Lingual Transfer for Domain-Spacific Task]{Medical MT5: Cross-Lingual Transfer for Domain-Spacific Task}
\label{ch:medicalmt5}

In this chapter we will introduce Medical mT5, an open-source multilingual text-to-text large language model for the medical domain. We will leverage all the data-transfer and model-transfer techniques developed in the previous chapters. We will build a multilingual pre-training, fine-tuning, and evaluation framework for the medical domain. Medical mT5 demonstrates the importance of the technology and knowledge developed in this thesis, resulting in the first multilingual text-to-text medical model when it was created.

\section{Motivation and Contributions}

As it is the case for many application domains, there is an increasing interest in applying Artificial Intelligence (AI) and Natural Language Processing (NLP) techniques to assist medical experts in their everyday activities. With this aim in mind, a number of language models have been trained or adapted to the medical domain. These include encoder-only models such as  SciBERT (\cite{beltagy2019scibert}), BioBERT (\cite{DBLP:journals/bioinformatics/LeeYKKKSK20}) or PubmedBERT (\cite{DBLP:journals/health/GuTCLULNGP22}). These models have obtained state-of-the-art results in discriminative tasks, with the advent of text-to-text and text-generation models, a new generation of language models has been developed. These models are typically much larger and have a much broader scope than the encoder-only models. Examples of these models include SciFive (\cite{DBLP:journals/corr/abs-2106-03598}), BioGPT (\cite{10.1093/bib/bbac409})
Med-PaLM (\cite{singhal-palm}), PMC-LLaMA (\cite{wu2023pmcllama}) or ClinicalGPT (\cite{Wang2023ClinicalGPTLL}).

However, the development of all the aforementioned text-to-text LLMs has been focused on a single language, usually English. As a consequence, there is a lack of high-quality multilingual data for pre-training models, a lack of models themselves, and a lack of high-quality multilingual evaluation benchmarks for the medical domain. Although there have been efforts to generate evaluation data in languages other than English (\cite{Wang2023ClinicalGPTLL,carrino-etal-2022-pretrained}), they have consisted largely of monolingual approaches.

To address these issues, we have compiled, to the best of our knowledge, the largest multilingual corpus for training LLMs adapted to the medical domain. Our corpus includes 3B words in four languages, namely, English, Spanish, French, and Italian. While relatively small when compared to existing English datasets (\cite{wu2023pmcllama}), it allowed us to build Medical mT5, the first open-source text-to-text multilingual model for the medical domain. Additionally, we have built a multilingual evaluation framework for the medical domain that can be used to evaluate the performance of any future multilingual model in the medical domain. 

Medical mT5 has been built on the work presented in previous chapters. We use the data transfer approach to reduce the cost of annotating new multilingual evaluation benchmarks for the medical domain. We also use constrained decoding to achieve high-quality zero-shot model-based cross-lingual transfer. Medical mT5 demonstrates the benefits of the techniques developed during this thesis and their application in real-world scenarios where data is scarce.

Medical mT5 outperforms similarly-sized text-to-text models for the Spanish, French, and Italian benchmarks while being competitive in English to current state-of-the-art text-to-text (\cite{mt5,chung-flan-instruction-models}) and encoder-only models (\cite{DBLP:journals/bioinformatics/LeeYKKKSK20,DBLP:conf/iclr/HeLGC21/deberta}). The results show that continuing pre-training of a multilingual text-to-text model such as mT5 allows to successfully adapt it to the medical domain, even when the amount of domain-specific data is relatively modest (ranging between 1B words for English and Spanish to 150M in Italian). Summarizing, the contributions of this chapter are:

\begin{itemize}
    \item The collection of the largest publicly available in-domain medical multilingual corpus for Spanish, French, and Italian languages. Together with the already existing English data, we release a corpus of 3 billion tokens\footnote{\url{https://hf.co/datasets/HiTZ/Multilingual-Medical-Corpus}}.
    \item We use the data-transfer approaches developed in previous chapters to build two new datasets for Spanish, French, and Italian on Argument Mining\footnote{\url{https://hf.co/datasets/HiTZ/multilingual-abstrct}} and generative Question-Answering\footnote{\url{https://hf.co/datasets/HiTZ/Multilingual-BioASQ-6B}} tasks, generated by taking their original English versions as a starting point.
    \item the public release of two Medical mT5 versions: a 770M\footnote{\url{https://hf.co/HiTZ/Medical-mT5-large}} and 3B\footnote{\url{https://hf.co/HiTZ/Medical-mT5-xl}} parameter text-to-text open-source models which obtain state-of-the-art results in multilingual sequence labeling for the medical domain, most notably in multi-task and zero-shot cross-lingual settings.
\end{itemize}

Other benefits of our Medical mT5 models include the comparatively low hardware requirements needed for both fine-tuning on downstream tasks (the large 770M version easily fits in a 24GB GPU) and for inference (a 12GB GPU should be enough). As an example, a LLaMA 7B model (\cite{wu2023pmcllama}) requires at least 4 80GB A100 GPUs. This makes our models more accessible to the research community and to small and medium-sized companies.

\section{Related Work}

As it has been the case in most application domains, Large Language Models (LLMs) have facilitated significant improvements in the state-of-the-art for medical NLP tasks (\cite{singhal-palm,wu2023pmcllama,mayer2021enhancing}). The most popular approaches use models pre-trained on medical corpora such as SciBERT (\cite{beltagy2019scibert}), BioBERT (\cite{DBLP:journals/bioinformatics/LeeYKKKSK20}), PubmedBERT (\cite{DBLP:journals/health/GuTCLULNGP22}), BSC-BIO (\cite{carrino-etal-2022-pretrained}), or BioLinkBERT (\cite{DBLP:conf/acl/YasunagaLL22}).

While the previous encoder-only models focused on discriminative tasks, the emergence of generative models such as LLaMa (\cite{touvron2023llama}), PaLM (\cite{singhal-palm}), and GPT-3 (\cite{brown2020language}) has generated significant interest in adapting such LLMs to the medical domain. These models include, but are not limited to, SciFive (\cite{DBLP:journals/corr/abs-2106-03598}), an English T5 encoder-decoder model adapted to the scientific domain, and decoder models such as BioGPT (\cite{10.1093/bib/bbac409}), Med-PaLM (\cite{singhal-palm}), PMC-LLaMA (\cite{wu2023pmcllama}), and ClinicalGPT (\cite{Wang2023ClinicalGPTLL}). 

Additionally, a range of abstractive question-answering tasks has been proposed as evaluation benchmarks, on which the larger models (\cite{wu2023pmcllama,singhal-palm,Wang2023ClinicalGPTLL}) achieve the best results. While interesting, both these LLMs and benchmarks have been developed with a focus on a single language, usually English. 

Furthermore, these LLMs require hardware that is simply not affordable for the majority of end-users and researchers. To address these issues, we propose Medical mT5, a multilingual text-to-text model adapted to the medical domain which, despite its relatively modest size and low running costs, obtains competitive results, notably in multi-task and zero-shot cross-lingual settings.

\section{Compiling a Multilingual Corpus for the Medical Domain}\label{sec7:corpus}

\begin{wraptable}{r}{0.45\textwidth}
    \centering
    \vspace{-3.0em}
    \intextsep=0pt
    \adjustbox{max width=0.94\linewidth,max totalheight=0.9\textheight}{
    \begin{tabular}{@{}l|l|r@{}}
    \toprule
    \textbf{Language} & \textbf{Source} & \textbf{Words} \\
    \midrule
    \multirow{4}{*}{English} & ClinicalTrials & 127.4M \\
     & EMEA & 12M \\
     & PubMed & 968.4M \\
     & \textbf{Total} & \textbf{1.1B} \\
    \midrule 
    \multirow{7}{*}{Spanish} & EMEA & 13.6M \\
     & PubMed & 8.4M \\
     & Medical Crawler & 918M \\
     & SPACC & 350K \\
     & UFAL & 10.5M \\
     & WikiMed & 5.2M \\
     & \textbf{Total} & \textbf{960M} \\
    \midrule
    \multirow{6}{*}{French} & PubMed & 1.4M \\
     & Science Direct & 15.2M \\
     & Wikipedia - Médecine & 5M \\
     & EDP & 48K \\
     & Google Patents & 654M \\
    & \textbf{Total} & \textbf{676M} \\
    \midrule
    \multirow{14}{*}{Italian} & Medical Commoncrawl - IT & 67M \\
     & Drug instructions & 30.5M \\
     & Wikipedia - Medicina & 13.3M \\
     & E3C Corpus - IT & 11.6M \\
     & Medicine descriptions & 6.3M \\
     & Medical theses & 5.8M \\
     & Medical websites & 4M \\
     & PubMed & 2.3M \\
     & Supplement description & 1.3M \\
     & Medical notes & 975K \\
     & Pathologies & 157K \\
     & Medical test simulations & 26K \\
     & Clinical cases & 20K \\
     & \textbf{Total} & \textbf{143M} \\
     \midrule
    \textbf{Total} & & \textbf{3.02B} \\
    \bottomrule
    \end{tabular}
    }
    \caption{Data sources and word counts by language.}
    \label{tab7:merged-data}
    \vspace{-4.0em}
    \end{wraptable}

Obtaining good quality medical corpora is usually difficult due to the sensitive nature of the data. This is even more challenging for non-English languages, as the availability of data for other languages is in general more restricted. Despite these issues, we have successfully gathered and curated a diverse collection of public relevant corpora of medical texts in English, French, Italian and Spanish to generate the Medical mT5 model. The data sources are summarized in Table \ref{tab7:merged-data}.

\subsection{English}

As listed in table \ref{tab7:merged-data}, we collected around 1B words from three sources related to the medical domain: (i) \textbf{ClinicalTrials} is a set of documents of clinical studies from all over the world; (ii) \textbf{EMEA} is an English-Spanish parallel corpus with documents provided by the European Medicines Agency (\cite{TIEDEMANN12.463}) and, (iii) \textbf{PubMed}, which contains data from various sources such as MEDLINE, life science journals and online books, provides the bulk of the English data.

\subsection{Spanish}

Apart from \textbf{EMEA} and \textbf{PubMed}, which we also used for Spanish, the biggest portion of the data came from the \textbf{Medical Crawler}, a biomedical corpus compiled by \citet{carrino-etal-2022-pretrained}. Additionally, we also included \textbf{SPACC}, \textbf{UFAL} and \textbf{WikiMed}, a corpus built ad-hoc from Wikipedia entries. Table \ref{tab7:merged-data} provides the details of the collected data, which amounts to $\approx$1B words.

\subsection{French}

A total of 7,192,779 sentences and 670,972,717 words were compiled using the data sources listed in Table~\ref{tab7:merged-data}. \textbf{Science Direct} offers a collection of scientific and medical publications. We filtered relevant articles with the keyword ``Médecine'', and the obtained XML documents were parsed to extract the \texttt{<dc:description>} tag.
As for Spanish, we took advantage of \textbf{Wikipedia} and \textbf{PubMed} as a source of medical knowledge. PubMed data was extracted using the \texttt{Bio.Entrez} package\footnote{\url{https://biopython.org/docs/1.75/api/Bio.Entrez.html}}. For wikipedia we obtain HTML formatted data from the category ``Category:Médecine''. The \textbf{EDP French/English Parallel Medical Corpus}~(\cite{DBLP:conf/wmt/Jimeno-YepesNNV17}) provides bilingual content from journals that address domains such as dentistry and life sciences. From this source, we downloaded the dataset labeled ``EDP French corpus, text format''. Finally, \textbf{Google Patents} is a comprehensive repository of patent data from around the world. Google Patents data were retrieved by filtering using the IPC code and abstract language. A final French language verification step was undertaken by applying the \texttt{langdetect} package (version 1.0.9).

\subsection{Italian}

The crawling and pre-processing of the Italian split of the corpus followed the methodology described by \citet{carrino-etal-2022-pretrained}. First, we compiled a list of 504 medical terms, which we use as seeds to scrape the Italian split of the \textbf{MC4 Common Crawl Corpus} by only selecting the pages which contained at least one of the keywords in their URL domain. To create the list, we extracted 600 keyword terms related to medicine from the \textit{Dizionario analogico della Lingua Italiana} (Zanichelli). We excluded some sectors and discarded terms that may lead to ambiguous queries (e.g., actions, which contained mainly verbs, proverbs, general terms like ``assistente'', etc.). We normalized rare variants (``bacteriologia'' to ``batteriologia'') and stemmed all terms without lemmatizing, as most terms are already lemmatized in the dictionary; we performed univerbation of multiword units (e.g., ``esamedelleurine'', ``follow-up''), and removed the duplicates. This resulted in a corpus of 67 million tokens, which we joined with other sources of text such as \textbf{Medical dissertations}, \textbf{Drug use instructions}, \textbf{PubMed abstracts}, etc. as detailed in Table \ref{tab7:merged-data}, resulting in a $\approx$145M word corpus. 

\newcommand{\customsectiontitle}{
  \includegraphics[width=0.7cm]{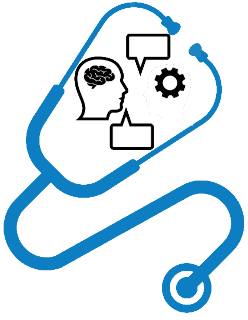} Medical mT5
}
\section[Medical mT5]{\customsectiontitle}
\label{sec7:medicalmt5}

Multilingual T5 (mT5) (\cite{mt5}) is an extension of the original T5 (\cite{DBLP:journals/jmlr/RaffelSRLNMZLL20-T5}) framework, which is optimized for multilingual tasks. The T5 model is grounded in the Transformer encoder-decoder architecture (\cite{DBLP:conf/nips/VaswaniSPUJGKP17}). With its decoder block, T5 is capable of generating sequences of tokens in an auto-regressive fashion. T5 was designed to convert every NLP problem into a text-to-text task, and mT5 extends this strategy to a multitude of languages, leveraging a shared vocabulary for diverse scripts. mT5 was trained using mC4, a 1 Trillion token Common Crawl-based dataset covering 101 languages. The pre-training is based on a masked language modeling ``span-corruption'' objective, where consecutive spans of input tokens are replaced with a mask and the model is trained to reconstruct the masked-out tokens.

\subsection{Pre-training Medical mT5}

\begin{table}[htb]
\centering
\adjustbox{max width=0.8\linewidth,max totalheight=0.8\textheight}{
\begin{tabular}{@{}lrr@{}}
\toprule
 & Medical-mT5-large & Medical-mT5-xl \\ \midrule
Param. no. & 738M & 3B \\
Sequence Lenght & 1024 & 480 \\
Token/step & 65536 & 30720 \\
Epochs & 1 & 1 \\
Total Tokens & 4.5B & 4.5B \\
Optimizer & Adafactor & Adafactor \\
LR & 0.001 & 0.001 \\
Scheduler & Constant & Constant \\
Hardware & 4xA100 & 4xA100 \\
Time (h) & 10.5 & 20.5 \\
CO\textsubscript{2}eq (kg) & 2.9 & 5.6 \\ \bottomrule
\end{tabular}}
\caption{Pre-Training settings for Medical mT5.}
\label{tab7:PreTraining}
\end{table}

Medical mT5 is built upon the same architecture as mT5 (\cite{mt5}). We release two diffent models: Medical-mT5-large (738M parameters) and Medical-mT5-xl (3 billion parameters). Both models were initialized using the pre-trained weights of their corresponding mT5 checkpoints and continued their pre-training using the 3B word medical domain dataset described in Section \ref{sec7:corpus} (with x2 oversampling for the Italian split). To prevent over-fitting, we run the training for only one epoch, as preliminary experiments showed that performance degraded with more epochs. We adhered to the self-supervised parameter settings described in \citet{mt5} and detailed in Table \ref{tab7:PreTraining}. It should be noted that Medical-mT5-large was trained with a sequence length of 1024 tokens whereas Medical-mT5-xl was limited to a sequence length of 480 tokens due to GPU memory limitations. Medical mT5 was trained using the Flax implementation of mT5 in the Hugging Face Transformers library (\cite{wolf-etal-2020-huggingface-transformers}). All experiments were conducted on our private servers, employing 4xA100 80GB GPUs. We made calculations for a carbon footprint estimation based on a 400W consumption per GPU and a carbon intensity of 0.171 kg/kWh\footnote{Sourced from \url{https://app.electricitymaps.com/map}}.

\section[Generating New Multilingual Benchmarks]{Generating New Multilingual Benchmarks: Real-World Application of Data Transfer}\label{sec7:new-benchmarks}

There is a lack of multilingual evaluation benchmarks for the medical domain. The only available benchmark in English, Spanish, French, and Italian is the relatively small e3C (\cite{e3c}). While medical domain evaluation datasets are scarce for Spanish, French and Italian, many datasets exist for English. Therefore, this is a good opportunity to apply the data transfer techniques developed in previous chapters to generate data for other languages. We focused on two different types of tasks: (i) a sequence labeling task, \textbf{Argument Mining}, which involves detecting and classifying the argument component spans and their relations, and (ii) \textbf{Abstractive Question Answering}, where the model is expected to generate an answer in response to an input question. In both cases we used existing labeled English data as a starting point.

\subsection{Argument Mining}

\begin{figure}
    \centering
    \includegraphics[width=\textwidth]{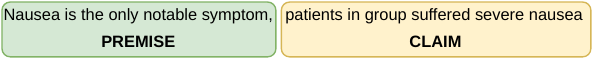}
    \caption{Example of an annotated abstract from the AbstRCT dataset.}
    \label{fig7:abstrct_example}
\end{figure}

The AbstRCT dataset is composed by English medical and scientific texts collected from the MEDLINE database and manually annotated with two types of argument components: Claims and Premises (\cite{mayer2021enhancing}). An example of the task is illustrated in Figure \ref{fig7:abstrct_example}

A `claim'  is a concluding statement made by the author about the outcome of the study. In the medical domain, it may be an assertion of a diagnosis or a treatment. A `premise' corresponds to an observation or measurement in the study (ground truth), which supports or attacks another argument component, usually a claim. It is important that they are observed facts and, therefore, credible without further evidence.

\begin{figure}[htbp]
  \centering
  \includegraphics[width=\textwidth]{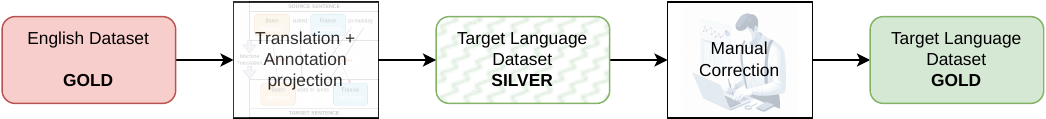}
  \caption{Data construction process for generating the Spanish, French and Italian versions of the AbstRCT dataset.}
  \label{fig7:dataconstruction}
\end{figure}

We generated French and Italian parallel versions of the dataset using the same method as for Spanish in \cite{yegingbergenova-cross}. First, the English dataset is translated into the target language using the machine translation model NLLB200-3.3B (\cite{DBLP:journals/corr/abs-2207-04672}). Then, the labels in the source language are transferred to the target language using AWESOME align (\cite{dou-neubig-2021-word}) and the annotation projection algorithm developed in Chapter \ref{ch:model-vs-data}. Finally, to ensure the quality of the generated dataset, the projections are manually reviewed by an expert in the target language. Thanks to this process, the manual annotation labor is significantly reduced compared to annotating the data from scratch. This process is illustrated in Figure \ref{fig7:dataconstruction}.

The AbstRCT dataset is divided into three splits, neoplasm, glaucoma and mixed. Following previous work, we fine-tune the models with the first one and then evaluate the in-domain performance on the neoplasm test split and the cross-domain performance on the glaucoma and mixed splits. Previous works using the AbstRCT datasets have employed different definitions of the $F_1$ score metric, such as token-level $F_1$ (\cite{mayer2021enhancing,yegingbergenova-cross}). However, in this paper, we report results using the standard sequence level $F_1$ score (\cite{DBLP:conf/conll/SangM03}), a much more strict metric, which explains the lower results for all the models.

\subsection{Question Answering}\label{sec:QA_explained}

We use the BioASQ-6B English Question Answering dataset (\cite{bioasq}) to generate parallel French, Italian and Spanish versions. Given a biomedical question and a set of snippets of text with relevant information about the question, the model must generate the \textit{ideal} answer. This task is similar to the Retrieval Augmented Generation (RAG) task (\cite{NEURIPS2020_6b493230}), where the model must generate an answer given a context or set of contexts. A set of ideal gold answers are provided to assess the performance of the models. We machine-translated the questions and ideal answers into French, Italian and Spanish using the NLLB200 3B parameter model (\cite{DBLP:journals/corr/abs-2207-04672}). In this case, as this is not a sequence labeling task, no annotation projection is needed. Nevertheless, the quality of a small set of translations was manually reviewed to ensure the quality of the generated data. 

\section{Experimental Setup}

In this section we describe the datasets and evaluation tasks used to measure the performance of Medical mT5. We also provide the details of the training and evaluation process, and baseline models used for comparison.

\subsection{Datasets}

\begin{table}[htb]
    \centering
    \small
    \adjustbox{max width=\textwidth}{
    \begin{tabular}{@{}ccccc@{}}
    \toprule
    Representation & Task & Dataset & Languages & Entity Type \\ \midrule
     & \cellcolor{ForestGreen!10} & \cellcolor{ForestGreen!10}NCBI-Disease, \cite{ncbi-disease} & \cellcolor{ForestGreen!10}EN & \cellcolor{ForestGreen!10}Disease \\
     & \cellcolor{ForestGreen!10} & \cellcolor{ForestGreen!10}BC5CDR Disease, \cite{bc5cdr} & \cellcolor{ForestGreen!10}EN & \cellcolor{ForestGreen!10}Disease \\
     & \cellcolor{ForestGreen!10} & \cellcolor{ForestGreen!10}BC5CDR Chemical, \cite{bc5cdr} & \cellcolor{ForestGreen!10}EN & \cellcolor{ForestGreen!10}Chemical \\
     & \cellcolor{ForestGreen!10} & \cellcolor{ForestGreen!10}DIANN, \cite{diann} & \cellcolor{ForestGreen!10}EN, ES & \cellcolor{ForestGreen!10}Disability \\
     & \cellcolor{ForestGreen!10} & \cellcolor{ForestGreen!10}E3C, \cite{e3c} & \cellcolor{ForestGreen!10}EN, ES, FR, IT & \cellcolor{ForestGreen!10}Clinical Entity \\
     & \multirow{-6}{*}{\cellcolor{ForestGreen!10}\begin{tabular}[c]{@{}c@{}}Named Entity \\ Recognition\end{tabular}} & \cellcolor{ForestGreen!10}PharmaCoNER, \cite{pharmaconer} & \cellcolor{ForestGreen!10}ES & \cellcolor{ForestGreen!10}Pharmacological \\
    \multirow{-7}{*}{\begin{tabular}[c]{@{}c@{}}Sequence\\ Labelling\end{tabular}} & \cellcolor{CornflowerBlue!10}\begin{tabular}[c]{@{}c@{}}Argument \\ Mining\end{tabular} & \cellcolor{CornflowerBlue!10}AbstRCT, \cite{mayer2021enhancing} & \cellcolor{CornflowerBlue!10}EN, ES, FR, IT & \cellcolor{CornflowerBlue!10}Claims and Premises \\ \midrule
    \begin{tabular}[c]{@{}c@{}}Generative \\ Question \\ Answering\end{tabular} & \cellcolor{Yellow!10}\begin{tabular}[c]{@{}c@{}}Question \\ Answering\end{tabular} & \cellcolor{Yellow!10}BioASQ 6B, \cite{bioasq} & \cellcolor{Yellow!10}EN, ES, FR, IT & \cellcolor{Yellow!10}Biomedical QA \\ \bottomrule
    \end{tabular}}
    \caption{List of evaluation tasks used to measure the performance of Medical mT5.}
    \label{tab7:tasks}
    
\end{table}

The list of tasks used for evaluation is listed in Table \ref{tab7:tasks}. The \textbf{Sequence labeling tasks} include medical NER, detecting and classifying named entities according to some pre-defined categories, and Argument Mining, described in Section \ref{sec7:new-benchmarks}. Performance for every sequence labeling task is evaluated using standard sequence level $F_1$ score (\cite{DBLP:conf/conll/SangM03}). 
We also evaluate the performance of Medical mT5 on the \textbf{Generative Question Answering task} using the BioASQ dataset, described in Section \ref{sec:QA_explained}. 

\subsection{Conversion to Text-to-Text Format}

Medical mT5 is a text-to-text model. This means that, given a text input, it learns to generate a text as output. Therefore, every evaluation task must be converted into a text-to-text format (\cite{mt5}). In our experiments the output text is always generated using beam search with 4 beams.

\begin{figure}[htbp]
  \centering
  \includegraphics[width=0.8\linewidth]{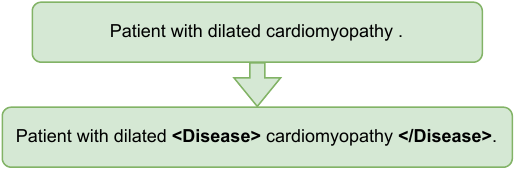}
  \caption{Text-to-Text representation of the Sequence Labeling task. Given an input sentence, the model is expected to generate the same sentence annotated with html-style tags.}
  \label{fig7:SL}
\end{figure}

To address sequence labeling tasks, we use the same approach presented in Chaper \ref{ch:model-transfer}. As illustrated in Figure \ref{fig7:SL}, Text-to-text models such as Medical mT5 are prompted with the sentence to label. The expected output is the same sentence annotated with HTML-style tags. The HTML tags for each task are added as special tokens to the model vocabulary. We use constrained decoding to ensure that the output contains the same words as the input and a valid HTML annotation. The constrined decoding algorithm is the one presented in Chapter \ref{ch:model-transfer}.

\begin{figure}[htbp]
  \centering
  \includegraphics[width=0.7\linewidth]{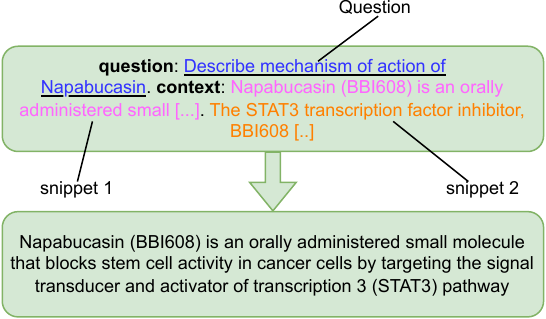}
  \caption{Text-to-Text representation of the BioASQ task. Given a question and a set of relevant snippets, the model generates an answer.}
  \label{fig:BioASQ}
\end{figure}

With respect to the BioASQ \textbf{Abstractive Question Answering task}, the input prompt contains the question and a context. As shown in Figure \ref{fig:BioASQ}, the context is generated by concatenating all the provided possible snippets. The expected output should be the generated answer to the question, which is then compared to the gold ideal answer. 

\subsection{Baselines}

As we have developed Medical mT5 by continuing the training of mT5 checkpoints, our primary point of comparison should be mT5 (\cite{mt5}). Thus, our first objective
is to assess whether training the model on our multilingual medical-domain
corpus enhances its performance for tasks specific to this domain. Furthermore, we also benchmark our model against SciFive (Pubmed+PMC) a T5-based 738M parameter model
(\cite{DBLP:journals/corr/abs-2106-03598}) trained exclusively on a corpus of 78B words containing scientific and medical English data. Additionally, we compare the performance of Medical mT5 with Flan-T5 (\cite{chung-flan-instruction-models}), which also adopts the T5 architecture but has been finetuned on a huge instruction-following dataset for almost 2K tasks. Flan-T5 achieves state-of-the-art performance in numerous benchmarks, including some from the medical domain (\cite{singhal-palm}). We tested all three types of text-to-text models under identical settings and hyperparameters.

We also measure Medical mT5 with the performance of encoder-only models in sequence labeling tasks. We report results with mDeBERTaV3 (\cite{DBLP:conf/iclr/HeLGC21/deberta}) which is widely used for sequence labeling and excels in multilingual tasks (\cite{adelani-etal-2022-masakhaner,Agerri2022LessonsLF}). Although we also tested XLM-RoBERTa (\cite{conneau-etal-2020-unsupervised}) and GLOT500 (\cite{DBLP:conf/acl/ImaniLKSSKMSMYS23}), their results were worse than those obtained by mDeBERTaV3. Finally, we also compare with BioBERT v1.1 (\cite{DBLP:journals/bioinformatics/LeeYKKKSK20}), which has been pretrained on a large English-only biomedical dataset. We do not evaluate the performance of encoder-only models in the question-answering task, as their architecture is not designed for text generation.

\subsection{Hyperparameters settings}

For sequence labeling, when using encoder-decoder models, we use a learning rate of $1 \times 10^{-4}$, a batch size of 8, and a maximum sequence length of 256 tokens. We use the Adafactor optimizer (\cite{DBLP:conf/icml/ShazeerS18}) with cosine learning rate decay to 0 and 500 warmup steps The number of epochs varies depending on the task. For the E3C dataset, which is very small, we use 100 epochs. For the other datasets, we use 45 epochs. When training the model in a multi-task setting, we use 12 epochs. For the question-answering task, we use 15 epochs. We use a beam size of 4 for all the tasks and no sampling. Models are trained using bfloat16 precision.

For encoder-only models, we use a batch size of 32, a learning rate of $5 \times 10^{-5}$, and 40 epochs. We use the AdamW optimizer (\cite{DBLP:journals/corr/abs-1711-05101}) with a cosine learning rate scheduler that decays the learning rate to 0. We use a maximum sequence length of 256 tokens. Encoder-only models are trained using fp16 precision.

For all models, we evaluate the model during training on the validation set periodically and select the model at the epoch with the highest performance on the validation set.

\section{Experimental Results}
In this section we present the evaluation results of Medical mT5 on Sequence Labeling and Question Answering tasks.
\subsection{Sequence labeling Tasks}
In this section we report on the performance of Medical mT5 and of the baselines in the
\textbf{sequence labeling tasks} across different settings.

\begin{table}[htb]
\centering
\small
\adjustbox{max width=0.98\linewidth,max totalheight=0.65\textheight}{
\begin{tabular}{@{}llccccccc|cc@{}}
Lang & Dataset & \rotatebox{90}{mT5\textsubscript{large}} & \rotatebox{90}{mT5\textsubscript{XL}} & \rotatebox{90}{SciFive} & \rotatebox{90}{FlanT5\textsubscript{large}} & \rotatebox{90}{FlanT5\textsubscript{XL}} & \rotatebox{90}{mDeBERTa\textsubscript{V3 base}} & \rotatebox{90}{BioBERT} & \rotatebox{90}{MedMT5\textsubscript{large}} & \rotatebox{90}{MedMT5\textsubscript{XL}} \\ 
\midrule
EN & NCBI-Disease & 85.1 & 87.7 & \textbf{89.4} & 88.6 & 89.3 & 85.7 & 87.4 & 89.1 & 87.2 \\ \midrule
EN & BC5CDR Disease & 78.5 & 81.4 & 85.4 & 85.0 & \textbf{85.8} & 82.5 & 84.3 & 84.4 & 82.4 \\
EN & BC5CDR Chemical & 89.1 & 90.8 & \textbf{93.3} & 92.0 & 92.9 & 91.1 & 92.9 & 92.8 & 91.3 \\ \midrule
EN & DIANN & 70.1 & 77.8 & 71.9 & 74.4 & 74.2 & \textbf{80.3} & 79.0 & 74.8 & 77.6 \\
\rowcolor{CornflowerBlue!15}ES & DIANN & 72.4 & 74.9 & 70.5 & 70.7 & 70.9 & \textbf{78.3} & 70.2 & 74.9 & 74.8 \\ \midrule
EN & E3C & 54.3 & 60.1 & 62.8 & \textbf{64.2} & 63.1 & 58.2 & 58.6 & 59.4 & 57.9 \\
\rowcolor{CornflowerBlue!15}ES & E3C & 61.6 & 71.7 & 62.7 & 64.4 & 67.1 & 65.9 & 57.4 & \textbf{72.2} & 69.5 \\
\rowcolor{CornflowerBlue!15}FR & E3C & 55.6 & 64.9 & 61.7 & 65.2 & 64.3 & 62.0 & 53.3 & 65.2 & \textbf{65.8} \\
\rowcolor{CornflowerBlue!15}IT & E3C & 61.8 & 63.8 & 59.6 & 61.9 & 65.1 & 63.9 & 52.1 & \textbf{67.5} & 65.9 \\ \midrule
\rowcolor{CornflowerBlue!15}ES & PharmaCoNER & 86.3 & 90.6 & 87.5 & 88.5 & 89.1 & 89.4 & 88.6 & \textbf{90.8} & 90.1 \\ \midrule
EN            & Neoplasm               & 70.4      & 71.1   & 74.4          & \textbf{74.3} & 73.4          & 64.5            & 67.5 & 73.9          & 73.2          \\
EN            & Glaucoma               & 70.7      & 75.1   & 77.1          & \textbf{78.4} & 78.0          & 71.2            & 74.8 & 76.2          & 76.4          \\
EN            & Mixed                  & 68.5      & 73.0   & 73.4          & 73.2          & \textbf{74.5} & 63.4            & 69.6 & 72.2          & 72.0          \\
\rowcolor{CornflowerBlue!15}ES            & Neoplasm               & 69.0      & 56.1   & 71.4          & 72.5          & \textbf{73.9} & 63.0            & 57.1 & 72.1          & 71.8          \\
\rowcolor{CornflowerBlue!15}ES            & Glaucoma               & 69.3      & 70.7   & 73.9          & 73.8          & 75.2          & 68.6            & 64.5 & \textbf{77.1} & 75.5          \\
\rowcolor{CornflowerBlue!15}ES            & Mixed                  & 68.4      & 66.2   & 69.2          & 69.3          & 71.6          & 61.3            & 58.9 & \textbf{72.4} & 71.4          \\
\rowcolor{CornflowerBlue!15}FR            & Neoplasm               & 70.5      & 66.6   & \textbf{74.0} & 72.4          & 73.7          & 63.9            & 59.0 & 72.9          & 71.2          \\
\rowcolor{CornflowerBlue!15}FR            & Glaucoma               & 71.1      & 69.2   & 77.8          & 74.8          & 77.2          & 60.3            & 65.6 & \textbf{79.5} & 75.8          \\
\rowcolor{CornflowerBlue!15}FR            & Mixed                  & 68.3      & 65.4   & 72.0          & 70.9          & \textbf{74.3} & 64.1            & 61.3 & 73.3          & 69.7          \\
\rowcolor{CornflowerBlue!15}IT            & Neoplasm               & 68.1      & 69.9   & 70.1          & 70.9          & 72.0          & 64.4            & 54.8 & 71.2          & \textbf{73.1} \\
\rowcolor{CornflowerBlue!15}IT            & Glaucoma               & 69.2      & 71.5   & 73.7          & 74.0          & 75.9          & 74.7            & 65.8 & 75.7          & \textbf{78.7} \\
\rowcolor{CornflowerBlue!15}IT            & Mixed                  & 66.3      & 67.7   & 67.4          & 69.9          & 70.0          & 61.3            & 57.4 & 70.6          & \textbf{71.9} \\ \midrule
\rowcolor{ForestGreen!10}\multicolumn{2}{c}{AVERAGE}            & 70.2      & 72.1   & 73.6          & 74.1          & 75.1          & 69.9            & 67.3                     & \textbf{75.4} & 74.7          \\
\rowcolor{ForestGreen!10}\multicolumn{2}{l}{AVERAGE ES, FR, IT} & 68.4      & 69.2   & 70.8          & 71.4          & 72.9          & 67.2            & 61.9                     & \textbf{74.0} & 73.2   \\ \bottomrule

\end{tabular}
}
\caption{Single-task supervised F1 scores for Sequence Labelling.}
\label{tab7:SingleTask}
\end{table}
\paragraph{Single Task Monolingual Supervised Results:} The results when fine-tuning and evaluating the models for each dataset and language are shown in Table \ref{tab7:SingleTask}.  The first observation is that Medical-mT5-large significantly outperforms both mT5-large and mT5-XL, demonstrating the benefits of further training these models with our multilingual medical domain corpus.

When comparing Medical mT5 with FlanT5 and SciFive, the latter models are systematically superior on English. This was anticipated since both have been pre-trained with a much larger amount of English-only data specific to the medical domain. With respect to encoder-only models, they achieve in general worse results than text-to-text models across all tasks and languages (except for the DIANN dataset). It is also noteworthy that FlanT5-XL exhibits robust performance across all datasets and languages, even though it was fine-tuned with English-only data not specific to the medical domain. Nonetheless, Medical-mT5-large obtains in general better results for French, Spanish and Italian while being much smaller in size (738M parameters vs 3B parameters), showing the impact of training Medical mT5 with domain-specific data for those languages.

\begin{table}[htb]
\centering
\small
\adjustbox{max width=0.98\linewidth,max totalheight=0.9\textheight}{
\begin{tabular}{@{}llccc|ccc@{}}
\toprule
\multirow{2}{*}{Lang} & \multirow{2}{*}{Dataset} & \multicolumn{3}{c}{Single Task} & \multicolumn{3}{c}{MultiTask} \\ 
 &  &  FlanT5\textsubscript{XL} & MedMT5\textsubscript{large} & MedMT5\textsubscript{XL} &  FlanT5\textsubscript{XL} & MedMT5\textsubscript{large} & MedMT5\textsubscript{XL} \\ \midrule
EN & NCBI-Disease & \textbf{89.3} & 89.1 & 87.2 & 87.6 & 87.6 & 86.9 \\ \midrule
EN & BC5CDR Disease & \textbf{85.8} & 84.4 & 82.4 & 85.1 & 83.4 & 83.0 \\
EN & BC5CDR Chemical & \textbf{92.9} & 92.8 & 91.3 & 92.7 & 92.5 & 91.6 \\ \midrule
EN & DIANN & 74.2 & 74.8 & 77.6 & \textbf{80.0} & 75.4 & 75.3 \\
\rowcolor{CornflowerBlue!15}ES & DIANN & 70.9 & 74.9 & 74.8 & \textbf{77.1} & 72.6 & 73.6 \\ \midrule
EN & E3C & \textbf{63.1} & 59.4 & 57.9 & 62.1 & 60.9 & 62.0 \\
\rowcolor{CornflowerBlue!15}ES & E3C & 67.1 & 72.2 & 69.5 & 66.5 & \textbf{74.9} & 73.3 \\
\rowcolor{CornflowerBlue!15}FR & E3C & 64.3 & 65.2 & \textbf{65.8} & 62.9 & 65.4 & 65.1 \\
\rowcolor{CornflowerBlue!15}IT & E3C & 65.1 & \textbf{67.5} & 65.9 & 60.7 & 66.9 & 65.1 \\ \midrule
\rowcolor{CornflowerBlue!15}ES & PharmaCoNER & 89.1 & \textbf{90.8} & 90.1 & 89.9 & 90.3 & 89.5 \\ \midrule
EN & Neoplasm & 73.4 & \textbf{73.9} & 73.2 & 73.1 & 72.3 & 72.9 \\
EN & Glaucoma & \textbf{78.0} & 76.2 & 76.4 & 76.4 & 76.8 & 77.5 \\
EN & Mixed & \textbf{74.5} & 72.2 & 72.0 & 71.5 & 70.9 & 73.0 \\
\rowcolor{CornflowerBlue!15}ES & Neoplasm & \textbf{73.9} & 72.1 & 71.8 & 73.5 & 73.5 & 73.7 \\
\rowcolor{CornflowerBlue!15}ES & Glaucoma & 75.2 & 77.1 & 75.5 & 77.1 & 77.7 & \textbf{79.3} \\
\rowcolor{CornflowerBlue!15}ES & Mixed & 71.6 & 72.4 & 71.4 & 70.0 & 71.8 & \textbf{72.8} \\
\rowcolor{CornflowerBlue!15}FR & Neoplasm & 73.7 & 72.9 & 71.2 & \textbf{74.0} & 72.9 & 73.6 \\
\rowcolor{CornflowerBlue!15}FR & Glaucoma & 77.2 & \textbf{79.5} & 75.8 & 76.6 & 77.0 & 79.4 \\
\rowcolor{CornflowerBlue!15}FR & Mixed & \textbf{74.3} & 73.3 & 69.7 & 71.8 & 71.2 & 73.0 \\
\rowcolor{CornflowerBlue!15}IT & Neoplasm & 72.0 & 71.2 & 73.1 & 71.9 & \textbf{74.6} & 74.0 \\
\rowcolor{CornflowerBlue!15}IT & Glaucoma & 75.9 & 75.7 & 78.7 & 77.6 & 78.5 & \textbf{78.9} \\
\rowcolor{CornflowerBlue!15}IT & Mixed & 70.0 & 70.6 & 71.9 & 69.9 & 72.5 & \textbf{73.3} \\ \midrule
\rowcolor{ForestGreen!10}\multicolumn{2}{c}{AVERAGE} & 75.1 & 75.4 & 74.7 & 75.2 & 76.2 & \textbf{76.7} \\
\rowcolor{ForestGreen!10}\multicolumn{2}{c}{AVERAGE ES, FR, IT} & 72.9 & 74.0 & 73.2 & 73.1 & 74.8 & \textbf{75.3} \\ \bottomrule
\end{tabular}
}
\caption{Multi-task supervised F1 scores for Sequence Labelling.}
\label{tab7:MultiTask}
\end{table}

\paragraph{Multi-Task Supervised Results:} Text-to-text models have
demonstrated improved performance when trained in multi-task settings
(\cite{chung-flan-instruction-models}). Following this, we also experimented with fine-tuning them across all the sequence labeling tasks simultaneously. To inform the model about which labels should be classified for each input example, we add the list of predefined labels from the corresponding dataset to the beginning of the input sentence. For instance, the input depicted in Figure \ref{fig7:SL} is adjusted to \textit{``<Disease> Patient with dilated cardiomyopathy''}. A comparison of the Single Task and Multi-Task settings is presented in Table \ref{tab7:MultiTask}. It can be seen that in this setting Medical mT5 achieves the best overall results for Spanish, French and Italian. On average, Medical-mT5-xl also obtains the best performance, slightly improving over the results of FlanT5-XL and Medical-mT5-large.

\begin{table}[htb]
\centering
\small
\adjustbox{max width=0.98\linewidth,max totalheight=0.9\textheight}{
\begin{tabular}{@{}llcccc|cc@{}}
\toprule
Lang & Dataset & mT5\textsubscript{XL} & SciFive & FlanT5\textsubscript{XL} & mDeBERTa\textsubscript{V3 base} & MedMT5\textsubscript{large} & MedMT5\textsubscript{XL} \\ \midrule
ES         & Neoplasm       & 71.4          & 69.8    & 67.9      & 65.1            & \textbf{72.4} & 71.7          \\
ES         & Glaucoma       & \textbf{74.1} & 71.5    & 70.6      & 68.3            & 72.4          & 73.2          \\
ES         & Mixed          & \textbf{69.4} & 67.0    & 66.7      & 60.9            & 68.1          & 68.8          \\
FR         & Neoplasm       & 71.6          & 68.6    & 69.9      & 60.5            & 72.4          & \textbf{72.8} \\
FR         & Glaucoma       & 75.8          & 74.5    & 71.0      & 68.7            & 72.3          & \textbf{76.7} \\
FR         & Mixed          & \textbf{73.0} & 68.5    & 68.2      & 59.3            & 70.4          & 72.4          \\
IT         & Neoplasm       & 70.6          & 63.1    & 67.3      & 62.4            & 72.9          & \textbf{73.2} \\
IT         & Glaucoma       & 76.7          & 71.6    & 72.0      & 70.2            & 75.4          & \textbf{79.0} \\
IT         & Mixed          & 69.9          & 62.5    & 66.9      & 62.1            & 71.7          & \textbf{71.9} \\ \midrule
\rowcolor{ForestGreen!10} \multicolumn{2}{c}{AVERAGE} & 72.5          & 68.6    & 69.0      & 64.2            & 72.0          & \textbf{73.3} \\ \bottomrule
\end{tabular}
}
\caption{Zero-shot F1 scores for Argument Mining. Models have been trained in English and evaluated in Spanish, French and Italian.}
\label{tab7:ZeroShot}
\end{table}

\paragraph{Zero-shot Cross-Lingual Transfer Results:} Manually annotated medical domain datasets for languages other than English are scarce. Therefore, developing models that can successfully generate predictions for languages different to those used for fine-tuning is crucial. We evaluate this ability to perform zero-shot cross-lingual transfer by fine-tuning Medical mT5 and the baselines on the English AbsRCT Neoplasm dataset, and then evaluating them on the Neoplasm, Glaucoma, and Mixed datasets for Spanish, French, and Italian. The results are presented in Table \ref{tab7:ZeroShot}. Results show that Medical mT5 outperforms any other model. Moreover, Medical-mT5-xl achieves significantly better results than Medical-mT5-large. 

To summarize, Medical mT5 stands out for its superior performance in the evaluation for Spanish, French, and Italian languages, especially for the multitask and the zero-shot transfer settings. These capabilities can help mitigate the scarcity of manually annotated medical data for other target languages. In contrast, SciFive and FlanT5, having been trained on extensive English-only datasets, emerge as the top choices when the primary focus is on English-only tasks.

Finally, despite Medical-mT5-xl being larger than Medical-mT5-large (3B vs 738M), its performance is worse in the single-task evaluation setting. This behaviour is not observed in the multi-task and zero-shot experiments, leading us to hypothesize that the larger Medical-mT5-xl model is more prone to overfit in the single-task supervised setting.

\subsection{Abstractive Question Answering}

In this section we explore the text generation capabilities of Medical mT5 and other baseline text-to-text models on the BioASQ question answering dataset described in Section
\ref{sec:QA_explained}. Previous work typically evaluates the performance 
on this task using the ROUGE score (\cite{bioasq}) to
compare the gold standard answer with the answer generated by the model.
However, we find this metric inadequate for medical domain tasks as it does not
address crucial aspects of the generation such as factuality, potential harm, and bias
(\cite{singhal-palm}). Consequently, we enlisted medical
professionals to analyze the answers produced by the models.

\begin{table}[htb]
\centering
\adjustbox{max width=0.9\textwidth}{

\begin{tabular}{@{}lp{16cm}@{}}
\toprule
\rowcolor{CornflowerBlue!15} Ideal Answer & Which type of urinary incontinence is diagnosed with the Q tip test?  Stress urinary incontinence is diagnosed with the Q tip test. The test evaluates urethral mobility.  \\ \midrule
mT5-large & Q-tip test is used for urodynamic assessment of patients with severe urinary incontinence.  \\
FlanT5-large & The Q tip test is used for assessing urethral mobility.  \\
SciFive & Q tip test is used for evaluation of stress urinary incontinence. \\
MedMT5-large & Q-tip test is used for evaluation of stress urinary incontinence.  \\ \midrule
\rowcolor{CornflowerBlue!15} Ideal Answer & Which are the main manifestations of Ohdo syndrome? Severe ID, absent or deficient language, skeletal manifestations including bilateral patella dislocations.  \\ \midrule
mT5-large & Skeletal manifestations in Ohdo syndrome are a case with bilateral patella dislocations where surgical intervention has been indicated.  \\
FlanT5-large & The main manifestations of Ohdo syndrome are: 1) severe ID, 2) absent or deficient language and 3) milder, clinical manifestation in heterozygotes. \\
SciFive & Ohdo syndrome is characterized by severe ID, absent or deficient language and, milder, clinical manifestation in heterozygotes.  \\
MedMT5-large & The main manifestations of Ohdo syndrome are: 1) absent or deficient language and 2) mildder clinical manifestation in heterozygotes.  \\ \bottomrule
\end{tabular}}
\caption{Examples of answers generated by each model for two different BioASQ questions together with the rank assigned by medics.}
\label{tab7:BioASQ_example}
\end{table}

During annotation, medical doctors were displayed the question, the ideal gold answers and
the answers generated by each model. If required, they could also inspect
the snippets that provide context to answer each of the questions. We narrowed the
evaluation to Medical-mT5-large, mT5-large, FlanT5-large and SciFive. The evaluation
was conducted by medical doctors proficient/native speakers of English, French and Spanish. For each question, doctors were asked to rank the
answers generated by the models as the best, second-best, third-best, and worst
answer.

Two Spanish medical doctors proficient or native in English and Spanish analyzed 50 English examples and 252 Spanish. For the French language, 3 French clinicians analyzed 186 answers, of which 47 were done by 2 doctors to calculate IAA (Cohen's Kappa Score: 0.28 and Average Spearman's Rank Correlation: 0.48), which indicates a low level of agreement. This exercise provided interesting insights with respect to the performance of the models in text generation tasks in the medical domain. First, medical doctors could not in general establish significant differences between the
answers generated by each of the models; predictions were far too similar, and all tended to
fail on the same questions. As an example, Table \ref{tab7:BioASQ_example} shows the answers to two different questions. As it can be observed, the answers generated by each model are very similar, and the doctors ended up ranking them primarily based on style. 

The final result of the manual analysis is that all the models were chosen a similar number of times as the best. 
We believe that this demonstrates the difficulty of performing and obtaining meaningful evaluation results for this kind of tasks on this specific domain. This is supported by the low IAA agreement obtained in the French annotation. This issue has also emerged in prior research and was partially addressed by employing a very large number of experts and asking them to respond with a yes/no to a set of predefined potential issues in the model output (\cite{singhal-palm}). Still, the variance on the answers provided by the experts was significant.

However, there could be other underlying reasons for this behaviour. First, perhaps the T5 architecture is not ideally suited for text generation as formulated in the BioASQ task, as these models are trained on a masking reconstruction objective rather than on direct text generation tasks. Consequently, the knowledge acquired during pre-training might not generalize well when the models are subsequently trained for text generation purposes. Second, perhaps using much larger models such as MedPaLM (\cite{singhal-palm}) may generate better answer generation, but models of 540B parameters are currently unusable for the large majority of the NLP research labs, including ours. Nonetheless, it should be stressed that research on appropriate evaluation metrics for these tasks is still a difficult challenge which requires further investigation. 

In any case, our results demonstrate the potential of a text-to-text model such as Medical mT5 for multilingual sequence labeling in the medical domain, establishing new state-of-the-art results in the multi-task and zero-shot cross-lingual settings.

\section{Conclusion}

In this chapter, we have presented Medical mT5, the first open-source multilingual text-to-text LLM for the medical domain. Its development has required the compilation of a new 3B word corpus in English, French, Italian and Spanish specific to the medical domain. Furthermore, motivated by the lack of multilingual benchmarks, we have generated evaluation benchmarks for French, Italian and Spanish for Argument Mining and Abstractive Question Answering. 

A comprehensive experimentation on sequence labeling tasks shows that Medical mT5 outperforms strong text-to-text baselines of similarly-sized models in the multi-task and zero-shot cross-lingual evaluation settings. This is particularly interesting as these settings fully exploit the multilingual nature of a text-to-text model such as Medical mT5.

Furthermore, our experiments on Abstractive Question Answering show the inherent difficulty of evaluating generative tasks for this specific domain, where complex issues such as truthfulness and veracity are difficult to capture by automatic metrics. Manual evaluation is not ideal either, as medical doctors were not able to clearly distinguish between the quality of the answers generated by the different models. In line with previous work (\cite{singhal-palm}), we hope our research will bring further attention to this problem and encourage further research on evaluation methods.

Medical mT5 has been built on the work presented in previous chapters. We use the data transfer approach to develop new multilingual evaluation benchmarks for the medical domain. We also use constrained decoding to achieve high-quality zero-shot model-based cross-lingual transfer. Medical mT5 demonstrates the benefits of the techniques developed during this thesis and their application in real-world scenarios where data is scarce.

Regarding the languages chosen for this chapter, acquiring medical domain data is extremely challenging, even for languages such as the ones included. Furthermore, the choice of languages was also influenced by the availability of native medical doctors to do the manual evaluation for Abstractive Question Answering. In any case, we hope that our research will encourage more researchers to join our effort and gather data for their respective languages, thereby creating larger, multilingual medical domain datasets encompassing more languages in the future. 
\selectlanguage{english}
\chapter[Conclusion and future work]{Conclusion and future work}
\label{ch:final-chapter}

In this thesis, we have developed novel cross-lingual transfer learning methods aimed at addressing the resource constraints of low-resource languages. By leveraging both data-based and model-based approaches, we have demonstrated the potential to significantly improve performance on sequence labeling tasks across diverse languages and domains. Our proposed methods, including T-Projection and constrained decoding algorithms, achieve state-of-the-art results, highlighting the effectiveness of modern Machine Translation and multilingual models in facilitating knowledge transfer. The real-world application to the medical domain further underscores the practical impact of our research. By contributing open-source tools, datasets, and models, this work not only bridges the gap between high-resource and low-resource languages but also sets the stage for future advancements in multilingual NLP. The main contributions of this thesis are summarized as follows:

\begin{itemize}
    \item \textbf{We improved data-based cross-lingual transfer} approaches by developing a novel annotation projection method, namely \textbf{T-Projection} (\cite{garcia-ferrero-etal-2023-projection}) in Chapter \ref{ch:data-transfer}. It leverages state-of-the-art text-to-text multilingual models and Machine Translation systems to project annotations from high-resource to low-resource languages. T-Projection significantly outperforms previous annotation projection methods by a wide margin. This method allows us to \textbf{automatically generate high-quality labeled data for low-resource languages}.
    
    \item \textbf{We enhanced the model-based cross-lingual transfer approach} by, in Chapter \ref{fig:model_transfer} proposing a \textbf{constrained decoding algorithm} that enables model-based \textbf{cross-lingual transfer for sequence labeling tasks with text-to-text models}. This algorithm allows for the use of more powerful models that demonstrate superior zero-shot model-based transfer capabilities. Given the prevailing focus on text-to-text large language models (LLMs) in current research, and the infrequent training of new encoder-only models, this represents significant progress in the field.
    
    \item We expanded \textbf{NLP research in the medical domain for more languages} by developing a multilingual text-to-text open-source model for the medical domain, called \textbf{Medical mT5} (\cite{garcia-ferrero-etal-2024-medmt5}), presented in Chapter \ref{ch:medicalmt5}. By applying the model-based and data-based cross-lingual transfer learning methods developed in this thesis, we have implemented a \textbf{multilingual pre-training, fine-tuning, and evaluation framework for the medical domain}. Medical mT5 demonstrates the importance of the technology and knowledge developed in this thesis, resulting in the \textbf{first multilingual text-to-text medical model} at the time it was created.
    
    \item We conducted a \textbf{comprehensive evaluation of different cross-lingual transfer learning methods across a wide range of tasks, languages, and domains} (\cite{garcia-ferrero-etal-2022-model}), contributing to a better understanding of the situations in which each method is most effective. We demonstrated that \textbf{both our proposed data-based and model-based methods are effective in different scenarios}, and that they can be combined to achieve even better results. We also showed that the proposed methods are robust across different languages and domains and can be easily adapted to new tasks and languages.
    
    \item We released a large collection of \textbf{open-source software, datasets, and models} to facilitate the development of multilingual NLP research. By making our work \textbf{easily accessible} to other researchers, we enable them to \textbf{replicate} our experiments and \textbf{build upon our work}. We expect that the insights and methods developed in this thesis will be applicable to a wide range of NLP tasks, languages, and domains, thus contributing to the advancement of NLP in low-resource languages.
     
\end{itemize}

In terms of \textbf{publications}, this thesis contains 3 papers published in international conferences: 2 at EMNLP and 1 at LREC-COLING. Additionally, we published 3 closely related papers that were not included in this manuscript, 1 at ICLR, 1 at EMLP and 1 at an ACL Workshop. Finally, we submited other peer-reviewed papers, including 1 EMNLP, 2 at Ikergazte, 1 at ACL Workshops, and 1 at SEPLN. Between these papers, the paper \textit{Twitterreko Euskal Komunitatearen Eduki Azterketa Pandemia Garaian} was awarded the Most Relevant Research for the Development of the Basque Country \textbf{award at IkerGazte} 2021. Additionally, the \textit{NoticIA: A Clickbait Article Summarization Dataset in Spanish} project was selected as the \textbf{winner of the \#Somos600M 2024 Hackathon} by the SomosNLP community. This thesis has created interest in the community, as evidenced by the more than  \textbf{250 citations} of the papers published during this PhD, more than  \textbf{56,000 downloads} of the open-source models and datasets released on the Hugging Face Hub and more than  \textbf{550 stars} on the GitHub repositories. It also demonstrated by the  \textbf{talks} given at the OntarioTech University (graduate NLP course), the SEPLN symposium, and the Universitat de Barcelona (AI4HF consortium meeting).  
Finally, outside of the scope of the thesis, we have also contributed to the fair and unbiased evaluation of models by organizing The 1st Workshop on Data Contamination (CONDA) at ACL 2024 (\href{https://conda-workshop.github.io/}{https://conda-workshop.github.io/}, \cite{sainz-etal-2024-data}).

\section{Future work}

With the introduction of Large Language Models (LLMs), which have been pre-trained on massive amounts of text, instructions and further improved with reinforcement learning such as GPT4 (\cite{openai2024gpt4technicalreport}) or LLama-3 (\cite{llama3modelcard}), the field of NLP is evolving rapidly. The focus has shifted from training task-specific models to training multi-task models that can handle a wide range of tasks given a task description or prompt. Multi-task models have proven to outperform tasks specific models (\cite{DBLP:journals/corr/abs-2108-03265}) and can even perform well on tasks that they have not been trained on (\cite{brown2020language}). 

While these models are revolutionizing the field, they require a large amount of data, in terms of unstructured text and, more importantly, they require high-quality instruction-tuning data to achieve the capability of performing tasks given a prompt. This data includes a huge variety of tasks formulated as text-to-text instructions, such as dialogues, examples or summarization, code writing, translation, text generation, etc. Building large-scale instruction-tuning datasets is a very expensive and time-consuming process, therefore for now it has been limited to a very few companies with the monetary resources to do so. 

In this context, the next step in the field of cross-lingual transfer learning is to develop instruction-tuning datasets for low-resource languages. This requires the development of both, data-transfer and model-transfer methods, as well as overcoming other challenges such as adapting LLMs to the culture of the community in which they are to be used. Some of the future work that we plan to explore are:

\begin{itemize}
    \item Exploring the use of Machine Translation to generate instruction-tuning data for low-resource languages based on the already existing instruction-tuning datasets in high-resource languages. However, current sentence-level Machine Translation systems such as M2M100 (\cite{aharoni-etal-2019-massively}) or NLLB200 (\cite{DBLP:journals/corr/abs-2207-04672}) are not able to handle long contexts, which is a requirement for instruction-tuning data, which can be very long. In addition, translating complex structures that mix code, mathematical formulas, and other elements with natural language remains a challenge for current Machine Translation systems. While LLMs have shown great proficiency at document-level translation (\cite{xu2023paradigm,tower_llm_2024}) they are still only proficient at translating between high-resource languages (\cite{DBLP:journals/corr/abs-2311-07978,DBLP:conf/africanlp/OjoO23}). Developing a new generation of long-context Machine Translation systems, able to handle complex structures, for a wide range of languages can enable the generation of instruction-tuning data for low-resource languages.
    \item Synthetic data generation using LLMs (\cite{OpenHermes}) has shown promising results (\cite{zou2023representationengineeringtopdownapproach,DBLP:journals/corr/abs-2404-07503}). This process involves using an already pre-trained LLM and a set of prompts to generate instruction-tuning data for a wide range of tasks. This synthetic data can then be used to instruction-tune a new model that is superior to the original LLM. However, current methods are still limited to English. Model-based cross-lingual transfer can be used to generate synthetic data for low-resource languages. This means that a model pre-trained with unstructured text from many languages and instruction-tuned in only a few high-resource languages may be able to generate synthetic data for all the languages it has been pre-trained on. Similar to the model-based cross-lingual transfer experiments in this thesis, the model can be fine-tuned on a few examples of the target language and then used to generate synthetic data. This synthetic data can be used to train a model for the target language. A better understanding of the model-based cross-lingual transfer methods with LLMs and synthetic data generation methods can enable the development of instruction-tuning datasets for low-resource languages. 
    \item Building LLMs that can receive text in a language and produce plausible output text in that language is only the first step in developing LLMs for low-resource languages. For example, in the case of the Basque language, the latest generation of LLMs can process text in Basque. However, the models fail to answer questions pertinent to Basque culture, while they correctly answer questions about global culture (\cite{etxaniz2024bertaqalanguagemodelsknow}). For an LLM to be useful for a community, it must not only be able to process text in that language and produce plausible output text in the same language, but it must also encode knowledge about the community's culture. As shown in the experiments in Chapter \ref{ch:model-vs-data}, model-based and data-based cross-lingual transfer methods are not enough to overcome this issue. In fact, training the model with English-translated data can exacerbate the problem. Therefore, future research in cross-lingual transfer should not only focus on developing LLMs that can process text in a wide range of languages but also on finding methods to efficiently teach the models about the culture of the community in which the LLM is to be used.

\end{itemize}

\bibliography{bibliografia}
\cleardoublepage

\renewcommand{\thesection}{A.\arabic{section}} 
\phantomsection 
\addcontentsline{toc}{chapter}{Appendix}
\appendix
\selectlanguage{english}
\section{Original papers}

In this appendix, we present the original papers presented in the manuscript of this thesis in the recommended reading order. 
\cleardoublepage

\newcommand{\notitlesection}[1]{%
  \par\refstepcounter{section}
  \sectionmark{#1}
  \addcontentsline{toc}{section}{\protect\numberline{\thesection}#1}
}

\notitlesection{García-Ferrero et al. (Findings of the Association for Computational Linguistics: EMNLP 2022)}
\includepdf[scale=0.85, pages=-, lastpage=14, offset=-25 15]{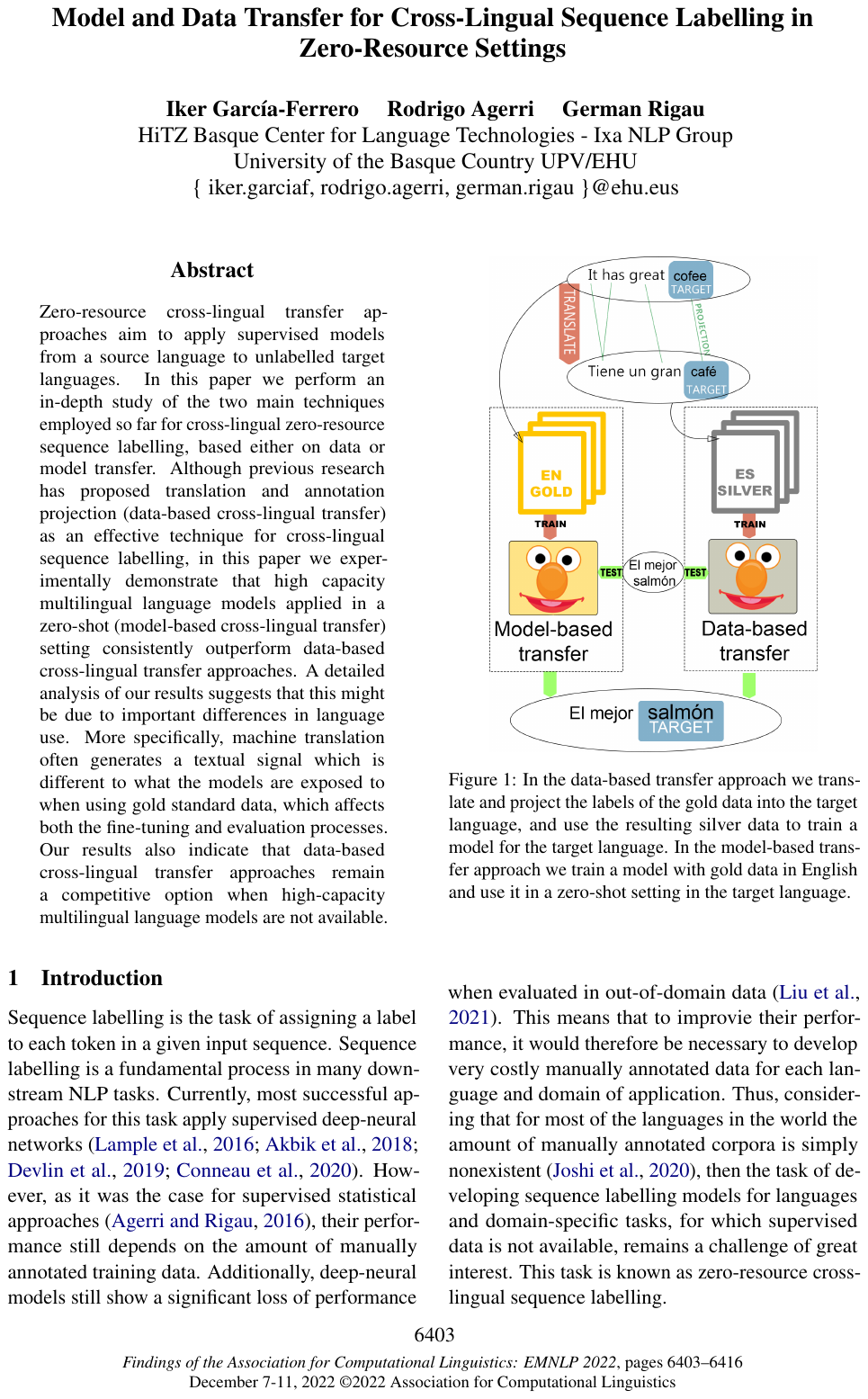}
\cleardoublepage

\notitlesection{García-Ferrero et al. (Findings of the Association for Computational Linguistics: EMNLP 2023)}
\includepdf[scale=0.85, pages=-, lastpage=15, offset=-25 15]{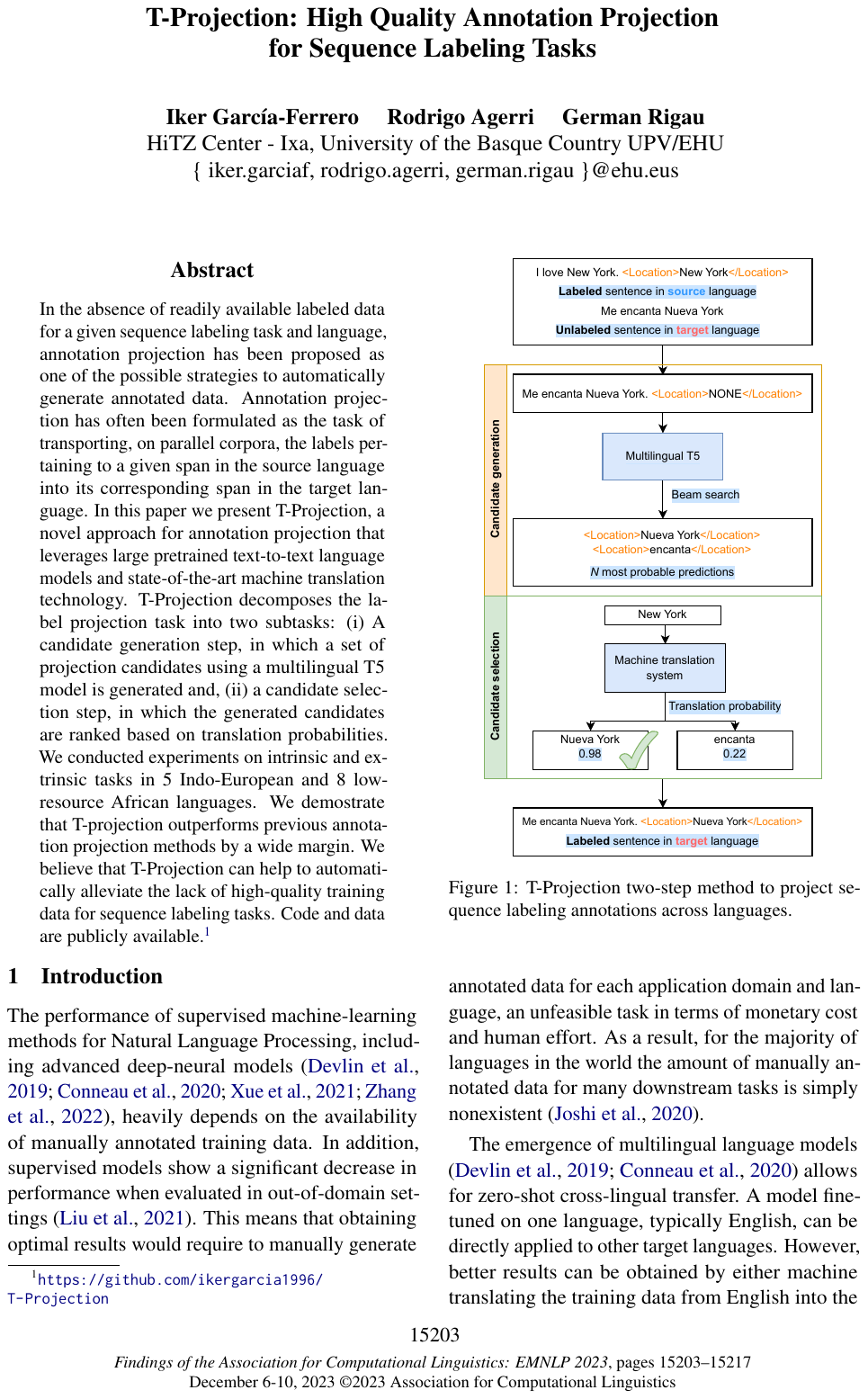}
\cleardoublepage

\notitlesection{García-Ferrero et al. (LREC-COLING 2024)}
\includepdf[scale=0.85, pages=-, lastpage=13, offset=-25 15]{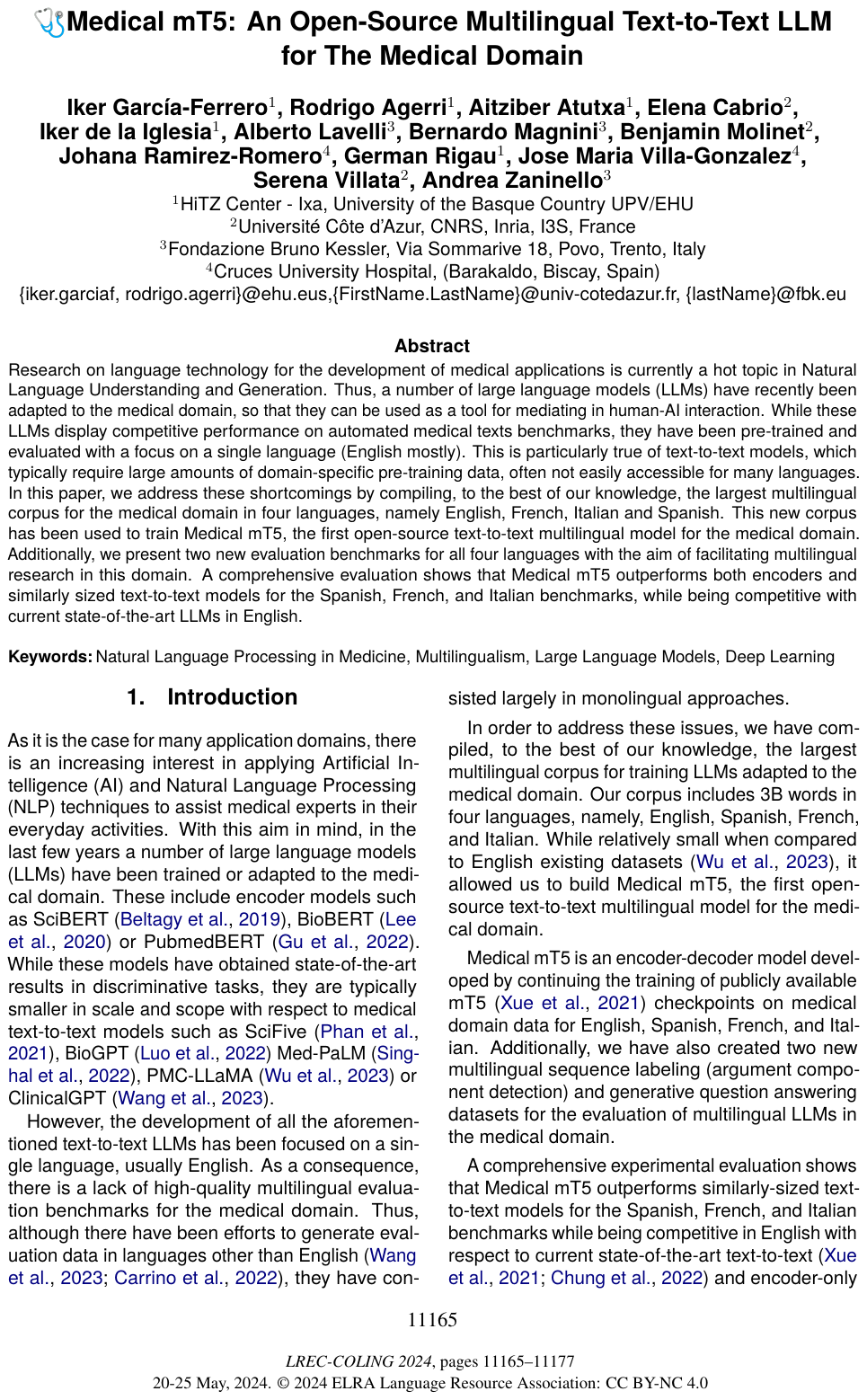}
\cleardoublepage

\cleardoublepage

\end{document}